\documentclass{article}
\usepackage{arxiv}
\usepackage[numbers,square,sort&compress]{natbib}

\usepackage{amsmath,amsfonts,bm}

\def\eqref#1{equation~\ref{#1}}

\def\1{\bm{1}}

\def\vzero{{\bm{0}}}

\def\va{{\bm{a}}}

\def\ve{{\bm{e}}}

\def\vh{{\bm{h}}}

\def\vr{{\bm{r}}}

\def\vu{{\bm{u}}}
\def\vv{{\bm{v}}}

\def\vx{{\bm{x}}}
\def\vy{{\bm{y}}}

\def\mI{{\bm{I}}}
\def\mJ{{\bm{J}}}

\DeclareMathAlphabet{\mathsfit}{\encodingdefault}{\sfdefault}{m}{sl}
\SetMathAlphabet{\mathsfit}{bold}{\encodingdefault}{\sfdefault}{bx}{n}

\newcommand{\method}{PerturbCellRL}
\providecommand{\vX}{{\bm{X}}}
\newif\ifeditmode
\editmodefalse
\AtBeginDocument{\ifeditmode\else\colorlet{blue}{black}\fi}

\usepackage[utf8]{inputenc}
\usepackage[T1]{fontenc}
\usepackage{url}
\usepackage{booktabs}
\usepackage{graphicx}
\usepackage{nicefrac}
\usepackage{microtype}
\usepackage{xcolor}
\usepackage{needspace}

\usepackage{colortbl}
\definecolor{light-light-gray}{gray}{0.92} 
\usepackage{wrapfig}
\usepackage{algorithm}
\usepackage{algpseudocode}
\let\arxivfloatalgorithm\algorithm
\renewcommand{\algorithm}[1][t]{\arxivfloatalgorithm[H]}
\usepackage{multirow}
\usepackage{enumitem}
\usepackage[normalem]{ulem}
\usepackage{array}
\usepackage{longtable}
\usepackage[section]{placeins}
\usepackage{amsthm}
\usepackage{amssymb}

\newcolumntype{R}[1]{>{\raggedleft\arraybackslash}p{#1}}

\newtheorem*{witnessrestatement}{Proposition~\ref{thm:witness}}
\newtheorem{corollary}{Corollary}
\newtheorem{proposition}{Proposition}
\newtheorem{lemma}{Lemma}
\theoremstyle{definition}

\theoremstyle{remark}
\newtheorem*{remark}{Remark}

\usepackage{hyperref}
\hypersetup{
  hidelinks,
  pdftitle={\method{}: Aligning Distributions and Grounding Biology via Post-Training Perturbation Generators},
  pdfauthor={Dongxia Wu, Mingyu Li, Yuhui Zhang, Anurendra Kumar, Emma Lundberg, Serena Yeung-Levy, Emily B. Fox}
}

\title{\method{}: Aligning Distributions and Grounding Biology via Post-Training Perturbation Generators}
\renewcommand{\shorttitle}{\emph{\method{}}}

\author{
    \textbf{Dongxia Wu}$^{1,2,*}$ \qquad \textbf{Mingyu Li}$^{1,*}$ \qquad
    \textbf{Yuhui Zhang}$^{1}$ \qquad \textbf{Anurendra Kumar}$^{1}$\\[4pt]
    \textbf{Emma Lundberg}$^{1,3}$ \qquad \textbf{Serena Yeung-Levy}$^{1}$ \qquad
    \textbf{Emily B. Fox}$^{1}$\\[8pt]
    {\normalfont\small $^{1}$Stanford University, Stanford, CA}\\[2pt]
    {\normalfont\small $^{2}$Mohamed bin Zayed University of Artificial Intelligence, Abu Dhabi, UAE}\\[2pt]
    {\normalfont\small $^{3}$ KTH Royal Institute of Technology, Stockholm, Sweden.}\\
    [6pt]
    {\normalfont\footnotesize\texttt{\{dowu,mrvlim,yuhuiz,anurendk,emmalu,syyeung,ebfox\}@stanford.edu}}\\[2pt]
    {\normalfont\footnotesize\texttt{dongxia.wu@mbzuai.ac.ae}}
}

\date{}

\newcommand{\blfootnote}[1]{\begingroup\renewcommand\thefootnote{}\footnotetext{#1}\addtocounter{footnote}{-1}\endgroup}

\begin{document}

\maketitle

\begin{NoHyper}
\blfootnote{$^*$ Equal contribution.}
\end{NoHyper}

\begin{abstract}

Single-cell perturbation models can reduce costly wet-lab screening
by predicting how cells respond transcriptionally to interventions.
Recent advances in flow-matching have enabled population-level prediction of cellular responses.
However, flow-matching training can fail to recover certain target distributions even within the model family, limiting its ability to capture cellular heterogeneity. We first prove that post-training can recover these
distributions, then introduce \method{}, a reinforcement learning framework that post-trains single-cell perturbation generators using per-cell rewards. The central component is a gene-expression energy witness
that translates population-level discrepancies into per-cell feedback. We further prove that this reward's policy gradient points toward better distributional alignment. Two complementary rewards, calibrated on real cells, penalize atypical expression
profiles and insufficient pathway-level responses to perturbations. Across genetic and chemical perturbation benchmarks \textcolor{blue}{and multiple generators (scDFM, scLDM)},
\method{} substantially improves distributional alignment and recovers pathway enrichment patterns more faithfully.
These results establish reward-guided post-training as an effective strategy for improving both distributional accuracy and biological fidelity in perturbation prediction.

\end{abstract}

\section{Introduction}
\label{sec:intro}

Predicting how single-cell gene expression responds to genetic or chemical perturbations is
a core capability of the envisioned virtual cell~\citep{bunne2024build,roohani2025virtual}. In single-cell perturbation screens,
a central task is to assess how an intervention changes a cell
population relative to a control. These changes can involve both shifts in
\textbf{average expression} and \textbf{cell-to-cell variation}. For example, the same \textcolor{blue}{shift in average expression could
come from a perturbed population in which all cells shift slightly,
or one in which a small subset shifts strongly.} A predictive model should therefore capture not
only average expression but also the heterogeneity of cellular responses\textcolor{blue}{~\citep{bunne2023learning,dong2023causal},
as measured by distributional metrics such as the energy distance~\citep{peidli2024scperturb,atanackovic2025meta,yu2026scdfm}}.
Since gene-expression measurements are destructive, control and perturbed cells are observed as unpaired
populations\textcolor{blue}{: the data reveal how the perturbed population is distributed, but not how each individual cell would have responded,}
making the response distribution a natural prediction target. Recent flow-matching models~\citep{lipman2022flow,liu2022flow} provide
a well-suited framework for learning these perturbation-specific distributions of single-cell gene expression~\citep{yu2026scdfm,klein2025cellflow}.

\begin{wrapfigure}[18]{R}[0pt]{0.5\textwidth}
\centering
\includegraphics[width=\linewidth]{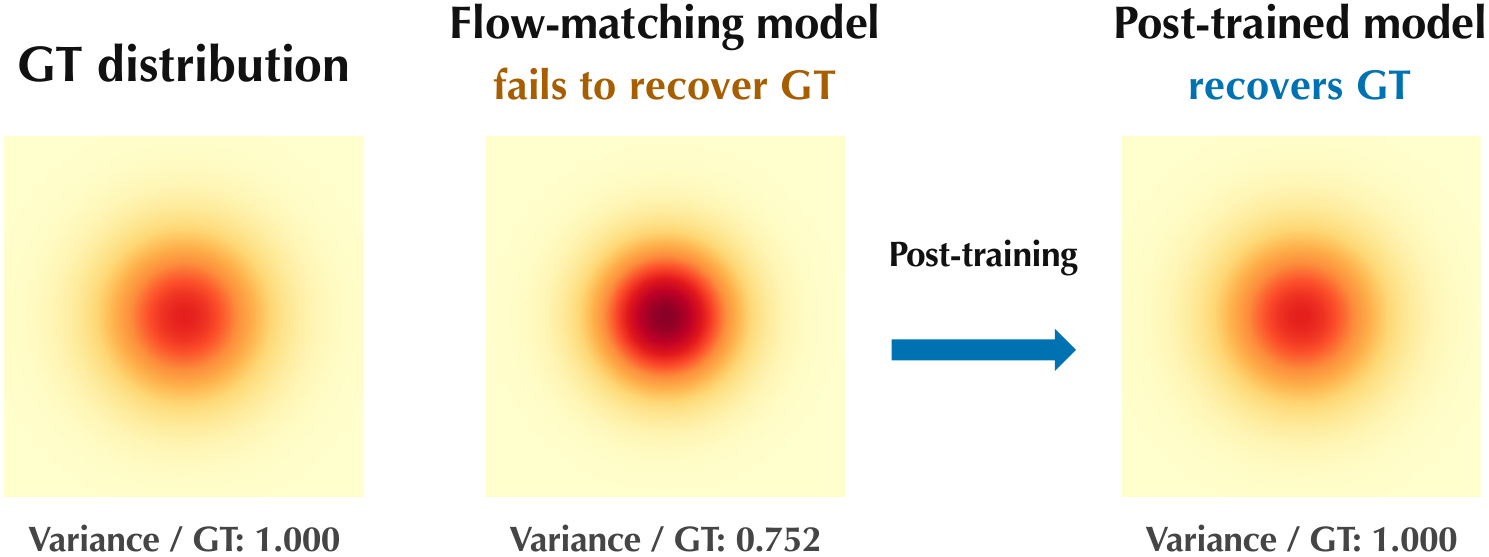}
\caption{\textbf{Post-training resolves a flow-matching mismatch.}
Left: the ground-truth (GT) density in a two-dimensional Gaussian example.
Middle: the flow-matching optimum has the correct mean
but only 75.2\% of the GT variance.
Right: post-training recovers the GT distribution
within the same model family (Proposition~\ref{prop:fm_counterexample}).}
\label{fig:fm_counterexample}
\end{wrapfigure}

However, via a counterexample, we prove that minimizing the flow-matching loss can fail to recover a target distribution that lies within the model family, while post-training recovers the target within the same family
(Figure~\ref{fig:fm_counterexample}).
This gap motivates post-training with direct distributional feedback
on generated cell populations.

Reinforcement learning provides a practical framework:
sample cellular responses, evaluate them with reward functions,
and update the generator to favor higher-reward
outputs~\citep{black2023training,fan2023dpok,liu2025flow,xue2025dancegrpo,li2025mixgrpo,zheng2025diffusionnft,wu2026cellfluxrl}.
We seek to assign a reward to each generated cell,
yet distributional agreement depends on the population as a whole.
The question is how to translate population discrepancies
into per-cell rewards that improve distributional alignment
while preserving cellular heterogeneity and biological fidelity.

To address this, we introduce \textbf{\method{}}, a reinforcement learning framework for post-training single-cell perturbation generators with per-cell rewards. Its central component is an energy witness in gene-expression space that scores each generated cell against generated and real target populations. This reward encourages proximity to real responses while discouraging crowding among generated cells. We prove that optimizing this per-cell reward can improve alignment between generated and target distributions under suitable assumptions. This connection provides a theoretical basis for using individual-cell feedback to improve population-level predictions. 

\textcolor{blue}{Because distributional feedback averages over all cells,
it can overlook individual cells that are atypical or lack the perturbation response.
We therefore complement it with two single-cell rewards.}
Expression support penalizes unusually extreme expression profiles,
while pathway discrimination penalizes weak evidence
for the requested perturbation relative to the control.
We use DiffusionNFT~\citep{zheng2025diffusionnft}
to integrate these rewards
and post-train the perturbation generator.

\textcolor{blue}{We post-train two flow-matching generators, scDFM~\citep{yu2026scdfm} and scLDM~\citep{palla2025scalable}.}
We evaluate \method{} \textcolor{blue}{with scDFM} on the Norman and ComboSciPlex benchmarks
for genetic and chemical perturbations.
Across four evaluation settings, \method{} achieves the best distributional alignments
among all compared methods, and also recovers pathway enrichment patterns more accurately.
\textcolor{blue}{With scLDM on the Replogle benchmark,
where test perturbations are unseen in the test cell line,
\method{} improves all nine evaluation metrics.}
Ablations examine how the three rewards jointly shape
distributional alignment, expression support, and pathway responses.

In summary, our theoretical analyses identify opportunities
to improve distributional alignment beyond flow-matching training alone.
Building on these insights, \method{} uses direct feedback
on generated responses to guide post-training toward
better distributional alignment and single-cell biological fidelity.
These improvements may support downstream applications in drug discovery
and personalized medicine by reducing reliance on costly wet-lab experiments.

\section{Related Work}
\label{sec:related_work}
\paragraph{Single-cell perturbation prediction with large-scale generative modeling.}
Single-cell perturbation prediction aims to infer how cellular transcriptomes change under genetic
or chemical interventions. This problem has received substantial attention, driven by the
increasing availability of large-scale perturbation datasets and advances in generative modeling.

On the data side, datasets such as  Norman~\citep{norman2019exploring}, ComboSciPlex~\citep{mathur2022combi} and Virtual Cell Challenge~\citep{roohani2025virtual}
provide
challenging genetic and chemical perturbation benchmarks.
scPerturb~\citep{peidli2024scperturb} highlights the broader
availability of harmonized single-cell perturbation data.

On the modeling side, early deep generative approaches, notably scVI~\citep{lopez2018deep},
established probabilistic representation learning for single-cell transcriptomics.
Subsequent perturbation prediction methods have used conditional autoencoders,
graph neural networks, transformer architectures, and generative models
to predict responses under unseen perturbations or cellular
contexts~\citep{lotfollahi2023predicting,roohani2024predicting,adduri2025predicting,bereketmodelling,10.1093/bioinformatics/btz158}.
More recently, state-of-the-art methods have begun to adopt flow
matching~\citep{lipman2022flow,lipman2024flow,liu2022flow}, which provides
a natural framework for conditional generation.
This is particularly well suited to perturbation prediction,
where models generate perturbed cells conditioned on control states
and perturbation labels~\citep{yu2026scdfm, klein2025cellflow, zhang2025cellflux}.
In this work, we build on this line of research, using scDFM~\citep{yu2026scdfm} as our base generator and further improving it through post-training.

\paragraph{Reinforcement learning for biological alignment.}
Although generative models can produce expression-like cell profiles,
incorporating biological priors into their predictions remains challenging.
This issue is especially important in single-cell perturbation
prediction, where a generated cell may appear realistic at the expression level while its perturbation effect is biologically inconsistent.
A trustworthy prediction should match the observed population-level
expression distribution, keep individual cells within the range of real
expression profiles, and carry the pathway-level response to the applied perturbation.
However, these objectives are often non-differentiable,
making them difficult to optimize directly.

One promising solution is reinforcement learning (RL), which has recently been used
to align diffusion and flow models with human preferences
or task-specific rewards~\citep{black2023training,fan2023dpok,liu2025flow,xue2025dancegrpo,li2025mixgrpo,wu2026cellfluxrl}.
Flow and diffusion models introduce challenges because
exact sample likelihoods are often intractable.
Prior work such as FlowGRPO and MixGRPO formulate
this problem as a Markov decision process, while recent
methods such as DiffusionNFT~\citep{zheng2025diffusionnft} introduce forward-process
objectives for online RL, improving both training efficiency
and stability.
In this work, we adapt this alignment perspective to
transcriptomic perturbation prediction, where rewards are derived
from distributional feedback and calibrated single-cell biological criteria
rather than visual quality or human-preference scores.
\section{Method}
\label{sec:method}

\subsection{Problem Setup and Motivation}
\label{sec:problem_setup}
\label{sec:problem_formulation}

\paragraph{Problem Setup.}
Let $\mathcal{X}\subseteq\mathbb{R}^G$ be the expression space over $G$ genes.
For perturbation $c\in\mathcal{C}$, let $P_0$ and $P_c$ denote
the control and target perturbed populations.
We use $\vu$ and $\vx_1$ for control
and real perturbed profiles, respectively.
Training triples $(\vu,c,\vx_1)$ are drawn from the observed data;
noise $\vx_0\sim\mathcal{N}(\vzero,\mI)$ and time $t\sim\mathcal{U}[0,1]$
are sampled independently of each other and these triples.

\textcolor{blue}{Conditional flow-matching generators} such as scDFM~\citep{yu2026scdfm}
learn a conditional velocity field $\vv_\theta$ by minimizing
\begin{equation}
\label{eq:flow_matching_objective}
\mathcal{L}_{\mathrm{FM}}(\theta)
=\mathbb{E}_{\vu,c,\vx_1,\vx_0,t}
\left[
\left\|\vv_\theta(\vx_t,t\mid\vu,c)-(\vx_1-\vx_0)\right\|_2^2
\right],
\qquad
\vx_t=(1-t)\vx_0+t\vx_1.
\end{equation}

The flow $\dot Z_t=\vv_\theta(Z_t,t\mid\vu,c)$, initialized at $Z_0=\vx_0$,
generates $\hat{\vx}_1=Z_1$ with conditional distribution $\pi_\theta(\cdot\mid\vu,c)$.
\textcolor{blue}{We retain scDFM's input $\vu$ to keep the base model unchanged;
since $\vu$ and $\vx_1$ are independent given $c$,
the flow-matching optimum does not depend on $\vu$ within a context.
Our second base model, scLDM~\citep{palla2025scalable}, applies flow matching
in a learned latent space without $\vu$ (Appendix~\ref{app:scldm}).}
Marginalizing over control profiles gives the generated population
\begin{equation}
\label{eq:generated_population}
Q_{\theta,c}(\mathrm{d}\hat{\vx}_1)
=
\int
\pi_\theta(\mathrm{d}\hat{\vx}_1\mid\vu,c)\,P_0(\mathrm{d}\vu).
\end{equation}

\paragraph{Motivation.}
Flow matching provides a strong starting point.
However, we show that flow matching can fail
to recover a target distribution even when that distribution
lies within the model family (Proposition~\ref{prop:fm_counterexample}).
Appropriate post-training can correct this mismatch.

\begin{proposition}[Flow-matching mismatch and recovery by post-training]
\label{prop:fm_counterexample}
There exists a model family whose parameterization includes the target distribution
\textcolor{blue}{but not the marginal velocity that flow matching regresses onto},
yet the minimizer of its flow-matching loss
generates a different distribution.
Starting from this minimizer, witness-based post-training within the same family
converges to the target distribution
for a sufficiently small constant step size.
\end{proposition}

\textcolor{blue}{Flow matching selects the member whose velocity is closest to this marginal velocity,
which need not be the member whose distribution is closest to the target.}
Intuitively, post-training pulls the generated distribution toward the target
by rewarding samples that improve population-level agreement.
Appendix~\ref{app:fm_counterexample} gives the construction and proof.

This mismatch motivates post-training with direct feedback on generated cells.
Starting from pretrained scDFM parameters $\theta_0$,
we aim to improve both population-level alignment and single-cell biological fidelity
while staying close to the pretrained generator.
We partition perturbations into disjoint sets
$\mathcal{C}_{\mathrm{train}}$ and $\mathcal{C}_{\mathrm{test}}$,
using only training conditions for post-training
and evaluating on held-out test conditions.
We optimize the following objective over training perturbations:
\begin{equation}
\label{eq:post_training_objective}
\begin{aligned}
\min_{\theta}\quad&
\mathbb{E}_{c\sim\mathrm{Unif}(\mathcal{C}_{\mathrm{train}})}
\left[
D\!\left(Q_{\theta,c},P_c\right)
-\lambda_{\mathrm{bio}}\,
\mathbb{E}_{\hat{\vx}_1\sim Q_{\theta,c}}
\left[r_{\mathrm{bio}}(\hat{\vx}_1,c)\right]
\right]
\\
\text{subject to}\quad&
\mathbb{E}_{\substack{
c\sim\mathrm{Unif}(\mathcal{C}_{\mathrm{train}})\\
\vu\sim P_0
}}
\left[
D_{\mathrm{KL}}\!\left(
\pi_\theta(\cdot\mid\vu,c)
\,\Vert\,
\pi_{\theta_0}(\cdot\mid\vu,c)
\right)
\right]\leq\epsilon.
\end{aligned}
\end{equation}
Here, $D$ measures population discrepancy
(e.g., energy distance),
$r_{\mathrm{bio}}$ rewards single-cell biological fidelity,
$\lambda_{\mathrm{bio}}\geq0$ balances the two objectives,
and $\epsilon$ bounds deviation from the pretrained generator.
\textcolor{blue}{A per-cell witness reward descends $D$ (Proposition~\ref{thm:witness}),
and the support and pathway rewards instantiate $r_{\mathrm{bio}}$ (Section~\ref{sec:single_cell_calibration}).}

\subsection{RL Post-Training}
\label{sec:rl_algorithm}

\setcounter{topnumber}{1}
\begin{figure}[t]
    \centering
    \includegraphics[width=\linewidth]{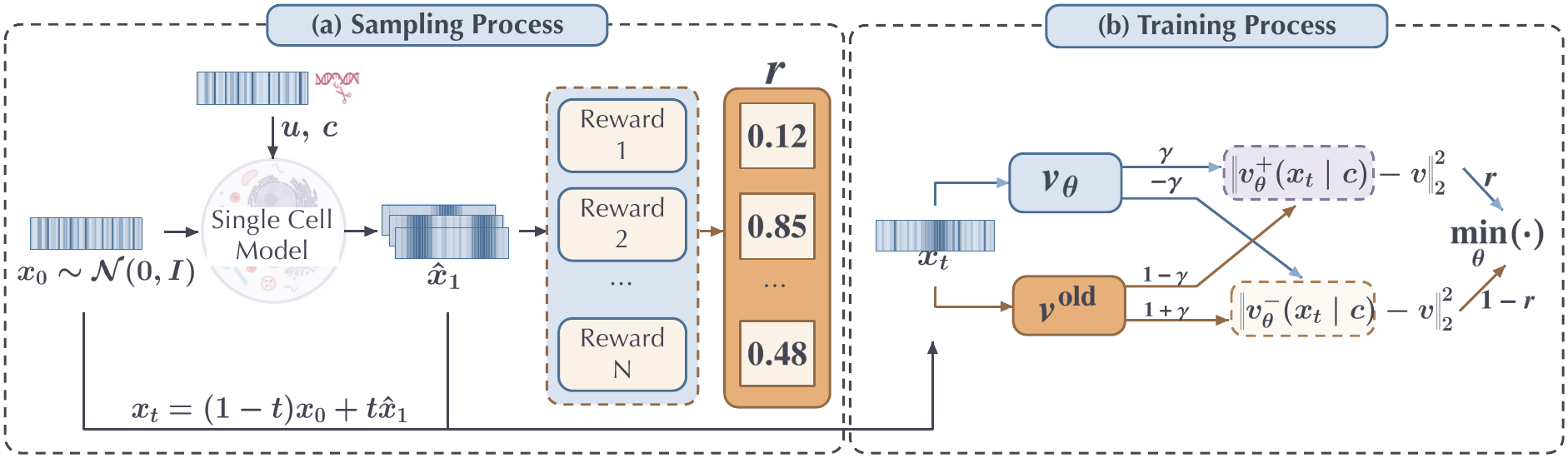}
    \caption{\textbf{\emph{\method{}} algorithm.}
    RL post-training seeks to increase the
    likelihood of high-reward samples and decrease
    the likelihood of low-reward samples.
    The core training loop consists of interleaved
    sampling and training phases.
    (a) Sampling: we generate multiple rollouts from
    a fixed control expression and perturbation condition,
    scoring each with the reward suite.
    (b) Training: because exact likelihoods under flow matching
    are intractable, we construct positive and negative velocities
    from the rollout batch and optimize them contrastively,
    following DiffusionNFT~\citep{zheng2025diffusionnft}.
    The branch scale $\gamma$ and preference weight $r$
    control the contrastive update.}
    \label{fig:algorithm}
\end{figure}

\paragraph{Algorithm overview.}
We adopt DiffusionNFT~\citep{zheng2025diffusionnft} for reward-guided velocity fitting.
Each RL iteration alternates between sampling and optimization:
we generate multiple cell profiles per control--perturbation pair,
score them using the reward suite in Section~\ref{sec:reward_models},
and update the model based on relative reward rankings within each group
(Figure~\ref{fig:algorithm}).

\paragraph{Algorithm detail.}
At the start of each iteration,
let $\vv^{\mathrm{old}}$ be a frozen copy of the current velocity field.
We use it to generate samples;
it remains fixed throughout optimization.
For a generated profile $\hat{\vx}_1$ conditioned on $(\vu,c)$,
define the training state and target velocity as
\begin{equation}
\label{eq:nft_training_state}
\vx_t=(1-t)\vx_0+t\hat{\vx}_1,
\qquad
\vv=\hat{\vx}_1-\vx_0,
\end{equation}
where $\vx_0\sim\mathcal{N}(\vzero,\mI)$ and $t\sim\mathcal{U}[0,1]$
are resampled independently at each training step.
All velocity fields below are evaluated at
\mbox{$(\vx_t,t\mid\vu,c)$}, with arguments omitted for brevity.
NFT defines positive and negative velocity branches
that interpolate toward and away from $\vv_\theta$
relative to $\vv^{\mathrm{old}}$:
\begin{equation}
\label{eq:implicit_pos}
\vv_\theta^+
=\gamma\vv_\theta+(1-\gamma)\vv^{\mathrm{old}},
\qquad
\vv_\theta^-
=-\gamma\vv_\theta+(1+\gamma)\vv^{\mathrm{old}},
\end{equation}
where $\gamma>0$ controls branch scale.
We standardize each cell's reward $R$ across rollouts
sharing the same control and perturbation pair
to obtain advantage $A$.
Section~\ref{sec:reward_combination} specifies reward construction.
Clipping and rescaling $A$ to $[-1,1]$ gives $\widetilde{A}$.
We define the preference weight and optimize
\begin{equation}
\label{eq:nft_loss}
r=\tfrac{1+\widetilde{A}}{2}\in[0,1],
\quad
\mathcal{L}(\theta)
=\mathbb{E}\Big[
 r\|\vv_\theta^+-\vv\|_2^2
 +(1-r)\|\vv_\theta^--\vv\|_2^2
+\lambda_{\mathrm{reg}}
\|\vv_\theta-\vv_{\theta_0}\|_2^2\Big].
\end{equation}
The preference weights $r$ and $1{-}r$
determine the relative strength of the positive and negative velocity targets.
The last term regularizes the model
toward the fixed pretrained generator $\vv_{\theta_0}$.
After optimization, we update the frozen sampling policy
$\vv^{\mathrm{old}}\leftarrow\vv_\theta$ and begin the next iteration.
Appendix~\ref{app:algorithm} provides the full procedure.

\subsection{Reward Suite}
\label{sec:reward_models}

\begin{figure}[t]
\centering
\includegraphics[width=\linewidth]{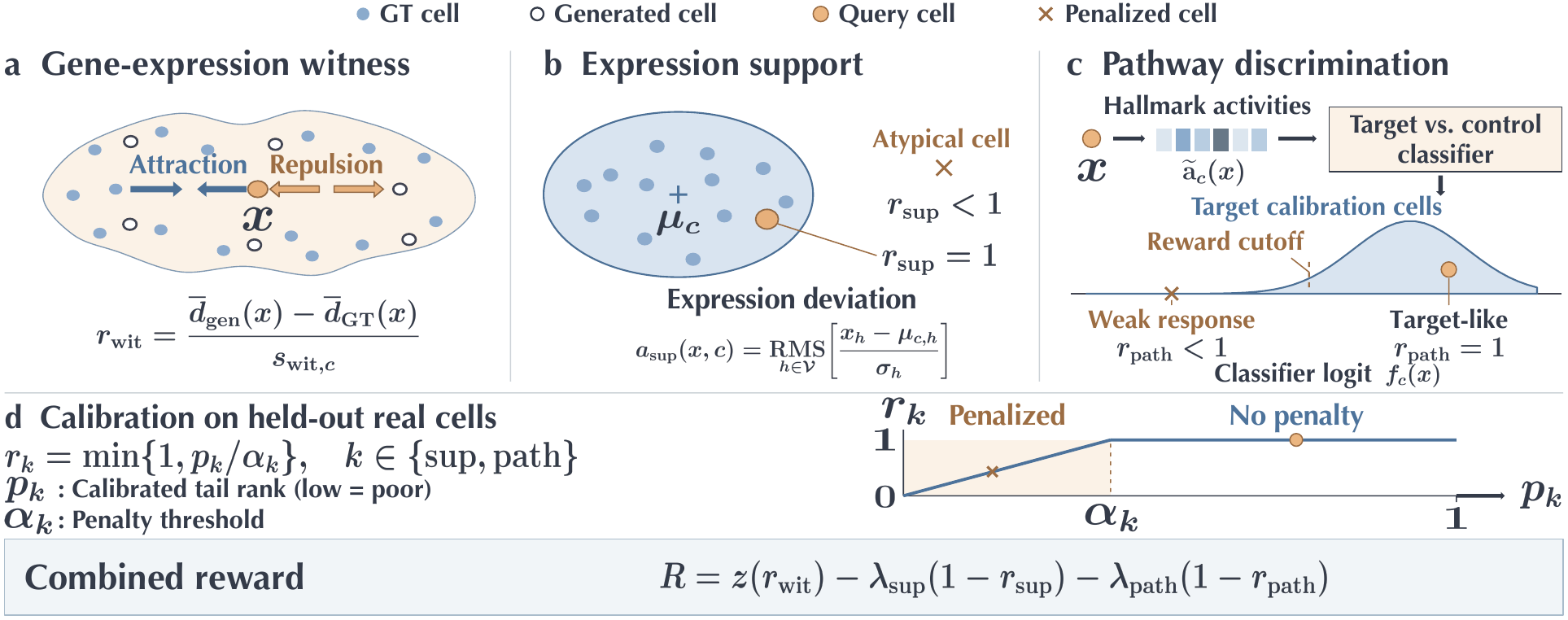}
\caption{\textbf{Complementary population and single-cell rewards.}
(a)~The gene-expression witness rewards closeness to real cells
while penalizing crowding among generated cells.
(b)~Expression support penalizes unusually large standardized radial deviations.
(c)~Pathway discrimination penalizes weak target-versus-control evidence
from Hallmark activities.
Bottom panels show calibration and reward combination:
tail ranks $p_k$ and cutoffs $\alpha_k$ define
$r_k=\min\{1,p_k/\alpha_k\}$ for $k\in\{\mathrm{sup},\mathrm{path}\}$.
}
\label{fig:reward_suite}
\end{figure}

Under ideal on-policy updates, per-cell witness rewards provide
a descent direction for population discrepancy, as shown in \textcolor{blue}{Proposition}~\ref{thm:witness}.
We instantiate this principle in gene-expression space,
then combine the population reward with
single-cell expression support and perturbation-specific pathway discrimination rewards,
as shown in Figure~\ref{fig:reward_suite}.

\subsubsection{Population Alignment via an Energy Witness}
\label{sec:population_witnesses}

We first establish a theoretical link between per-cell rewards and population-level distributional alignment.
Fix a perturbation $c$, with generated population $Q=Q_{\theta,c}$
and ground-truth population $P=P_c$.
For a fixed feature map $\phi:\mathcal{X}\to\mathbb{R}^d$
(we present the theory for general $\phi$ and instantiate it
with the identity map $\phi(\vx)=\vx$ on gene expression below),
let $d_\phi(\hat{\vx}_1,\vx_1)=\|\phi(\hat{\vx}_1)-\phi(\vx_1)\|_2$.
The energy distance in this space is
\begin{equation}
\label{eq:energy_distance}
D_\phi(Q,P)
=2\mathbb{E}_{Q\otimes P}d_\phi
-\mathbb{E}_{Q\otimes Q}d_\phi
-\mathbb{E}_{P\otimes P}d_\phi.
\end{equation}
We assign each cell the corresponding energy witness:
\begin{equation}
\label{eq:energy_witness}
w^\phi_{Q,P}(\hat{\vx}_1)
=\mathbb{E}_{\hat{\vx}_1'\sim Q}d_\phi(\hat{\vx}_1,\hat{\vx}_1')
-\mathbb{E}_{\vx_1\sim P}d_\phi(\hat{\vx}_1,\vx_1).
\end{equation}
Intuitively, this reward encourages generated cells to be close to real cells
while discouraging them from clustering together.

\begin{proposition}[Witness policy-gradient identity]
\label{thm:witness}
Suppose $P_c$ and $Q_{\theta,c}$ have finite first feature moments.
Assume conditional densities are differentiable with parameter-independent support,
and differentiation commutes with the relevant expectations.
Treat the current witness $w^\phi_{Q_{\theta,c},P_c}$ as a fixed reward.
Its on-policy gradient satisfies
\begin{equation}
\label{eq:witness_policy_gradient}
\begin{aligned}
g_\phi(\theta)
&:=\mathbb{E}_{\substack{\vu\sim P_0\\
\hat{\vx}_1\sim\pi_\theta(\cdot\mid\vu,c)}}
\left[w^\phi_{Q_{\theta,c},P_c}(\hat{\vx}_1)
\nabla_\theta\log\pi_\theta(\hat{\vx}_1\mid\vu,c)\right]
\\
&=-\frac{1}{2}\nabla_\theta D_\phi(Q_{\theta,c},P_c).
\end{aligned}
\end{equation}
Whenever $g_\phi(\theta)\ne0$, it is a descent direction for $D_\phi$.
\end{proposition}

The witness thus provides per-cell credit that yields a policy gradient
reducing population discrepancy.
When discrepancy reaches zero, the generated and target distributions
agree in the chosen feature space.
\textcolor{blue}{The identity follows from the score-function identity
and the first variation of the energy distance~\citep{bellemare2017cramer};
Appendix~\ref{app:witness_theory} gives the derivation and this characterization.}
\textcolor{blue}{Appendix~\ref{app:transfer} extends this descent property to the implemented NFT update.}

\paragraph{Energy witness in gene-expression space.}
The population reward $r_{\mathrm{wit}}$ measures distributional agreement in gene-expression space.
We use the identity map $\phi_{\mathrm{wit}}(\vx)=\vx$
on the current gene panel.
Using real responses $\{\vx_{1,j}\}$ from $P_c$, we define
\begin{equation}
\label{eq:gene_witness_reward}
r_{\mathrm{wit},i}=\frac{\widehat{w}^{\phi_{\mathrm{wit}}}_i}{s_{\mathrm{wit},c}},
\qquad
s_{\mathrm{wit},c}=\left(
\operatorname{median}_{j<\ell}\|\vx_{1,j}-\vx_{1,\ell}\|_2^2
\right)^{1/2}>0,
\end{equation}
where $\widehat{w}^{\phi_{\mathrm{wit}}}_i$ estimates the energy witness
$w^{\phi_{\mathrm{wit}}}_{Q_{\theta,c},P_c}(\hat{\vx}_{1,i})$ in \eqref{eq:energy_witness}
from generated and real cells;
Appendix~\ref{app:empirical_witness} details the sampling and estimator.
Because $s_{\mathrm{wit},c}$ is a positive constant derived from the target population,
this scaling preserves the gradient direction established in \textcolor{blue}{Proposition}~\ref{thm:witness}.

\subsubsection{Single-Cell Fidelity via Calibration}
\label{sec:single_cell_calibration}

\textcolor{blue}{The energy distance changes little when a few generated cells are atypical or lack the perturbation response:
if a fraction $\varepsilon$ comes from a wrong distribution $R$,
then $D_\phi((1-\varepsilon)P+\varepsilon R,P)=\varepsilon^2 D_\phi(R,P)$, only $0.25\%$ of $D_\phi(R,P)$ for $\varepsilon=5\%$.
We therefore add two single-cell rewards, expression support and pathway discrimination,
which penalize only cells that score worse than typical held-out real cells.}

\paragraph{Expression support.}
To construct the expression support reward $r_{\mathrm{sup}}$,
we measure how far a cell deviates
from the typical expression profile of its condition.
Let $\mu_{c,h}$ be the reference mean for gene $h$
under condition $c$, and $\sigma_h$ its standard deviation
across reference cells.
For variable genes $\mathcal{V}=\{h:\sigma_h>0\}$,
define the radial nonconformity score
\begin{equation}
\label{eq:support_nonconformity}
a_{\mathrm{sup}}(\vx,c)=\left[
\frac{1}{|\mathcal{V}|}\sum_{h\in\mathcal{V}}
\left(\frac{x_h-\mu_{c,h}}{\sigma_h}\right)^2
\right]^{1/2}.
\end{equation}
Higher values indicate more extreme expression profiles.
Let $p_{\mathrm{sup}}(\vx,c)$ be the upper-tail rank of this score
among real calibration scores pooled across conditions.
A small $p_{\mathrm{sup}}$ indicates an unusually large radial deviation.
The expression support reward is
\begin{equation}
\label{eq:support_reward}
r_{\mathrm{sup}}(\vx,c)=\min\left\{1,\frac{p_{\mathrm{sup}}(\vx,c)}{\alpha_{\mathrm{sup}}}\right\},
\qquad \alpha_{\mathrm{sup}}\in(0,1).
\end{equation}

\paragraph{Pathway discrimination.}
Expression support alone does not assess whether a cell
carries the response associated with the requested perturbation.
The pathway discrimination reward $r_{\mathrm{path}}$ assesses evidence
for this response relative to control cells.
We represent these responses using Hallmark gene sets~\citep{liberzon2015molecular}.
Let $\mathcal{G}_p$ contain pathway $p$'s genes
present in the current gene panel.
For pathways with $|\mathcal{G}_p|\geq5$,
we define the standardized activity vector directly:
\begin{equation}
\label{eq:pathway_activity}
\widetilde{\va}_c(\vx)=
\left(
\frac{|\mathcal{G}_p|^{-1}\sum_{h\in\mathcal{G}_p}x_h-m_{c,p}}{s_{c,p}}
\right)_{p:s_{c,p}>0},
\end{equation}
where $m_{c,p}$ and $s_{c,p}$ are pathway reference means
and standard deviations.

Following the treatment-versus-control classification approach of Augur~\citep{skinnider2021cell},
we fit a separate $\ell_1$-regularized logistic classifier for each condition.
The classifier logit is
$f_c(\vx)=\boldsymbol\omega_c^\top\widetilde{\va}_c(\vx)+b_c$.
Higher logits indicate stronger target-versus-control evidence.
Let $p_{\mathrm{path}}(\vx,c)$ be the lower-tail rank of this logit
among held-out real target cells under condition $c$.
A small $p_{\mathrm{path}}$ indicates unusually weak response evidence.
The pathway discrimination reward is
\begin{equation}
\label{eq:discrimination_reward}
r_{\mathrm{path}}(\vx,c)=\min\left\{1,\frac{p_{\mathrm{path}}(\vx,c)}{\alpha_{\mathrm{path}}}\right\},
\qquad \alpha_{\mathrm{path}}\in(0,1).
\end{equation}

Both rewards penalize cells whose rank falls below the threshold;
above it, the reward saturates at~$1$, avoiding incentives to minimize deviation
or maximize evidence beyond what real cells exhibit.
This leaves population matching to the gene-expression witness.
Appendix~\ref{app:single_cell_calibration} provides details.

\subsubsection{Reward Combination}
\label{sec:reward_combination}

The population reward $r_{\mathrm{wit}}$ is combined with
single-cell rewards $r_{\mathrm{sup}},r_{\mathrm{path}}\in[0,1]$.
Let $z$ denote standardization across rollouts
sharing the same control and perturbation pair.
Positive weights $\lambda_{\mathrm{sup}},\lambda_{\mathrm{path}}$ scale the two single-cell penalty terms.
The standardized gene-expression witness has unit weight.
For cell $i$, we define the combined reward
\begin{equation}
\label{eq:reward_combination}
R_i=z(r_{\mathrm{wit}})_i
-\lambda_{\mathrm{sup}}(1-r_{\mathrm{sup},i})-\lambda_{\mathrm{path}}(1-r_{\mathrm{path},i}).
\end{equation}
The gene-expression witness is standardized before being combined,
while the single-cell rewards retain their calibrated scales.
We then standardize $R$ within each group
to obtain the advantage $A_i=z(R)_i$,
which \eqref{eq:nft_loss} maps to the preference weight $r_i$
after clipping and rescaling.
The single-cell penalties thus modulate the relative ranking of rollouts
within each control--perturbation group, penalizing individually implausible cells
even when the population-level witness is favorable.

\setcounter{topnumber}{2}

\Needspace{8\baselineskip}
\section{Experiments}
\label{sec:experiments}

\subsection{Experimental Setup}
\label{sec:exp_details}

\begin{wrapfigure}{R}{0.45\textwidth}
\centering
\includegraphics[width=\linewidth]{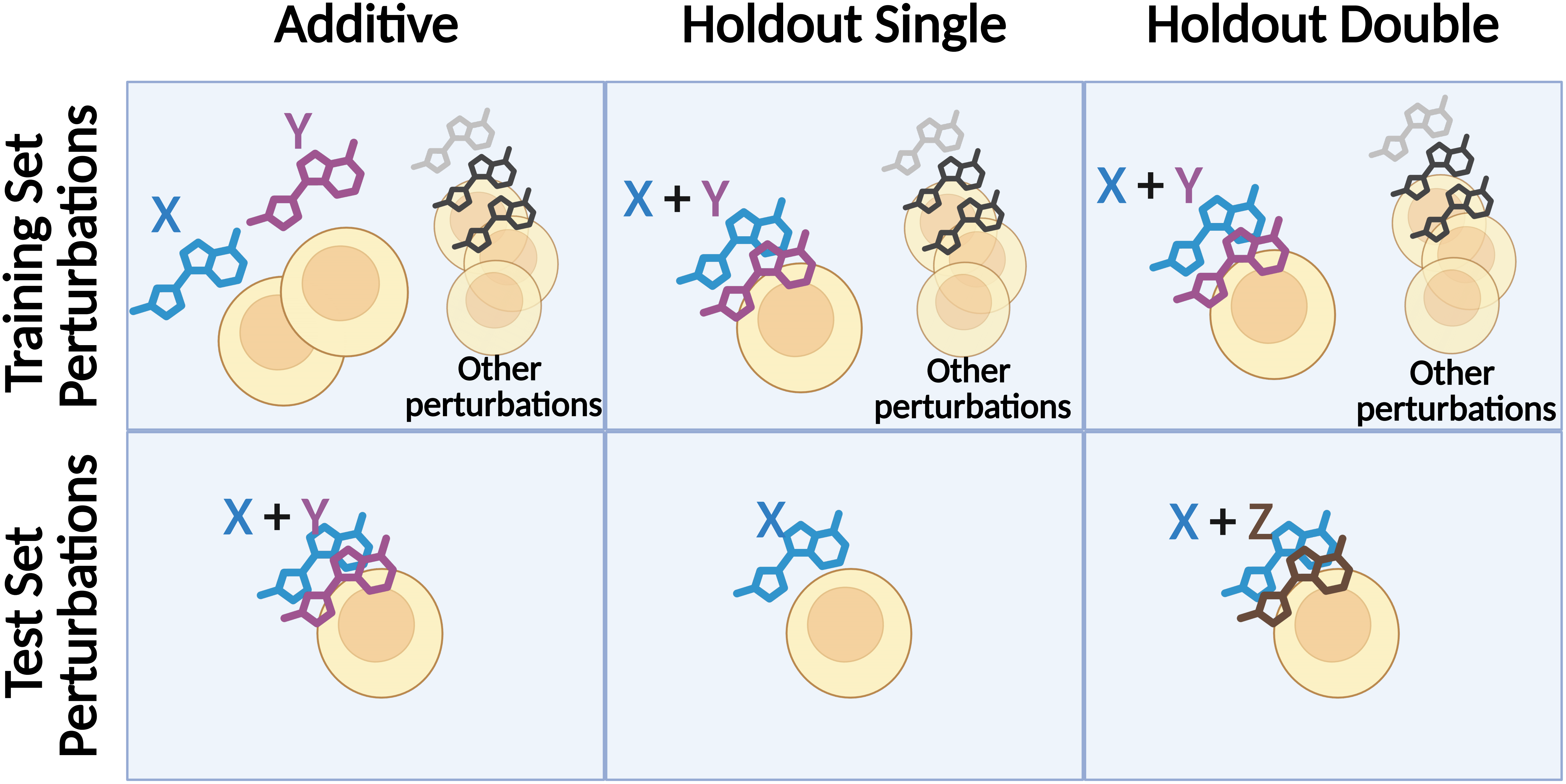}
\caption{Additive and holdout split protocols. Top rows show training perturbations; bottom rows show held-out test perturbations.}
\label{fig:additive_holdout}
\end{wrapfigure}
\paragraph{Datasets.}
We evaluate on the Norman~\citep{norman2019exploring}
and ComboSciPlex~\citep{mathur2022combi} datasets,
which cover genetic and chemical perturbations, respectively.
Following scDFM~\citep{yu2026scdfm}, we evaluate two Norman
train-test split protocols, illustrated in
Figure~\ref{fig:additive_holdout}.
In the additive split, all single-gene perturbations and a
subset of double-gene perturbations are used for training,
and the model predicts held-out double-gene perturbations.
In the holdout split, selected double-gene perturbations and
their constituent single-gene perturbations are held out for
testing, while the remaining perturbations are used for training.
We report holdout single and holdout double by averaging over
the single-gene and double-gene perturbations in the holdout
test set, respectively.
For both Norman split protocols, we use four random
train--test folds and report fold-averaged results.
\textcolor{blue}{For ComboSciPlex, we use the default scDFM split,
which holds out seven whole conditions (Appendix~\ref{app:dataset_details}).}
\textcolor{blue}{To test a second base model and transfer across cell lines,
we also use Replogle~\citep{replogle2022mapping}, Perturb-seq screens in four cell lines
(K562, RPE1, Jurkat, and HepG2), with the split of scLDM~\citep{palla2025scalable}:
$372$ gene edits are held out in HepG2,
and each is observed in training only in other cell lines (Appendix~\ref{app:scldm}).}

\paragraph{Baselines.}
We compare against Control (unperturbed cells),
Additive (a task-specific baseline for Norman additive
that linearly superposes single-gene effects),
GEARS~\citep{roohani2024predicting} (Norman only, as it models genetic perturbations),
CPA~\citep{lotfollahi2023predicting},
STATE~\citep{adduri2025predicting},
CellFlow~\citep{klein2025cellflow},
and scDFM~\citep{yu2026scdfm}.
For each test combination, Additive independently samples 128 training cells
without replacement from each single perturbation and the control group.
It sums the two sampled single-perturbation means,
subtracts the sampled control mean, clips negative values to zero,
and replicates the resulting vector 128 times for distributional evaluation.
Our primary comparison is with scDFM, which
serves as both the base model and
the strongest flow-matching baseline.
\textcolor{blue}{On Replogle, we post-train and compare with scLDM~\citep{palla2025scalable},
a latent flow-matching model that conditions on a perturbation token instead of a control cell
and decodes counts with a negative binomial decoder.
We keep its released checkpoint, split, sampler, and default guidance weight,
changing only post-training (Appendix~\ref{app:scldm}).}

\begin{table}[!t]
\centering
\caption{
\textbf{Population-level performance across Norman\textcolor{blue}{,}
ComboSciPlex\textcolor{blue}{, and Replogle} settings.}
Energy, MMD, and sliced Wasserstein-1 (SW1) compare cell distributions;
NES-Sp. measures the concordance of Reactome GSEA normalized enrichment scores;
the remaining metrics compare pseudobulk means.
Bold indicates best; underline indicates second best.
Norman results are averaged over four folds; ComboSciPlex uses
the default scDFM split.
CSP denotes ComboSciPlex; Add., H-S, and H-D denote
Norman additive, holdout single, and holdout double, respectively.
\textcolor{blue}{Rep.\ denotes Replogle, where \method{} post-trains scLDM
and results are averaged over $3$ generation seeds.}
}
\label{tab:pop_all_settings}
\setlength{\tabcolsep}{1.8pt}
\renewcommand{\arraystretch}{1.0}
\begin{tabular}{@{}llrrr@{\hspace{6pt}}rrrrrr@{}}
\toprule
& & \multicolumn{3}{
>{\columncolor{light-light-gray}[0pt][0pt]}c@{\hspace{6pt}}}{
\textbf{Distribution}}
& \cellcolor{light-light-gray}\textbf{Pathway}
& \multicolumn{5}{
>{\columncolor{light-light-gray}[0pt][0pt]}c}{
\textbf{Pseudobulk-mean}} \\
\cmidrule(lr){3-5}\cmidrule(lr){6-6}\cmidrule(lr){7-11}
\textbf{Set.} & \textbf{Model}
& \textbf{\shortstack{Energy\\$\downarrow$}}
& \textbf{\shortstack{MMD\\$\downarrow$}}
& \textbf{\shortstack{SW1\\$\downarrow$}}
& \textbf{\shortstack{NES-Sp.\\$\uparrow$}}
& \textbf{\shortstack{MAE\\$\downarrow$}}
& \textbf{\shortstack{DE-Sp.\\$\uparrow$}}
& \textbf{\shortstack{$\rho(\hat{\Delta})$\\$\uparrow$}}
& \textbf{\shortstack{$\rho(\Delta)$\\$\uparrow$}}
& \textbf{\shortstack{DS\\$\uparrow$}} \\
\midrule

\multirow{6}{*}{\textbf{CSP}}
& Control
& 7.8378
& 0.5303
& 0.2100
& \multicolumn{1}{c}{-}
& 0.0687
& \multicolumn{1}{c}{-}
& -0.3725
& \multicolumn{1}{c}{-}
& 0.5714 \\
& CPA
& 6.1738
& 0.4844
& 0.1863
& 0.3359
& 0.0690
& 0.2490
& -0.3994
& 0.2501
& 0.5714 \\
& STATE
& 0.7204
& 0.0601
& 0.0661
& 0.7805
& 0.0330
& 0.7317
& 0.7904
& 0.7915
& 0.8571 \\
& CellFlow
& 2.2834
& 0.2113
& 0.1154
& 0.8868
& 0.0470
& \textbf{0.8655}
& 0.2085
& 0.8254
& 0.5714 \\
& scDFM
& \underline{0.5177}
& \underline{0.0462}
& \underline{0.0654}
& \underline{0.8919}
& \underline{0.0242}
& 0.8363
& \underline{0.8418}
& \underline{0.8679}
& \underline{0.8776} \\
& \textbf{\emph{\method{}}}
& \textbf{0.4498}
& \textbf{0.0366}
& \textbf{0.0587}
& \textbf{0.8986}
& \textbf{0.0231}
& \underline{0.8416}
& \textbf{0.8536}
& \textbf{0.8770}
& \textbf{0.8980} \\

\midrule

\multirow{8}{*}{\textbf{Add.}}
& Control
& 5.9999
& 0.3236
& 0.1988
& \multicolumn{1}{c}{-}
& 0.0384
& \multicolumn{1}{c}{-}
& -0.1285
& \multicolumn{1}{c}{-}
& 0.5135 \\
& Additive
& 4.6519
& 0.2188
& 0.1818
& 0.7860
& 0.0243
& 0.4971
& \underline{0.8377}
& \textbf{0.8907}
& \underline{0.9646} \\
& GEARS
& 5.0438
& 0.2522
& 0.1869
& 0.6567
& 0.0381
& 0.7905
& 0.5840
& 0.7092
& 0.8240 \\
& CPA
& 4.6570
& 0.2918
& 0.1769
& 0.2509
& 0.0397
& 0.2128
& -0.0511
& 0.3171
& 0.5131 \\
& STATE
& 0.8992
& 0.0627
& 0.0770
& 0.6430
& 0.0306
& 0.6604
& 0.6641
& 0.7492
& 0.7841 \\
& CellFlow
& 1.4261
& 0.1156
& 0.1035
& 0.7757
& \underline{0.0223}
& 0.8458
& 0.7464
& 0.8264
& 0.7905 \\
& scDFM
& \underline{0.5562}
& \underline{0.0459}
& \underline{0.0738}
& \underline{0.7876}
& 0.0232
& \textbf{0.9017}
& 0.8333
& 0.8820
& \textbf{0.9743} \\
& \textbf{\emph{\method{}}}
& \textbf{0.4865}
& \textbf{0.0367}
& \textbf{0.0672}
& \textbf{0.7973}
& \textbf{0.0219}
& \underline{0.9007}
& \textbf{0.8479}
& \underline{0.8895}
& 0.9558 \\

\midrule

\multirow{7}{*}{\textbf{H-S}}
& Control
& 4.6858
& 0.2611
& 0.1714
& \multicolumn{1}{c}{-}
& 0.0247
& \multicolumn{1}{c}{-}
& 0.2657
& \multicolumn{1}{c}{-}
& 0.5216 \\
& GEARS
& 4.2806
& 0.2314
& 0.1663
& 0.5602
& 0.0259
& 0.7276
& 0.5424
& 0.5986
& 0.8180 \\
& CPA
& 3.5537
& 0.2428
& 0.1504
& 0.0400
& 0.0285
& 0.2979
& 0.2441
& 0.2194
& 0.5230 \\
& STATE
& 0.8626
& 0.0670
& 0.0743
& 0.5166
& 0.0295
& 0.6624
& 0.3793
& 0.5259
& 0.6716 \\
& CellFlow
& 1.2079
& 0.1088
& 0.0890
& 0.5910
& \underline{0.0195}
& 0.7847
& 0.5401
& 0.5826
& 0.6344 \\
& scDFM
& \underline{0.5642}
& \underline{0.0581}
& \underline{0.0736}
& \underline{0.5947}
& 0.0203
& \textbf{0.8372}
& \underline{0.6848}
& \underline{0.7182}
& \textbf{0.8904} \\
& \textbf{\emph{\method{}}}
& \textbf{0.5021}
& \textbf{0.0478}
& \textbf{0.0671}
& \textbf{0.6153}
& \textbf{0.0195}
& \underline{0.8342}
& \textbf{0.7099}
& \textbf{0.7281}
& \underline{0.8630} \\

\midrule

\multirow{7}{*}{\textbf{H-D}}
& Control
& 6.2336
& 0.3224
& 0.2066
& \multicolumn{1}{c}{-}
& 0.0414
& \multicolumn{1}{c}{-}
& -0.1412
& \multicolumn{1}{c}{-}
& 0.5333 \\
& GEARS
& 5.3218
& 0.2594
& 0.1951
& 0.6901
& 0.0398
& 0.8121
& 0.5473
& 0.7146
& 0.8433 \\
& CPA
& 4.9475
& 0.2953
& 0.1853
& 0.0737
& 0.0437
& 0.1827
& -0.0840
& 0.2666
& 0.5333 \\
& STATE
& 1.2475
& 0.0841
& 0.0934
& 0.6511
& 0.0380
& 0.6859
& 0.4868
& 0.6561
& 0.6933 \\
& CellFlow
& 1.7584
& 0.1346
& 0.1133
& \underline{0.7639}
& 0.0266
& 0.8368
& 0.5088
& 0.7334
& 0.7244 \\
& scDFM
& \underline{0.6398}
& \underline{0.0455}
& \underline{0.0767}
& 0.7602
& \underline{0.0251}
& \textbf{0.8845}
& \underline{0.7435}
& \underline{0.8279}
& \textbf{0.9144} \\
& \textbf{\emph{\method{}}}
& \textbf{0.5961}
& \textbf{0.0388}
& \textbf{0.0718}
& \textbf{0.7657}
& \textbf{0.0246}
& \underline{0.8777}
& \textbf{0.7526}
& \textbf{0.8312}
& \underline{0.8911} \\

\midrule

\multirow{2}{*}{\textcolor{blue}{\textbf{Rep.}}}
& \textcolor{blue}{scLDM}
& \textcolor{blue}{2.3945}
& \textcolor{blue}{0.0154}
& \textcolor{blue}{0.1979}
& \textcolor{blue}{0.3706}
& \textcolor{blue}{0.1329}
& \textcolor{blue}{0.3978}
& \textcolor{blue}{0.3167}
& \textcolor{blue}{0.3270}
& \textcolor{blue}{0.7518} \\
& \textcolor{blue}{\textbf{\emph{\method{}}}}
& \textcolor{blue}{\textbf{2.1088}}
& \textcolor{blue}{\textbf{0.0098}}
& \textcolor{blue}{\textbf{0.1829}}
& \textcolor{blue}{\textbf{0.4030}}
& \textcolor{blue}{\textbf{0.1210}}
& \textcolor{blue}{\textbf{0.4304}}
& \textcolor{blue}{\textbf{0.3369}}
& \textcolor{blue}{\textbf{0.3514}}
& \textcolor{blue}{\textbf{0.7617}} \\

\bottomrule
\end{tabular}
\end{table}

\paragraph{Evaluation metrics.}
We evaluate distributions, pathway enrichment, and pseudobulk means.
Distribution-level metrics are Energy distance, maximum mean discrepancy (MMD),
and sliced Wasserstein-1 distance (SW1)~\citep{bonneel2015sliced}.
For pathway enrichment, we report Reactome GSEA-NES Spearman
(NES-Sp.), following the evaluation approach of PertAdapt~\citep{bai2026pertadapt}.
Pseudobulk-mean metrics are mean absolute error (MAE),
Pearson $\rho(\Delta)$ and $\rho(\hat{\Delta})$,
DE-Spearman, and discrimination score (DS).
We average metrics across test conditions within each setting
and, for Norman, across the four folds.
Appendix~\ref{app:evaluation_metrics} provides definitions and computational details.

\begin{table}[!t]
\centering
\color{blue}
\caption{
\textbf{Single-reward ablation.}
scDFM (first row of each block) is post-trained with each reward alone (wit: witness; sup: support; path: pathway)
or with all three (last row, \method{}).
Metrics, Norman fold averaging, and bold/underline follow Table~\ref{tab:pop_all_settings}.
}
\label{tab:holdout_reward_ablation}
\renewcommand{\arraystretch}{1.0}
\resizebox{\linewidth}{!}{%
\begin{tabular}{@{}cccrrr@{\hspace{6pt}}r@{\hspace{6pt}}rrrrr@{\hspace{6pt}}rr@{}}
\toprule
\multicolumn{3}{c}{Training rewards}
& \multicolumn{3}{c}{\cellcolor{light-light-gray}\textbf{Distribution}}
& \cellcolor{light-light-gray}\textbf{Pathway}
& \multicolumn{5}{c}{\cellcolor{light-light-gray}\textbf{Pseudobulk-mean}}
& \multicolumn{2}{c}{\cellcolor{light-light-gray}\textbf{Reward}} \\
\cmidrule(lr){1-3}\cmidrule(lr){4-6}\cmidrule(lr){7-7}\cmidrule(lr){8-12}\cmidrule(l){13-14}
$\mathrm{wit}$ & $\mathrm{sup}$ & $\mathrm{path}$
& \textbf{\shortstack{Energy\\$\downarrow$}}
& \textbf{\shortstack{MMD\\$\downarrow$}}
& \textbf{\shortstack{SW1\\$\downarrow$}}
& \textbf{\shortstack{NES-Sp.\\$\uparrow$}}
& \textbf{\shortstack{MAE\\$\downarrow$}}
& \textbf{\shortstack{DE-Sp.\\$\uparrow$}}
& \textbf{\shortstack{$\rho(\hat{\Delta})$\\$\uparrow$}}
& \textbf{\shortstack{$\rho(\Delta)$\\$\uparrow$}}
& \textbf{\shortstack{DS\\$\uparrow$}}
& \textbf{\shortstack{$\mathrm{sup}$\\$\uparrow$}}
& \textbf{\shortstack{$\mathrm{path}$\\$\uparrow$}} \\
\midrule
\multicolumn{14}{@{}l}{\textit{ComboSciPlex}} \\
$\times$ & $\times$ & $\times$
& 0.5177 & 0.0462 & 0.0654 & 0.8919 & 0.0242 & 0.8363 & 0.8418 & 0.8679 & \underline{0.8776} & 0.9305 & 0.8735 \\
$\checkmark$ & $\times$ & $\times$
& \textbf{0.4413} & \textbf{0.0358} & \textbf{0.0581} & \textbf{0.9028} & \textbf{0.0231} & \underline{0.8410} & \textbf{0.8567} & \textbf{0.8797} & \textbf{0.8980} & 0.9536 & \textbf{0.8811} \\
$\times$ & $\checkmark$ & $\times$
& 0.4852 & 0.0410 & 0.0620 & 0.8935 & \underline{0.0237} & 0.8398 & 0.8442 & 0.8689 & \underline{0.8776} & \underline{0.9600} & 0.8775 \\
$\times$ & $\times$ & $\checkmark$
& 0.5013 & 0.0442 & 0.0641 & \underline{0.8991} & 0.0240 & 0.8395 & 0.8477 & 0.8737 & \underline{0.8776} & 0.9349 & 0.8791 \\
\addlinespace[2pt]
$\checkmark$ & $\checkmark$ & $\checkmark$
& \underline{0.4498} & \underline{0.0366} & \underline{0.0587} & 0.8986 & \textbf{0.0231} & \textbf{0.8416} & \underline{0.8536} & \underline{0.8770} & \textbf{0.8980} & \textbf{0.9632} & \underline{0.8809} \\
\midrule
\multicolumn{14}{@{}l}{\textit{Norman additive}} \\
$\times$ & $\times$ & $\times$
& 0.5562 & 0.0459 & 0.0738 & 0.7876 & 0.0232 & 0.9017 & 0.8333 & 0.8820 & \underline{0.9743} & 0.9088 & 0.9464 \\
$\checkmark$ & $\times$ & $\times$
& \textbf{0.4794} & \textbf{0.0358} & \textbf{0.0665} & \textbf{0.8003} & \underline{0.0220} & \textbf{0.9038} & \textbf{0.8495} & \textbf{0.8908} & 0.9567 & 0.9283 & \underline{0.9474} \\
$\times$ & $\checkmark$ & $\times$
& 0.5174 & 0.0409 & 0.0701 & 0.7864 & 0.0224 & 0.8988 & 0.8410 & 0.8848 & 0.9606 & \textbf{0.9411} & 0.9451 \\
$\times$ & $\times$ & $\checkmark$
& 0.5601 & 0.0459 & 0.0738 & 0.7916 & 0.0234 & \underline{0.9026} & 0.8333 & 0.8836 & \textbf{0.9775} & 0.9077 & \textbf{0.9504} \\
\addlinespace[2pt]
$\checkmark$ & $\checkmark$ & $\checkmark$
& \underline{0.4865} & \underline{0.0367} & \underline{0.0672} & \underline{0.7973} & \textbf{0.0219} & 0.9007 & \underline{0.8479} & \underline{0.8895} & 0.9558 & \underline{0.9395} & 0.9470 \\
\midrule
\multicolumn{14}{@{}l}{\textit{Norman holdout single}} \\
$\times$ & $\times$ & $\times$
& 0.5642 & 0.0581 & 0.0736 & 0.5947 & 0.0203 & \underline{0.8372} & 0.6848 & 0.7182 & \underline{0.8904} & 0.9264 & 0.9079 \\
$\checkmark$ & $\times$ & $\times$
& \textbf{0.4965} & \textbf{0.0467} & \textbf{0.0664} & 0.6130 & \textbf{0.0194} & 0.8319 & \textbf{0.7118} & 0.7239 & 0.8532 & 0.9474 & 0.9102 \\
$\times$ & $\checkmark$ & $\times$
& 0.5355 & 0.0525 & 0.0702 & 0.5851 & 0.0199 & 0.8300 & 0.6992 & 0.7170 & 0.8600 & \textbf{0.9596} & 0.9072 \\
$\times$ & $\times$ & $\checkmark$
& 0.5672 & 0.0585 & 0.0739 & \underline{0.6136} & 0.0204 & \textbf{0.8417} & 0.6786 & \textbf{0.7290} & \textbf{0.8949} & 0.9231 & \textbf{0.9156} \\
\addlinespace[2pt]
$\checkmark$ & $\checkmark$ & $\checkmark$
& \underline{0.5021} & \underline{0.0478} & \underline{0.0671} & \textbf{0.6153} & \underline{0.0195} & 0.8342 & \underline{0.7099} & \underline{0.7281} & 0.8630 & \underline{0.9503} & \underline{0.9128} \\
\midrule
\multicolumn{14}{@{}l}{\textit{Norman holdout double}} \\
$\times$ & $\times$ & $\times$
& 0.6398 & 0.0455 & 0.0767 & 0.7602 & 0.0251 & \underline{0.8845} & 0.7435 & 0.8279 & \underline{0.9144} & 0.8978 & 0.9250 \\
$\checkmark$ & $\times$ & $\times$
& \underline{0.5979} & \textbf{0.0385} & \textbf{0.0716} & 0.7633 & \underline{0.0247} & 0.8772 & 0.7493 & 0.8290 & 0.8822 & 0.9215 & 0.9264 \\
$\times$ & $\checkmark$ & $\times$
& 0.6251 & 0.0420 & 0.0743 & 0.7471 & 0.0250 & 0.8745 & 0.7432 & 0.8254 & 0.8811 & \textbf{0.9377} & 0.9238 \\
$\times$ & $\times$ & $\checkmark$
& 0.6384 & 0.0456 & 0.0768 & \textbf{0.7755} & 0.0251 & \textbf{0.8877} & \underline{0.7498} & \textbf{0.8337} & \textbf{0.9378} & 0.8916 & \textbf{0.9328} \\
\addlinespace[2pt]
$\checkmark$ & $\checkmark$ & $\checkmark$
& \textbf{0.5961} & \underline{0.0388} & \underline{0.0718} & \underline{0.7657} & \textbf{0.0246} & 0.8777 & \textbf{0.7526} & \underline{0.8312} & 0.8911 & \underline{0.9260} & \underline{0.9295} \\
\bottomrule
\end{tabular}%
}
\end{table}

\subsection{Main Results}
\label{sec:results_main}

\paragraph{\method{} substantially improves distributional alignment.}
As shown in Table~\ref{tab:pop_all_settings},
\method{} achieves the lowest Energy distance, MMD, and SW1
among all compared methods in all four settings.
Relative to its pretrained counterpart scDFM,
\method{} reduces MMD by $14.7\%$--$20.8\%$
and Energy distance by $6.8\%$--$13.1\%$.
The SW1 distance decreases by $6.4\%$--$10.2\%$ across the four settings.
On ComboSciPlex, MMD decreases from $0.0462$ to $0.0366$,
while Energy distance decreases from $0.5177$ to $0.4498$.
These gains are consistent across both Norman split protocols,
including held-out single- and double-gene perturbations.
\textcolor{blue}{On Replogle, where test edits are unseen in HepG2,
post-training scLDM lowers Energy by $11.9\%$, MMD by $36.6\%$, and SW1 by $7.6\%$.}
These improvements show that reward-guided post-training
shifts generated cell populations closer to the target distributions.

\paragraph{\method{} improves pathway enrichment concordance.}
\method{} achieves the highest mean Reactome NES-Sp.
among all methods in all four settings\textcolor{blue}{,
and on Replogle it raises NES-Sp.\ of scLDM from $0.371$ to $0.403$}.
These correlations indicate that \method{} better recovers
pathway enrichment rankings, suggesting improved biological fidelity.

\paragraph{\method{} remains competitive in pseudobulk-mean prediction.}
\method{} improves MAE and both Pearson correlations over scDFM
in all four settings, achieving the lowest MAE
and highest $\rho(\hat{\Delta})$ throughout.
For $\rho(\Delta)$, it ranks first on ComboSciPlex and both holdout settings,
and second on Norman additive.
\textcolor{blue}{DS and DE-Sp.\ decrease slightly on Norman ($1.9$--$3.1\%$ and $0.1$--$0.8\%$; Section~\ref{sec:ablations})
and improve on ComboSciPlex.
On Replogle, \method{} improves all five pseudobulk-mean metrics over scLDM.}
The results indicate that post-training improves prediction of
mean perturbation-induced expression changes.

\begin{figure}[!htb]
\centering
\includegraphics[width=0.70\linewidth]{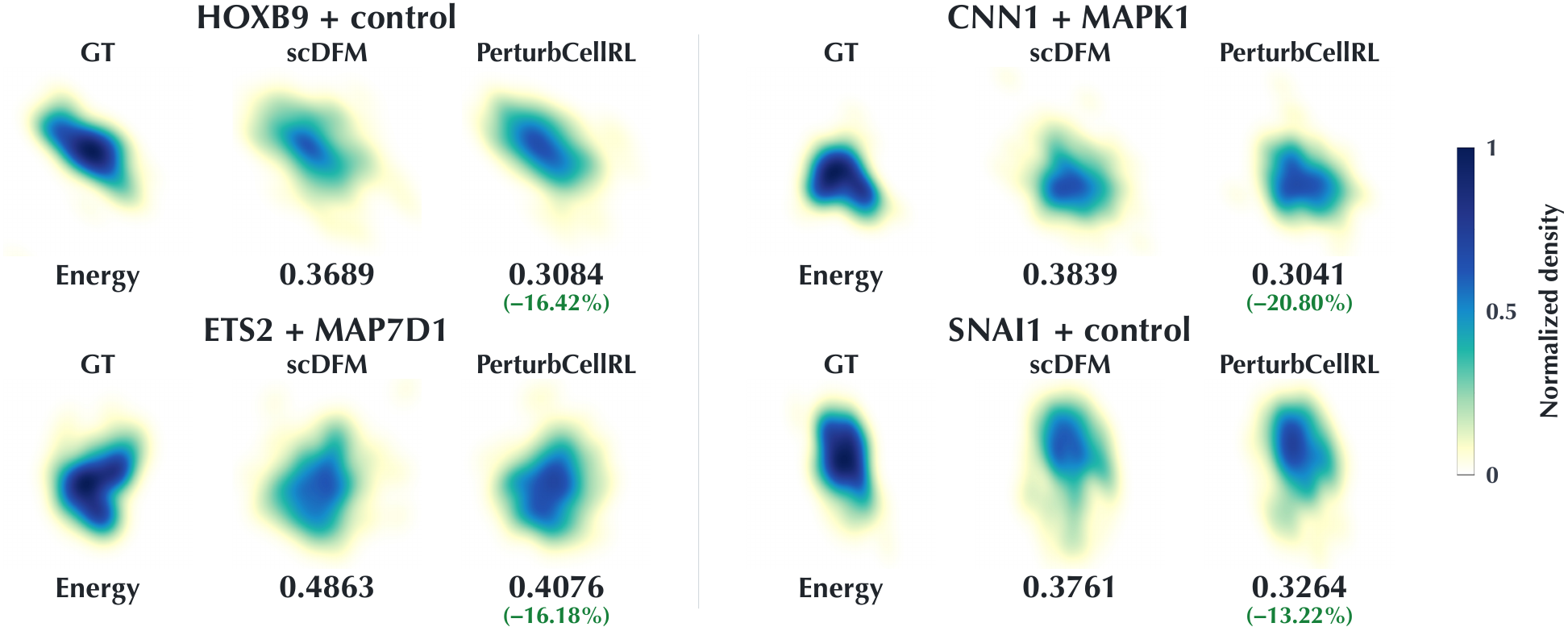}
\caption{
\textbf{Cell-density comparisons in UMAP space for four Norman holdout conditions.}
Each row compares ground truth (GT), scDFM, and \method{};
annotations report the Energy distance and its relative reduction.
}
\label{fig:visualization}
\end{figure}

\paragraph{Qualitative comparison.}
In target-fitted UMAP projections of held-out perturbations
(Figure~\ref{fig:visualization}; more in Appendix~\ref{app:additional_visualizations}),
\method{} predictions exhibit tighter overlap with the target high-density regions,
whereas scDFM often spreads into displaced areas of the embedding space.

\subsection{\texorpdfstring{\textcolor{blue}{Reward Ablations}}{Reward Ablations}}
\label{sec:ablations}

\begingroup\leavevmode\color{blue}%
Table~\ref{tab:holdout_reward_ablation} post-trains scDFM with each reward alone or all three;
$\mathrm{sup}$ and $\mathrm{path}$ are mean calibrated rewards of generated cells.
The \textbf{witness} reward targets distributional alignment:
alone, it gives the lowest MMD and SW1 in all four settings and the lowest Energy in three.
The \textbf{pathway} reward targets perturbation specificity:
alone, it improves NES-Sp., DE-Sp., and $\rho(\Delta)$ over scDFM in all four
and gives the highest DS and $\mathrm{path}$ on Norman.
The \textbf{support} reward targets single-cell plausibility:
alone, it gives the highest $\mathrm{sup}$ on Norman
and lowers Energy, MMD, SW1, and MAE over scDFM in all four.
\textbf{Combining all three} balances these gains:
\method{} is best or second best on every distribution metric in every setting,
improves every metric but DS and DE-Sp.\ over scDFM,
and recovers part of the DS that the witness alone loses on Norman holdout.
On scLDM, each reward again leads the metrics it targets (Appendix~\ref{app:scldm}).
\endgroup
\makeatletter
\@ifpackageloaded{arxiv}{}{\FloatBarrier}
\makeatother

\section{Conclusion}
\label{sec:conclusion}

We introduced \method{}, a framework that post-trains
single-cell perturbation generators using population-level and single-cell feedback.
Our approach addresses a key limitation of flow-matching generators:
the learned generator can fail to reproduce a target distribution even when it lies within the model family. We propose three complementary rewards that guide further optimization. A gene-expression energy witness translates population discrepancies into per-cell rewards,
while the calibrated expression-support and pathway-discrimination rewards evaluate individual generated cells for biological fidelity.
\textcolor{blue}{Applied to two flow-matching generators, scDFM and scLDM,}
\method{} improves distributional alignment and biological fidelity
across genetic and chemical perturbation benchmarks\textcolor{blue}{,
including perturbations that are unseen in the test cell line}.
Reward ablations \textcolor{blue}{show that the rewards are complementary.}
Together, these results establish reward-guided post-training
as a principled approach to aligning predicted cell populations with observed responses
while enforcing single-cell biological fidelity.

\paragraph{Limitations and future work.}
Our rewards require access to real target cells and are limited by the coverage of the selected gene and pathway features.
Moreover, the single-cell biology rewards are manually designed from domain expertise.
\textcolor{blue}{No cell type, cell line, or donor in our test sets is entirely absent from training.}
\textcolor{blue}{We retain scDFM's control-cell input $\vu$ to post-train it unchanged.
Within a context, the flow-matching optimum does not depend on $\vu$,
and whether $\vu$ helps in multi-context settings remains untested.
Our second base model, scLDM, has no such input (Appendix~\ref{app:scldm}).}
Future work could explore reference-free rewards with broader biological \textcolor{blue}{coverage,
transfer to unseen contexts,}
and prospective validation with domain biologists.

\section*{Acknowledgement}
This work was supported in part by ONR Grant N00014-22-1-2110 and the Stanford Institute for Human-Centered Artificial Intelligence (HAI). This work used Marlowe~\cite{marlowe2025}, Stanford University's GPU-based Computational Instrument, supported by Stanford HAI and Stanford Research Computing. EBF, SY, EL are Biohub, San Francisco, Investigator. EL and SY were supported by the Stanford Institute for Human-Centered AI. DW was supported by a startup grant from the Mohamed bin Zayed University of Artificial Intelligence (MBZUAI) under grant number 11-101-5050-62410-CMS000-BF0111-KD234-00. 

\bibliographystyle{unsrtnat}
\bibliography{main}

\newpage
\appendix
\clearpage
\section*{Appendix Contents}

\begingroup
\newcommand{\appendixentry}[2]{\noindent\textbf{\ref{#1}\quad\nameref{#1}}\nobreak\hfill\mbox{p.~\pageref{#1}}\par\nopagebreak
#2\par\medskip
}

\appendixentry{app:witness_theory}{We prove the witness policy-gradient identity
and its connection to energy-distance minimization.
}

\appendixentry{app:fm_counterexample}{We derive a flow-matching counterexample
and prove convergence of witness post-training within the same model family.
}

\appendixentry{app:algorithm}{We provide the post-training algorithm, empirical witness estimator,
implementation details, training \textcolor{blue}{hyperparameters, and computational cost.}
}

\appendixentry{app:single_cell_calibration}{We describe scorer fitting and held-out calibration
for the expression-support and pathway-discrimination rewards.
}

\appendixentry{app:evaluation_metrics}{We define the distribution-level, pathway-enrichment, and pseudobulk-mean metrics
used to evaluate perturbation predictions.
}

\appendixentry{app:dataset_details}{We document the Norman and ComboSciPlex dataset splits
and list their held-out perturbation conditions.
}

{\color{blue}\appendixentry{app:scldm}{To show that our post-training is a general framework rather than specific to scDFM,
we apply it to scLDM, a latent flow-matching generator with a different architecture and conditioning,
on its own Replogle benchmark with all three rewards,
where every test perturbation is held out in one cell line and seen in training only in other cell lines.
}}

{\color{blue}\appendixentry{app:transfer}{We show that the witness identity has an exact counterpart for the implemented NFT loss,
that with the exact witness the update has no stationary point away from the target up to the approximation error of the network and the reference term,
and that a functional NFT step changes the energy distance at first order by a descent term plus an explicit residual,
without assuming that the generator fits its own samples.
Finite-sample estimation, group standardization, and clipping enter the descent term only through one covariance.
}}

{\color{blue}\appendixentry{app:additional_quant}{We compare with a tuned classifier-free guidance baseline,
analyze how effect size affects its scores and the discrimination score,
and with direct differentiable energy-distance fine-tuning,
report variability across training seeds and folds,
track the energy distance during the first post-training updates,
and test whether generated cells depend on their control cell.
}}

\appendixentry{app:additional_visualizations}{We compare predicted and real cell densities
for additional double-gene and drug-combination perturbations.
}

\appendixentry{app:responsible_use}{We discuss intended use, potential prediction failures,
and the need for independent experimental validation.
}
\endgroup

\clearpage
\section{Witness Rewards and Distributional Alignment}
\label{app:witness_theory}

We restate \textcolor{blue}{Proposition}~\ref{thm:witness} and give the full derivation.

\begin{witnessrestatement}[Witness policy-gradient identity, restated]
Fix a perturbation $c$, control distribution $P_0$,
target distribution $P=P_c$, and feature map
$\phi:\mathcal{X}\to\mathbb{R}^d$.
Let $Q_{\theta,c}$ be the marginal distribution obtained by sampling
$\vu\sim P_0$ and then $\hat{\vx}_1\sim\pi_\theta(\cdot\mid\vu,c)$.
Define $d_\phi(\hat{\vx}_1,\vx_1)=\|\phi(\hat{\vx}_1)-\phi(\vx_1)\|_2$ and
\begin{equation}
\label{eq:witness_restatement_definitions}
\begin{aligned}
D_\phi(Q_{\theta,c},P)
&=2\mathbb{E}_{Q_{\theta,c}\otimes P}d_\phi
-\mathbb{E}_{Q_{\theta,c}\otimes Q_{\theta,c}}d_\phi
-\mathbb{E}_{P\otimes P}d_\phi,
\\
w^\phi_{Q_{\theta,c},P}(\hat{\vx}_1)
&=\mathbb{E}_{\hat{\vx}_1'\sim Q_{\theta,c}}d_\phi(\hat{\vx}_1,\hat{\vx}_1')
-\mathbb{E}_{\vx_1\sim P}d_\phi(\hat{\vx}_1,\vx_1).
\end{aligned}
\end{equation}
Suppose $P$ and $Q_{\theta,c}$ have finite first feature moments.
Assume conditional densities exist relative to a common dominating measure,
are differentiable, and have parameter-independent support.
Assume differentiation commutes with the relevant expectations,
and the resulting score-weighted integrals are finite.
Treat the current witness $w^\phi_{Q_{\theta,c},P}$ as a fixed reward.
Its on-policy gradient satisfies
\begin{equation}
\label{eq:witness_policy_gradient_restated}
\begin{aligned}
g_\phi(\theta)
&:=\mathbb{E}_{\substack{\vu\sim P_0\\
\hat{\vx}_1\sim\pi_\theta(\cdot\mid\vu,c)}}
\left[w^\phi_{Q_{\theta,c},P}(\hat{\vx}_1)
\nabla_\theta\log\pi_\theta(\hat{\vx}_1\mid\vu,c)\right]
\\
&=-\frac12\nabla_\theta D_\phi(Q_{\theta,c},P).
\end{aligned}
\end{equation}
Whenever $g_\phi(\theta)\ne0$, it is a descent direction for $D_\phi$.
\end{witnessrestatement}

\begingroup
\definecolor{witnesscross}{RGB}{0,102,153}
\definecolor{witnessself}{RGB}{163,74,0}
\definecolor{witnesstarget}{RGB}{85,85,85}
\newcommand{\wCross}[1]{\textcolor{witnesscross}{#1}}
\newcommand{\wSelf}[1]{\textcolor{witnessself}{#1}}
\newcommand{\wTarget}[1]{\textcolor{witnesstarget}{#1}}

\begin{proof}
We use direct differentiation of the energy-distance expectations,
as in \citet[Appendix A, Eq.~(8)]{bellemare2017cramer}.
Write $Q_\theta=Q_{\theta,c}$, $d=d_\phi$, and
$w_\theta=w^\phi_{Q_{\theta,c},P}$ throughout the proof.
The subscript on $w_\theta$ records which population defines the witness.
Only the generator depends on $\theta$;
$P_0$, $P$, and $d$ remain fixed.

\textbf{Step 1: Set up the distance terms.}
Let $\mathrm{d}\hat{\vx}_1$ denote integration against the common dominating measure.
The target $P$ need not have a density against this measure.
Let $q_\theta$ denote the density of $Q_\theta$.
The population definition in \eqref{eq:generated_population} reads
\begin{equation}
\label{eq:witness_marginal_density}
q_\theta(\hat{\vx}_1)
=\int\pi_\theta(\hat{\vx}_1\mid\vu,c)P_0(\mathrm{d}\vu).
\end{equation}
We distinguish three types of pairwise terms:
\wCross{generated--real}, \wSelf{generated--generated}, and \wTarget{real--real}.
The symbols $C_\theta$, $S_\theta$, and $R$ identify them independently
of color:
\begin{equation}
\label{eq:witness_pairwise_terms}
\begin{aligned}
\wCross{C_\theta}
&:=\mathbb{E}_{Q_\theta\otimes P}d
=\int\!\int d(\hat{\vx}_1,\vx_1)q_\theta(\hat{\vx}_1)
P(\mathrm{d}\vx_1)\,\mathrm{d}\hat{\vx}_1,
\\
\wSelf{S_\theta}
&:=\mathbb{E}_{Q_\theta\otimes Q_\theta}d
=\int\!\int d(\hat{\vx}_1,\hat{\vx}_1')q_\theta(\hat{\vx}_1)q_\theta(\hat{\vx}_1')
\,\mathrm{d}\hat{\vx}_1'\,\mathrm{d}\hat{\vx}_1,
\\
\wTarget{R}
&:=\mathbb{E}_{P\otimes P}d
=\int\!\int d(\vx_1,\vx_1')P(\mathrm{d}\vx_1')P(\mathrm{d}\vx_1).
\end{aligned}
\end{equation}
The density product in $S_\theta$ follows from independent sampling.
By definition,
\begin{equation}
\label{eq:witness_distance_decomposition}
D_\phi(Q_\theta,P)
=2\wCross{C_\theta}-\wSelf{S_\theta}-\wTarget{R}.
\end{equation}
For a fixed query $\hat{\vx}_1$, define its mean distances by
\begin{equation}
\label{eq:witness_mean_distances}
\begin{aligned}
\wCross{a(\hat{\vx}_1)}
&:=\int d(\hat{\vx}_1,\vx_1)P(\mathrm{d}\vx_1),
\\
\wSelf{b_\theta(\hat{\vx}_1)}
&:=\int d(\hat{\vx}_1,\hat{\vx}_1')q_\theta(\hat{\vx}_1')\,\mathrm{d}\hat{\vx}_1',
\\
w_\theta(\hat{\vx}_1)&=\wSelf{b_\theta(\hat{\vx}_1)}-\wCross{a(\hat{\vx}_1)}.
\end{aligned}
\end{equation}
Here $a$ is fixed, whereas $b_\theta$ depends on the generator.
Finite first feature moments make these distance expectations finite,
by $d(\hat{\vx}_1,\vx_1)\leq\|\phi(\hat{\vx}_1)\|_2+\|\phi(\vx_1)\|_2$.
All derivative and integration interchanges below use the stated assumptions.

\textbf{Step 2: Differentiate the generated--real term.}
In $C_\theta$, only $q_\theta(\hat{\vx}_1)$ depends on $\theta$.
Moving the derivative inside the integrals and collecting
terms independent of $\vx_1$ gives
\begin{equation}
\label{eq:witness_cross_derivative}
\begin{aligned}
\nabla_\theta\wCross{C_\theta}
&=\nabla_\theta\int\!\int
 d(\hat{\vx}_1,\vx_1)q_\theta(\hat{\vx}_1)P(\mathrm{d}\vx_1)\,\mathrm{d}\hat{\vx}_1
\\
&=\int\!\int d(\hat{\vx}_1,\vx_1)\nabla_\theta q_\theta(\hat{\vx}_1)
P(\mathrm{d}\vx_1)\,\mathrm{d}\hat{\vx}_1
\\
&=\int\left[\int d(\hat{\vx}_1,\vx_1)P(\mathrm{d}\vx_1)\right]
\nabla_\theta q_\theta(\hat{\vx}_1)\,\mathrm{d}\hat{\vx}_1
\\
&=\int\wCross{a(\hat{\vx}_1)}\nabla_\theta q_\theta(\hat{\vx}_1)\,\mathrm{d}\hat{\vx}_1.
\end{aligned}
\end{equation}

\textbf{Step 3: Differentiate both copies of the generated distribution.}
In $S_\theta$, both density factors depend on $\theta$.
The product rule therefore produces two terms:
\begin{equation}
\label{eq:witness_self_product_rule}
\begin{aligned}
\nabla_\theta\wSelf{S_\theta}
&=\nabla_\theta\int\!\int d(\hat{\vx}_1,\hat{\vx}_1')
q_\theta(\hat{\vx}_1)q_\theta(\hat{\vx}_1')\,\mathrm{d}\hat{\vx}_1'\,\mathrm{d}\hat{\vx}_1
\\
&=\int\!\int d(\hat{\vx}_1,\hat{\vx}_1')
\nabla_\theta[q_\theta(\hat{\vx}_1)q_\theta(\hat{\vx}_1')]
\,\mathrm{d}\hat{\vx}_1'\,\mathrm{d}\hat{\vx}_1
\\
&=\underbrace{\int\!\int d(\hat{\vx}_1,\hat{\vx}_1')
[\nabla_\theta q_\theta(\hat{\vx}_1)]q_\theta(\hat{\vx}_1')
\,\mathrm{d}\hat{\vx}_1'\,\mathrm{d}\hat{\vx}_1}_{\wSelf{I_1}}
\\
&\quad+\underbrace{\int\!\int d(\hat{\vx}_1,\hat{\vx}_1')
q_\theta(\hat{\vx}_1)[\nabla_\theta q_\theta(\hat{\vx}_1')]
\,\mathrm{d}\hat{\vx}_1'\,\mathrm{d}\hat{\vx}_1}_{\wSelf{I_2}}.
\end{aligned}
\end{equation}
For $I_1$, integrate over $\hat{\vx}_1'$ first:
\begin{equation}
\label{eq:witness_self_first_term}
\begin{aligned}
\wSelf{I_1}
&=\int\left[\int d(\hat{\vx}_1,\hat{\vx}_1')q_\theta(\hat{\vx}_1')
\,\mathrm{d}\hat{\vx}_1'\right]\nabla_\theta q_\theta(\hat{\vx}_1)\,\mathrm{d}\hat{\vx}_1
\\
&=\int\wSelf{b_\theta(\hat{\vx}_1)}
\nabla_\theta q_\theta(\hat{\vx}_1)\,\mathrm{d}\hat{\vx}_1.
\end{aligned}
\end{equation}
For $I_2$, first exchange the integration order.
Next, swap the dummy names $\hat{\vx}_1$ and $\hat{\vx}_1'$.
Finally, use the symmetry $d(\hat{\vx}_1',\hat{\vx}_1)=d(\hat{\vx}_1,\hat{\vx}_1')$:
\begin{equation}
\label{eq:witness_self_second_term}
\begin{aligned}
\wSelf{I_2}
&=\int\!\int d(\hat{\vx}_1,\hat{\vx}_1')q_\theta(\hat{\vx}_1)
\nabla_\theta q_\theta(\hat{\vx}_1')\,\mathrm{d}\hat{\vx}_1'\,\mathrm{d}\hat{\vx}_1
\\
&=\int\!\int d(\hat{\vx}_1,\hat{\vx}_1')q_\theta(\hat{\vx}_1)
\nabla_\theta q_\theta(\hat{\vx}_1')\,\mathrm{d}\hat{\vx}_1\,\mathrm{d}\hat{\vx}_1'
\\
&=\int\!\int d(\hat{\vx}_1',\hat{\vx}_1)q_\theta(\hat{\vx}_1')
\nabla_\theta q_\theta(\hat{\vx}_1)\,\mathrm{d}\hat{\vx}_1'\,\mathrm{d}\hat{\vx}_1
\\
&=\int\!\int d(\hat{\vx}_1,\hat{\vx}_1')q_\theta(\hat{\vx}_1')
\nabla_\theta q_\theta(\hat{\vx}_1)\,\mathrm{d}\hat{\vx}_1'\,\mathrm{d}\hat{\vx}_1
\\
&=\int\left[\int d(\hat{\vx}_1,\hat{\vx}_1')q_\theta(\hat{\vx}_1')
\,\mathrm{d}\hat{\vx}_1'\right]\nabla_\theta q_\theta(\hat{\vx}_1)\,\mathrm{d}\hat{\vx}_1
\\
&=\int\wSelf{b_\theta(\hat{\vx}_1)}
\nabla_\theta q_\theta(\hat{\vx}_1)\,\mathrm{d}\hat{\vx}_1
=\wSelf{I_1}.
\end{aligned}
\end{equation}
Combining \eqref{eq:witness_self_product_rule}, \eqref{eq:witness_self_first_term},
and \eqref{eq:witness_self_second_term} gives
\begin{equation}
\label{eq:witness_self_derivative}
\begin{aligned}
\nabla_\theta\wSelf{S_\theta}
&=\wSelf{I_1}+\wSelf{I_2}
\\
&=2\wSelf{I_1}
\\
&=2\int\wSelf{b_\theta(\hat{\vx}_1)}
\nabla_\theta q_\theta(\hat{\vx}_1)\,\mathrm{d}\hat{\vx}_1.
\end{aligned}
\end{equation}
This is the source of the factor $2$.

\textbf{Step 4: Combine all three distance terms.}
The real--real term depends only on $P$ and $d$,
so it has zero derivative:
\begin{equation}
\label{eq:witness_target_derivative}
\begin{aligned}
\nabla_\theta\wTarget{R}
&=\nabla_\theta\int\!\int
 d(\vx_1,\vx_1')P(\mathrm{d}\vx_1')P(\mathrm{d}\vx_1)
\\
&=0.
\end{aligned}
\end{equation}
Differentiating \eqref{eq:witness_distance_decomposition} and substituting
\eqref{eq:witness_cross_derivative}, \eqref{eq:witness_self_derivative},
and \eqref{eq:witness_target_derivative} yields
\begin{equation}
\label{eq:witness_density_derivative}
\begin{aligned}
\nabla_\theta D_\phi(Q_\theta,P)
&=2\nabla_\theta\wCross{C_\theta}
-\nabla_\theta\wSelf{S_\theta}-\nabla_\theta\wTarget{R}
\\
&=2\int\wCross{a(\hat{\vx}_1)}
\nabla_\theta q_\theta(\hat{\vx}_1)\,\mathrm{d}\hat{\vx}_1
\\
&\quad-2\int\wSelf{b_\theta(\hat{\vx}_1)}
\nabla_\theta q_\theta(\hat{\vx}_1)\,\mathrm{d}\hat{\vx}_1-0
\\
&=2\int[\wCross{a(\hat{\vx}_1)}-\wSelf{b_\theta(\hat{\vx}_1)}]
\nabla_\theta q_\theta(\hat{\vx}_1)\,\mathrm{d}\hat{\vx}_1
\\
&=-2\int[\wSelf{b_\theta(\hat{\vx}_1)}-\wCross{a(\hat{\vx}_1)}]
\nabla_\theta q_\theta(\hat{\vx}_1)\,\mathrm{d}\hat{\vx}_1
\\
&=-2\int w_\theta(\hat{\vx}_1)
\nabla_\theta q_\theta(\hat{\vx}_1)\,\mathrm{d}\hat{\vx}_1.
\end{aligned}
\end{equation}
The last equality uses the witness definition
in \eqref{eq:witness_mean_distances}.

\textbf{Step 5: Convert the density derivative into a policy gradient.}
On the support of the conditional density,
the derivative of the logarithm gives
\begin{equation}
\label{eq:witness_log_derivative}
\begin{aligned}
\nabla_\theta\log\pi_\theta(\hat{\vx}_1\mid\vu,c)
&=\frac{\nabla_\theta\pi_\theta(\hat{\vx}_1\mid\vu,c)}
{\pi_\theta(\hat{\vx}_1\mid\vu,c)},
\\
\nabla_\theta\pi_\theta(\hat{\vx}_1\mid\vu,c)
&=\pi_\theta(\hat{\vx}_1\mid\vu,c)
\nabla_\theta\log\pi_\theta(\hat{\vx}_1\mid\vu,c).
\end{aligned}
\end{equation}
These identities are used almost everywhere under the conditional law.
Since $P_0$ is fixed, differentiating
\eqref{eq:witness_marginal_density} gives
\begin{equation}
\label{eq:witness_marginal_derivative}
\begin{aligned}
\nabla_\theta q_\theta(\hat{\vx}_1)
&=\nabla_\theta\int\pi_\theta(\hat{\vx}_1\mid\vu,c)P_0(\mathrm{d}\vu)
\\
&=\int\nabla_\theta\pi_\theta(\hat{\vx}_1\mid\vu,c)P_0(\mathrm{d}\vu)
\\
&=\int\pi_\theta(\hat{\vx}_1\mid\vu,c)
\nabla_\theta\log\pi_\theta(\hat{\vx}_1\mid\vu,c)P_0(\mathrm{d}\vu).
\end{aligned}
\end{equation}
Substitute this into \eqref{eq:witness_density_derivative},
exchange the integrals over $\hat{\vx}_1$ and $\vu$,
and recognize the expectation under the sampling policy:
\begin{equation}
\label{eq:witness_score_identity}
\begin{aligned}
\nabla_\theta D_\phi(Q_\theta,P)
&=-2\int w_\theta(\hat{\vx}_1)
\nabla_\theta q_\theta(\hat{\vx}_1)\,\mathrm{d}\hat{\vx}_1
\\
&=-2\int w_\theta(\hat{\vx}_1)\left[
\int\pi_\theta(\hat{\vx}_1\mid\vu,c)
\nabla_\theta\log\pi_\theta(\hat{\vx}_1\mid\vu,c)
P_0(\mathrm{d}\vu)\right]\mathrm{d}\hat{\vx}_1
\\
&=-2\int P_0(\mathrm{d}\vu)\left[
\int w_\theta(\hat{\vx}_1)\pi_\theta(\hat{\vx}_1\mid\vu,c)
\nabla_\theta\log\pi_\theta(\hat{\vx}_1\mid\vu,c)
\,\mathrm{d}\hat{\vx}_1\right]
\\
&=-2\mathbb{E}_{\vu\sim P_0}
\mathbb{E}_{\hat{\vx}_1\sim\pi_\theta(\cdot\mid\vu,c)}
\left[w_\theta(\hat{\vx}_1)
\nabla_\theta\log\pi_\theta(\hat{\vx}_1\mid\vu,c)\right]
\\
&=-2g_\phi(\theta).
\end{aligned}
\end{equation}
Dividing both sides by $-2$ proves
\eqref{eq:witness_policy_gradient_restated}.

Finally, whenever $g_\phi(\theta)\ne0$,
\begin{equation}
\label{eq:witness_descent_direction}
\nabla_\theta D_\phi(Q_\theta,P)^\top g_\phi(\theta)
=-\tfrac12\|\nabla_\theta D_\phi(Q_\theta,P)\|_2^2<0,
\end{equation}
so $g_\phi(\theta)$ is a descent direction for $D_\phi$.
\end{proof}
\endgroup

\begin{corollary}[Witness gap and distributional agreement]
\label{cor:witness_gap}
For any $P,Q$ with finite first feature moments,
the witness $w=w^\phi_{Q,P}$ satisfies
\begin{equation}
\label{eq:witness_gap}
\mathbb{E}_{P}w-\mathbb{E}_{Q}w=D_\phi(Q,P)\geq0.
\end{equation}
Equality holds exactly when $Q_\phi=P_\phi$,
where $Q_\phi$ and $P_\phi$ are the corresponding feature distributions.
\end{corollary}

\begin{proof}
Expanding both witness expectations and using symmetry yields
\begin{equation}
\begin{aligned}
\mathbb{E}_{P}w-\mathbb{E}_{Q}w
&=2\mathbb{E}_{Q\otimes P}d_\phi
-\mathbb{E}_{Q\otimes Q}d_\phi
-\mathbb{E}_{P\otimes P}d_\phi
\\
&=D_\phi(Q,P).
\end{aligned}
\end{equation}
Finite feature moments ensure all expectations are finite.
Since $D_\phi(Q,P)=\operatorname{ED}(Q_\phi,P_\phi)$,
Euclidean energy distance is nonnegative and vanishes
exactly when these distributions agree~\citep{szekely2013energy}.
\end{proof}

\clearpage
\section{Flow-matching Mismatch and Recovery by Witness Post-training}
\label{app:fm_counterexample}

We prove Proposition~\ref{prop:fm_counterexample} using the two-dimensional Gaussian example
shown in Figure~\ref{fig:fm_counterexample}.
Both training stages optimize the same scalar parameter.
Let $Z\sim\mathcal{N}(0,I_2)$ and consider
\begin{equation}
v_b(x,t)=-bx,\qquad b\geq0.
\end{equation}
Its exact time-one map and endpoint distribution are
\begin{equation}
T_b(z)=e^{-b}z,\qquad Q_b=\mathcal{N}(0,e^{-2b}I_2).
\end{equation}
Set $\tau=e^{-1}$ and $P=Q_1=\mathcal{N}(0,\tau^2I_2)$.
The target is therefore exactly representable at $b=1$.
All endpoint distributions have the correct mean, zero.
Let $X_1\sim P$ and $t\sim\mathcal{U}[0,1]$
be independent of each other and of $Z$.
Training uses $X_t=(1-t)Z+tX_1$ and target velocity $X_1-Z$.
\textcolor{blue}{The velocity family is misspecified:
the marginal velocity that flow matching regresses onto is}
{\color{blue}\begin{equation}
u^\star(x,t)=\mathbb{E}[X_1-Z\mid X_t=x]=k(t)\,x,
\qquad
k(t)=\frac{t\tau^2-(1-t)}{(1-t)^2+t^2\tau^2},
\end{equation}}
\textcolor{blue}{whose coefficient varies from $k(0)=-1$ to $k(1)=1$,
so $u^\star\neq v_b$ for every $b$.}

\begin{proof}
\textbf{The population FM minimizer misses the target.}
The velocity-regression residual is
\begin{equation}
-bX_t-(X_1-Z)=[1-b(1-t)]Z-(1+bt)X_1.
\end{equation}
Independence and zero means eliminate the cross term.
Using $\mathbb{E}\|Z\|^2=2$ and $\mathbb{E}\|X_1\|^2=2\tau^2$ gives
\begin{equation}
\label{eq:counterexample_fm}
\begin{aligned}
L_{\mathrm{FM}}(b)
&=2\int_0^1
\left([1-b(1-t)]^2+\tau^2(1+bt)^2\right)\,\mathrm{d}t\\
&=\frac{2(1+\tau^2)}{3}b^2
-2(1-\tau^2)b+2(1+\tau^2).
\end{aligned}
\end{equation}
This quadratic is strictly convex.
Its unique minimizer is positive and therefore feasible:
\begin{equation}
\label{eq:counterexample_fm_minimizer}
b_{\mathrm{FM}}^\star=\frac{3(1-e^{-2})}{2(1+e^{-2})}
=\frac32\tanh(1)\approx1.142391.
\end{equation}
Since $e^2>5$, this minimizer exceeds $1$.
Consequently, $Q_{b_{\mathrm{FM}}^\star}\neq P$:
its mean is correct, but its variance is too small.
The mismatch holds under population training and exact flow integration.

\textbf{Closed-form terminal energy distance.}
Write $s=e^{-b}$ and $c_2=\mathbb{E}\|Z\|=\sqrt{\pi/2}$.
For independent $\hat{X}_1,\hat{X}_1'\sim Q_b$ and $X_1,X_1'\sim P$,
\begin{equation}
\begin{aligned}
\hat{X}_1-X_1&\sim\mathcal{N}(0,(s^2+\tau^2)I_2),\\
\hat{X}_1-\hat{X}_1'&\sim\mathcal{N}(0,2s^2I_2),\\
X_1-X_1'&\sim\mathcal{N}(0,2\tau^2I_2).
\end{aligned}
\end{equation}
Thus the Euclidean energy distance in \eqref{eq:energy_distance} is
\begin{equation}
\label{eq:counterexample_energy}
D(b):=D_{\mathrm{id}}(Q_b,P)
=c_2\left[2\sqrt{s^2+\tau^2}-\sqrt{2}(s+\tau)\right].
\end{equation}
The inequality $2(s^2+\tau^2)\geq(s+\tau)^2$
shows that $D(b)\geq0$, with equality exactly when $s=\tau$.
Therefore $b=1$ uniquely minimizes terminal energy distance.

\textbf{The witness gradient points toward the correct parameter.}
At each iteration, form the current energy witness
\begin{equation}
w_b(x)=\mathbb{E}_{\hat{X}_1'\sim Q_b}\|x-\hat{X}_1'\|
-\mathbb{E}_{X_1\sim P}\|x-X_1\|.
\end{equation}
Hold this witness fixed while differentiating the sampling policy.
The Gaussian density has common support $\mathbb{R}^2$, with score
\begin{equation}
\partial_b\log q_b(x)=2-e^{2b}\|x\|^2.
\end{equation}
On compact parameter intervals, Gaussian tails justify differentiation under expectations.
\textcolor{blue}{Proposition}~\ref{thm:witness}, with identity features and no conditioning, gives
\begin{equation}
\label{eq:counterexample_witness_gradient}
\begin{aligned}
G(b)&:=\mathbb{E}_{\hat{X}_1\sim Q_b}
\left[w_b(\hat{X}_1)\partial_b\log q_b(\hat{X}_1)\right]\\
&=-\frac12D'(b)
=c_2s\left[\frac{s}{\sqrt{s^2+\tau^2}}-\frac1{\sqrt2}\right].
\end{aligned}
\end{equation}
In particular,
\begin{equation}
\label{eq:counterexample_gradient_sign}
G(b)>0\ (b<1),\qquad G(1)=0,\qquad G(b)<0\ (b>1).
\end{equation}
There are no other stationary points at finite $b\geq0$.
The continuous parameter flow $\dot b=G(b)$
stays between its initialization and $1$ and converges to $1$.
Indeed, the trajectory is monotone and bounded;
any limit different from $1$ would have nonzero derivative.

\textbf{Discrete parameter updates also converge.}
We give an explicit sufficient constant step size.
Let $u=s/\sqrt{s^2+\tau^2}\in[0,1]$.
Differentiation of \eqref{eq:counterexample_witness_gradient} yields
\begin{equation}
G'(b)=-c_2s\left(2u-u^3-\frac1{\sqrt2}\right).
\end{equation}
Since $0\leq2u-u^3\leq\sqrt2$ and $s\leq1$,
\begin{equation}
|G'(b)|\leq\frac{c_2}{\sqrt2}
=\frac{\sqrt\pi}{2}=:L,\qquad b\geq0.
\end{equation}
For any finite initialization $b_0\geq0$ different from $1$,
let $I=[\min\{b_0,1\},\max\{b_0,1\}]$.
Define
\begin{equation}
k(b)=\begin{cases}
G(b)/(1-b),&b\neq1,\\
-G'(1)=c_2\tau/(2\sqrt2),&b=1.
\end{cases}
\end{equation}
This function is continuous and strictly positive on $I$.
The derivative bound and $G(1)=0$ imply $k(b)\leq L$.
Compactness gives $k_{\min}:=\min_{b\in I}k(b)>0$.
For any fixed $0<\eta\leq1/L=2/\sqrt\pi$, consider
\begin{equation}
\label{eq:counterexample_post_training}
b_{n+1}=b_n+\eta G(b_n).
\end{equation}
Then
\begin{equation}
b_{n+1}-1=[1-\eta k(b_n)](b_n-1).
\end{equation}
Each iterate lies between the previous iterate and $1$.
Inductively, all iterates remain in $I$, and
\begin{equation}
|b_n-1|\leq(1-\eta k_{\min})^n|b_0-1|\longrightarrow0.
\end{equation}
Thus $Q_{b_n}\to P$ and $D(b_n)\to0$.
Taking $b_0=b_{\mathrm{FM}}^\star$ proves the post-training claim.
Every update stays within the original parameterized model family.
\end{proof}

\paragraph{The mismatch also holds for unweighted data-prediction loss.}
For the same velocity family, define the clean-data prediction
\begin{equation}
\widehat X_{1,b}(x,t)=x+(1-t)v_b(x,t)=[1-b(1-t)]x.
\end{equation}
This is a denoising prediction, not an integrated ODE endpoint.
Because $X_1=X_t+(1-t)(X_1-Z)$, its residual satisfies
\begin{equation}
\widehat X_{1,b}(X_t,t)-X_1
=(1-t)[v_b(X_t,t)-(X_1-Z)].
\end{equation}
With uniform time sampling, the unweighted data loss becomes
\begin{equation}
\label{eq:counterexample_x_loss}
\begin{aligned}
L_x(b)&=\mathbb{E}\|\widehat X_{1,b}(X_t,t)-X_1\|^2\\
&=\frac{6+\tau^2}{15}b^2
-\frac{3-\tau^2}{3}b+\frac{2(1+\tau^2)}3.
\end{aligned}
\end{equation}
Its unique minimizer is
\begin{equation}
b_x^\star=\frac{5(3-e^{-2})}{2(6+e^{-2})}
\approx1.167281>1.
\end{equation}
The strict inequality follows from $e^2>7/3$.
The same witness updates converge from $b_x^\star$ to $1$.

\paragraph{Figure reproduction.}
Figure~\ref{fig:fm_counterexample} uses \eqref{eq:counterexample_post_training} with $\eta=1$
for $100$ updates from $b_{\mathrm{FM}}^\star$.
All three panels display exact Gaussian densities
on a shared grid with a common linear color scale.
The final parameter differs from $1$ by less than $10^{-8}$.
Reported variance ratios are $e^{-2b}/e^{-2}$ for either coordinate.
Training updates use exact population gradients.

\clearpage
\section{Algorithm Details}
\label{app:algorithm}

Algorithm~\ref{alg:perturbexprrl} summarizes the training procedure.
The implicit branches in equation~\ref{eq:implicit_pos}
express reward-guided updates through a single trainable velocity field.
The data-collection policy is frozen during each iteration
and synchronized with the current model at its end.
The pretrained reference remains fixed throughout training.
We separately maintain an exponential moving average (EMA)
for checkpointing and evaluation.

Each batch fixes a perturbation $c$ and independently samples
$K$ controls $\vu_g\sim P_0$, with $m$ rollouts per control.
The rollouts from each sampled control form a sampling group $g$.
We index generated endpoints by $i=(g,j)$.
Equation~\ref{eq:nft_loss} maps their advantages $A_i$
to preference weights $r_i$ before optimization,
as shown in Algorithm~\ref{alg:perturbexprrl}.
Training noise is sampled independently of the rollout starts.
All velocities in \eqref{eq:nft_loss} are evaluated
at the same state $\vx_{t_i}$, time $t_i$,
and conditioning $(\vu_g,c)$.
The expectation covers batches, endpoints, training noise, and times.

\begin{algorithm}[t]
\caption{\emph{\method{}}: energy-witness RL for scDFM}
\label{alg:perturbexprrl}
\begin{algorithmic}[1]
\Require Pretrained parameters $\theta_0$; training perturbations;
control distribution $P_0$; $K$ groups with $m$ rollouts each;
reward weights; $\gamma$, $C$, $\lambda_{\mathrm{ref}}$
\State Initialize $\theta,\theta^{\mathrm{old}},\theta^{\mathrm{EMA}}\leftarrow\theta_0$;
keep $\theta_0$ fixed
\For{each RL iteration}
    \State Sample $c\sim\mathrm{Unif}(\mathcal{C}_{\mathrm{train}})$
    and $\{\vu_g\}_{g=1}^K\sim P_0^{\otimes K}$
    \For{each group $g$ and rollout $j$}
        \State Draw rollout noise $\vx^{\mathrm{roll}}_{0,g,j}\sim\mathcal{N}(\vzero,\mI)$;
        integrate $\vv^{\mathrm{old}}$ conditioned on $(\vu_g,c)$
        to generate $\hat{\vx}_{1,g,j}$
    \EndFor
    \State Score the complete batch at $\max(\hat{\vx}_{1,g,j},0)$;
    compute $R_i$ using \eqref{eq:reward_combination}
    and standardize within groups to obtain $A_i=z(R)_i$
    \State Set $r_i=\tfrac12+\operatorname{clip}(A_i,-C,C)/(2C)$;
    detach endpoints and weights
    \State Shuffle endpoints and split into training microbatches;
    zero accumulated gradients
    \For{each microbatch}
        \State Independently draw fresh training noise
        $\vx_{0,i}\sim\mathcal{N}(\vzero,\mI)$
        and $t_i\sim\mathcal{U}[0,1]$ for each endpoint
        \State Construct $\vx_{t_i}=(1-t_i)\vx_{0,i}+t_i\hat{\vx}_{1,i}$
        and $\vv_i=\hat{\vx}_{1,i}-\vx_{0,i}$
        \State Compute \eqref{eq:nft_implementation_loss};
        accumulate gradients with respect to $\theta$ only
        \If{the accumulation window is complete}
            \State Clip gradients; update $\theta$;
            zero gradients; update $\theta^{\mathrm{EMA}}$
        \EndIf
    \EndFor
    \State Synchronize $\theta^{\mathrm{old}}\leftarrow\theta$
\EndFor
\State \Return EMA checkpoint for evaluation
\end{algorithmic}
\end{algorithm}

\paragraph{Empirical witness estimation.}
\label{app:empirical_witness}
Let $g(i)$ identify the sampling group of generated cell $\hat{\vx}_{1,i}$.
Each group corresponds to one independent control sampling event.
Repeated draws of the same real cell remain separate groups.
To estimate the expectation over $Q_{\theta,c}$,
we use $\mathcal{I}_i=\{k:c_k=c,\ g(k)\ne g(i)\}$.
This excludes $\hat{\vx}_{1,i}$ and rollouts sharing its sampled control,
so reference groups use independently sampled controls.
Given $M_c$ real responses $\{\vx_{1,j}\}_{j=1}^{M_c}$ from $P_c$,
the empirical witness is
\begin{equation}
\label{eq:empirical_witness}
\widehat{w}^\phi_i
=\frac{1}{|\mathcal{I}_i|}
\sum_{k\in\mathcal{I}_i}d_\phi(\hat{\vx}_{1,i},\hat{\vx}_{1,k})
-\frac{1}{M_c}\sum_{j=1}^{M_c}d_\phi(\hat{\vx}_{1,i},\vx_{1,j}).
\end{equation}

\paragraph{Implementation.}
Rewards use nonnegative-clipped endpoints, while the NFT target
uses the original, unclipped model endpoint.
Generated endpoints, advantages, and old/reference predictions are detached.
We use bf16 mixed precision for the model
and combine NFT loss operands in FP32.
Each endpoint receives one fresh noise draw and one training time.
The operator $z$ subtracts the within-group mean
and divides by the within-group standard deviation,
with variance computed using divisor $m$.
For joint training, the implementation raises an error
if the gene-expression witness or combined reward
has zero within-group standard deviation;
constant single-cell rewards are permitted.
Raw witness credits are not clipped;
clipping applies to the final advantages before NFT weighting.
Specifically, $\widetilde{A}_i=\operatorname{clip}(A_i,-C,C)/C$
in \eqref{eq:nft_loss}, where $C>0$ is the clipping threshold.
The implementation averages squared errors over $d$ velocity features
and multiplies the policy loss by $C$:
\begin{equation}
\label{eq:nft_implementation_loss}
\mathcal{L}_{\mathrm{impl}}(\theta)
=\mathbb{E}\Big[
\frac{C}{d}\big(r_i\|\vv_\theta^+-\vv\|_2^2
+(1-r_i)\|\vv_\theta^--\vv\|_2^2\big)
+\frac{\lambda_{\mathrm{ref}}}{d}
\|\vv_\theta-\vv_{\theta_0}\|_2^2\Big].
\end{equation}
The effective main-text weight is
$\lambda_{\mathrm{reg}}=\lambda_{\mathrm{ref}}/C$,
so $\mathcal{L}_{\mathrm{impl}}=(C/d)\mathcal{L}$.
Thus, \eqref{eq:nft_loss} preserves the relative policy and reference weights
while omitting a positive global scale.
Table~\ref{tab:nft_settings} reports the implementation weight $\lambda_{\mathrm{ref}}$.
Velocity regularization implements the proximity goal
in \eqref{eq:post_training_objective}.

\paragraph{Training settings.}
Table~\ref{tab:nft_settings} lists setting-specific parameters.
All runs use $200$ optimizer updates,
a learning rate of $5\times10^{-7}$,
$\gamma=1$, $C=2$,
and a maximum gradient norm of $0.25$.
Each iteration collects one sampling batch and trains
for one inner epoch, producing one optimizer update.
Rollouts use RK4 with $20$ time points ($19$ intervals).
Both the global seed and training-noise seed are $0$.
The old-policy decay is $0$, corresponding to direct synchronization.
The evaluation EMA uses decay
$\min\{(1+s)/(10+s),0.9\}$ at optimizer step $s$,
following the implementation's initial decay schedule. \textbf{Setting-specific hyperparameters in Table 3 were selected on a validation set of 5\% of the training data; test conditions were used only for final evaluation.}

\begin{table}[t]
\centering
\caption{NFT post-training settings.}
\label{tab:nft_settings}
\begin{tabular}{@{}lccc@{}}
\toprule
Parameter & Additive & Holdout & ComboSciPlex \\
\midrule
$\lambda_{\mathrm{sup}},\lambda_{\mathrm{path}}$
& $3,1$ & $1,3$ & $3,1$ \\
Groups $\times$ rollouts
& $8\times8$ & $16\times8$ & $8\times8$ \\
Microbatch / accumulation
& $32\,/\,2$ & $32\,/\,4$ & $32\,/\,2$ \\
Reference weight $\lambda_{\mathrm{ref}}$ & $2$ & $2$ & $1$ \\
Training $\alpha_{\mathrm{path}}$ & $0.05$ & $0.20$ & $0.05$ \\
\bottomrule
\end{tabular}
\end{table}

The standardized gene-expression witness has unit weight in all settings.
Support uses pooled calibration with $\alpha_{\mathrm{sup}}=0.05$.
Fixed reward evaluation uses $\alpha_{\mathrm{path}}=0.05$ for all settings,
including holdout, whose training threshold is $0.20$.
The witness excludes the query's entire sampling group;
additional mean and detection energy terms have zero weight.

\begingroup\color{blue}
\paragraph{Computational cost.}
\label{app:compute_cost}
Post-training adds no inference cost,
because the post-trained model keeps the architecture and sampler of scDFM.
Each iteration generates $N=Km$ cells with $T=20$ RK4 time points,
which costs $4(T-1)=76$ forward passes per cell without gradients.
The NFT update then uses one forward--backward pass per cell at a single time $t$,
plus two forward passes without gradients for the old and reference velocities,
so the gradient does not pass through the sampler.
The rewards require no network evaluations.
Each iteration therefore costs $O(NT)$ network evaluations.
Table~\ref{tab:compute_cost} reports the measured wall-clock time on one H100 GPU,
excluding periodic evaluation.
Post-training uses $200$ optimizer updates in total,
i.e., $0.75$--$1.5$ GPU-hours per setting.
\par\endgroup

\begin{table}[t]
\centering
\color{blue}
\caption{Post-training wall-clock time on one H100 GPU.}
\label{tab:compute_cost}
\begin{tabular}{@{}lccc@{}}
\toprule
Setting & Cells / iteration & Time / iteration & Total (200 iterations) \\
\midrule
Additive & $64$ & $14.2$\,s & $47$\,min \\
Holdout & $128$ & $\approx26$\,s & $\approx1.5$\,h \\
ComboSciPlex & $64$ & $13.4$\,s & $45$\,min \\
\bottomrule
\end{tabular}
\end{table}

\clearpage
\section{Single-Cell Reward Calibration}
\label{app:single_cell_calibration}

We separate scorer fitting from calibration following
the sample-splitting principle of conformal
outlier detection~\citep{angelopoulos2023conformal}.
For each scorer, we shuffle real cells within each condition
using seed $42$ and reserve the final half for calibration.
The first $\lfloor n_c/2\rfloor$ cells form the reference subset,
where $n_c$ is the condition's real-cell count.
The two scorers construct their splits separately.
Post-training scorers use training data only;
calibration cells are held out from scorer fitting,
but remain available as real targets for the population witness.
Once fitted and calibrated, both scorers remain fixed during generator post-training.

For expression support, each reference subset determines its condition mean.
Gene scales are sample standard deviations across pooled reference cells,
using divisor $n_{\mathrm{ref}}-1$.
Zero-variance genes are excluded, and calibration scores are pooled
across conditions with one entry per cell.
Specifically, we evaluate \eqref{eq:support_nonconformity}
for each calibration cell under its own condition.
Let $b_1,\ldots,b_{n_{\mathrm{sup}}}$ denote these scores,
where $n_{\mathrm{sup}}$ is the total calibration-cell count across conditions.
The upper-tail rank used in \eqref{eq:support_reward} is
\begin{equation}
\label{eq:support_calibration}
p_{\mathrm{sup}}(\vx,c)=
\frac{1+\sum_{j=1}^{n_{\mathrm{sup}}}
\mathbf{1}\{b_j\geq a_{\mathrm{sup}}(\vx,c)\}}{n_{\mathrm{sup}}+1}.
\end{equation}
A small $p_{\mathrm{sup}}$ identifies cells outside the calibrated radial envelope.
Pooling provides a shared reference across conditions.

For pathway discrimination, we subsample the target-reference subset
when necessary to match the available number of controls.
Each classifier uses equal numbers of target and control cells.
For \eqref{eq:pathway_activity}, the pathway means $m_{c,p}$
and sample standard deviations $s_{c,p}$ are fitted
on these combined reference cells; zero-variance pathways are excluded.
Thus, both standardization and pathway selection depend on condition $c$.
We use logistic regression with an $\ell_1$ penalty,
inverse regularization strength $C_{\mathrm{logit}}=1$,
the \texttt{liblinear} solver, seed $42$, and $1000$ maximum iterations.
Only held-out target cells calibrate the resulting logits.
Let $n_{\mathrm{path},c}$ be their number under condition $c$,
and let $t_{c,j}=f_c(\vx^{\mathrm{cal}}_{1,c,j})$
denote the logit of calibration cell $\vx^{\mathrm{cal}}_{1,c,j}$.
The lower-tail rank used in \eqref{eq:discrimination_reward} is
\begin{equation}
\label{eq:discrimination_calibration}
p_{\mathrm{path}}(\vx,c)=
\frac{1+\sum_{j=1}^{n_{\mathrm{path},c}}
\mathbf{1}\{t_{c,j}\leq f_c(\vx)\}}{n_{\mathrm{path},c}+1}.
\end{equation}
A small $p_{\mathrm{path}}$ indicates classifier evidence in the lower tail
of real target responses under the same condition.

Both tail ranks include ties and the finite-sample $+1$ correction.
The thresholds $\alpha_{\mathrm{sup}},\alpha_{\mathrm{path}}$ are specified in Appendix~\ref{app:algorithm}.
These ranks define empirical reward thresholds.
Conformal false-positive guarantees require exchangeability with calibration scores.

\clearpage
\section{Evaluation Metrics}
\label{app:evaluation_metrics}

\paragraph{Distribution-level metrics.}
We compare predicted and real cell populations using three metrics.
Energy distance~\citep{szekely2013energy} measures distributional differences
using Euclidean distances in expression space.
We use the empirical form of \eqref{eq:energy_distance},
including diagonal pairs in the within-population averages.
Maximum mean discrepancy (MMD)~\citep{gretton2012kernel} provides a complementary kernel-based comparison.
We report its unbiased squared estimator with an RBF kernel,
setting $\sigma^2$ to half the median pairwise squared distance
among real target cells for each perturbation.
Sliced Wasserstein-1 distance (SW1)~\citep{bonneel2015sliced} averages exact
one-dimensional Wasserstein-1 distances over 1,024 random unit directions.
We project all 1,000 expression features without additional transformation.
Directions are normalized isotropic Gaussian vectors,
generated in four blocks of 256 using NumPy seeds 0--3.
All methods and folds share these directions.
Each condition uses 128 predicted cells and all real target cells.
Exact quantile integration accommodates unequal population sizes.
We average conditions equally, then average Norman folds equally.
Lower values are better for all three distributional metrics.

\paragraph{Pseudobulk-mean metrics.}
We average expression within each perturbation
and report MAE, two Pearson correlations, DE-Spearman, and DS.
Mean absolute error (MAE) measures differences between predicted and real means.
Pearson $\rho(\Delta)$ correlates these means after subtracting the control mean.
Following Systema~\citep{vinas2025systema}, $\rho(\hat{\Delta})$ instead subtracts
the equally weighted mean of training perturbation centroids,
emphasizing perturbation-specific effects beyond shared responses.
We use Cell-Eval~\citep{adduri2025predicting} for DE-Spearman and discrimination score (DS).
DE-Spearman correlates predicted and real log fold changes
on genes significantly differentially expressed in real data
at a false discovery rate of $0.05$.
DS ranks the correct perturbation using $L_1$ distances between
absolute control-relative pseudobulk effects and reports $1-r/N$,
where $r$ is its zero-based rank among $N$ evaluated perturbations.
Lower MAE is better;
higher correlations and DS are better.

\paragraph{Pathway enrichment concordance.}
Following PertAdapt~\citep{bai2026pertadapt}, we compare GSEA normalized enrichment scores
(NES) across Reactome pathways~\citep{gillespie2022reactome} using Spearman correlation.
For each condition, we subtract the shared real control mean
from predicted and real pseudobulk means of log-normalized expression.
We rank all 1,000 measured genes by these signed effects.
Pathways must contain at least five measured genes;
the eligible pathways and gene background are shared across methods.
We run weighted preranked GSEA with GSEApy 1.3.1,
weight $1$, 1,000 gene-set permutations, and random seed $42$.
For each condition, NES-Sp. correlates predicted and real NES vectors
over all eligible pathways, without filtering by enrichment significance.
We average conditions equally within each setting and fold,
then average the four Norman folds equally.
Higher values indicate better agreement in pathway enrichment rankings.

\clearpage
\section{Dataset Details}
\label{app:dataset_details}

We use the Norman and ComboSciPlex splits from scDFM.
For each split, training conditions are all benchmark
conditions not listed as test conditions. We further hold out 5\% of these training data as validation set.

\paragraph{Norman additive split.}
Each additive fold holds out 37 double-gene perturbations.
All corresponding single-gene perturbations remain in training.
Each fold therefore has 189 train conditions.
Table~\ref{tab:norman_additive_split} lists test conditions.

\begingroup
\setlength{\tabcolsep}{3pt}
\begin{longtable}{@{}cp{0.86\linewidth}@{}}
\caption{
Norman additive held-out double-gene conditions.
}
\label{tab:norman_additive_split}\\
\toprule
Fold & Test conditions \\
\midrule
\endfirsthead
\toprule
Fold & Test conditions \\
\midrule
\endhead
\bottomrule
\endfoot
0 &
AHR+FEV,
BPGM+SAMD1,
CBL+UBASH3A,
CBL+UBASH3B,
CDKN1B+CDKN1A,
CEBPB+CEBPA,
CEBPB+PTPN12,
CEBPE+CEBPA,
CEBPE+KLF1,
CEBPE+RUNX1T1,
CNN1+MAPK1,
CNN1+UBASH3A,
DUSP9+MAPK1,
ETS2+IGDCC3,
ETS2+PRTG,
FEV+ISL2,
FOSB+CEBPB,
FOSB+CEBPE,
FOSB+OSR2,
FOXA3+FOXA1,
FOXA3+HOXB9,
IRF1+SET,
KIF18B+KIF2C,
KLF1+MAP2K6,
LYL1+IER5L,
MAP2K3+IKZF3,
MAP2K3+MAP2K6,
PTPN12+OSR2,
PTPN12+SNAI1,
PTPN12+ZBTB25,
SAMD1+PTPN12,
SET+KLF1,
TGFBR2+ETS2,
UBASH3B+CNN1,
UBASH3B+PTPN12,
UBASH3B+UBASH3A,
ZC3HAV1+HOXC13 \\
1 &
AHR+FEV,
AHR+KLF1,
BCL2L11+BAK1,
BPGM+ZBTB1,
CBL+PTPN9,
CBL+TGFBR2,
CBL+UBASH3A,
CBL+UBASH3B,
CDKN1B+CDKN1A,
CDKN1C+CDKN1A,
CEBPB+CEBPA,
CEBPE+CEBPB,
CEBPE+KLF1,
DUSP9+SNAI1,
ETS2+IKZF3,
ETS2+MAP7D1,
FOSB+CEBPE,
FOXA3+FOXF1,
FOXA3+FOXL2,
FOXA3+HOXB9,
IGDCC3+PRTG,
KIF18B+KIF2C,
KLF1+CEBPA,
KLF1+CLDN6,
MAP2K3+IKZF3,
MAP2K3+MAP2K6,
MAP2K6+SPI1,
MAPK1+PRTG,
PTPN12+PTPN9,
SAMD1+UBASH3B,
SET+CEBPE,
SGK1+S1PR2,
SGK1+TBX2,
TGFBR2+IGDCC3,
UBASH3B+CNN1,
UBASH3B+OSR2,
ZC3HAV1+CEBPE \\
2 &
BPGM+ZBTB1,
CBL+CNN1,
CBL+PTPN12,
CBL+TGFBR2,
CBL+UBASH3B,
CEBPE+CEBPA,
CEBPE+RUNX1T1,
CNN1+MAPK1,
CNN1+UBASH3A,
DUSP9+KLF1,
DUSP9+SNAI1,
ETS2+MAP7D1,
ETS2+MAPK1,
FEV+CBFA2T3,
FOSB+IKZF3,
FOSB+OSR2,
FOSB+PTPN12,
FOXA1+HOXB9,
FOXL2+MEIS1,
IGDCC3+PRTG,
JUN+CEBPA,
KLF1+CEBPA,
LYL1+IER5L,
MAP2K3+IKZF3,
MAP2K3+MAP2K6,
MAP2K6+IKZF3,
MAP2K6+SPI1,
MAPK1+IKZF3,
PTPN12+PTPN9,
SAMD1+UBASH3B,
TGFBR2+ETS2,
UBASH3B+CNN1,
UBASH3B+PTPN9,
UBASH3B+UBASH3A,
UBASH3B+ZBTB25,
ZC3HAV1+CEBPE,
ZNF318+FOXL2 \\
3 &
AHR+FEV,
AHR+KLF1,
CBL+UBASH3A,
CDKN1C+CDKN1A,
CEBPE+CNN1,
CEBPE+SPI1,
CNN1+MAPK1,
DUSP9+ETS2,
DUSP9+SNAI1,
ETS2+MAP7D1,
ETS2+PRTG,
FEV+CBFA2T3,
FEV+MAP7D1,
FOSB+CEBPB,
FOSB+CEBPE,
FOSB+OSR2,
FOSB+PTPN12,
FOXA1+FOXL2,
IGDCC3+PRTG,
IRF1+SET,
KIF18B+KIF2C,
KLF1+CLDN6,
LYL1+IER5L,
MAPK1+IKZF3,
MAPK1+PRTG,
PTPN12+SNAI1,
PTPN12+UBASH3A,
PTPN12+ZBTB25,
SAMD1+UBASH3B,
SAMD1+ZBTB1,
SET+CEBPE,
SGK1+S1PR2,
SGK1+TBX3,
TGFBR2+ETS2,
UBASH3B+ZBTB25,
ZBTB10+DLX2,
ZBTB10+SNAI1 \\
\end{longtable}
\endgroup

\paragraph{Norman holdout split.}
Each holdout fold contains 15 held-out double-gene conditions.
Their constituent single-gene perturbations are also held out.
The train condition counts are 188, 185, 190, and 188.
Table~\ref{tab:norman_holdout_split} lists test conditions.

\begingroup
\setlength{\tabcolsep}{3pt}
\begin{longtable}{@{}cp{0.41\linewidth}p{0.41\linewidth}@{}}
\caption{
Norman holdout held-out conditions.
}
\label{tab:norman_holdout_split}\\
\toprule
Fold & Single-gene test conditions & Double-gene test conditions \\
\midrule
\endfirsthead
\toprule
Fold & Single-gene test conditions & Double-gene test conditions \\
\midrule
\endhead
\bottomrule
\endfoot
0 &
BPGM,
CEBPA,
CEBPB,
CEBPE,
CNN1,
ETS2,
FEV,
FOSB,
FOXA1,
FOXA3,
HOXB9,
HOXC13,
IER5L,
IGDCC3,
ISL2,
LYL1,
PTPN12,
SAMD1,
SNAI1,
UBASH3A,
UBASH3B,
ZBTB25,
ZC3HAV1 &
BPGM+SAMD1,
CEBPB+PTPN12,
CEBPE+CEBPA,
CNN1+UBASH3A,
ETS2+IGDCC3,
FEV+ISL2,
FOSB+CEBPB,
FOXA3+FOXA1,
FOXA3+HOXB9,
LYL1+IER5L,
PTPN12+SNAI1,
PTPN12+ZBTB25,
SAMD1+PTPN12,
UBASH3B+UBASH3A,
ZC3HAV1+HOXC13 \\
1 &
BAK1,
BCL2L11,
CBL,
CEBPA,
CEBPB,
CEBPE,
CLDN6,
ETS2,
FOSB,
IGDCC3,
IKZF3,
KIF18B,
KIF2C,
KLF1,
MAP2K3,
MAP2K6,
MAP7D1,
PRTG,
S1PR2,
SAMD1,
SGK1,
SPI1,
TBX2,
TGFBR2,
UBASH3A,
UBASH3B &
BCL2L11+BAK1,
CBL+UBASH3A,
CEBPB+CEBPA,
ETS2+MAP7D1,
FOSB+CEBPE,
IGDCC3+PRTG,
KIF18B+KIF2C,
KLF1+CEBPA,
KLF1+CLDN6,
MAP2K3+IKZF3,
MAP2K6+SPI1,
SAMD1+UBASH3B,
SGK1+S1PR2,
SGK1+TBX2,
TGFBR2+IGDCC3 \\
2 &
CBL,
CEBPA,
CEBPE,
CNN1,
DUSP9,
FOSB,
FOXA1,
HOXB9,
IER5L,
JUN,
KLF1,
LYL1,
MAP2K3,
MAP2K6,
MAPK1,
PTPN12,
RUNX1T1,
SNAI1,
SPI1,
TGFBR2,
UBASH3B &
CBL+CNN1,
CBL+PTPN12,
CBL+TGFBR2,
CEBPE+CEBPA,
CEBPE+RUNX1T1,
CNN1+MAPK1,
DUSP9+SNAI1,
FOSB+PTPN12,
FOXA1+HOXB9,
JUN+CEBPA,
KLF1+CEBPA,
LYL1+IER5L,
MAP2K3+MAP2K6,
MAP2K6+SPI1,
UBASH3B+CNN1 \\
3 &
AHR,
CEBPE,
CLDN6,
CNN1,
DLX2,
ETS2,
FEV,
IER5L,
IGDCC3,
IKZF3,
KLF1,
LYL1,
MAP7D1,
MAPK1,
PRTG,
PTPN12,
S1PR2,
SGK1,
SNAI1,
SPI1,
TBX3,
UBASH3A,
ZBTB10 &
AHR+FEV,
CEBPE+CNN1,
CEBPE+SPI1,
CNN1+MAPK1,
ETS2+PRTG,
FEV+MAP7D1,
IGDCC3+PRTG,
KLF1+CLDN6,
LYL1+IER5L,
MAPK1+IKZF3,
PTPN12+UBASH3A,
SGK1+S1PR2,
SGK1+TBX3,
ZBTB10+DLX2,
ZBTB10+SNAI1 \\
\end{longtable}
\endgroup

\paragraph{ComboSciPlex split.}
ComboSciPlex uses the default scDFM split.
The seven held-out conditions are listed in
Table~\ref{tab:combosciplex_split}.
All other ComboSciPlex conditions are used for training.
This gives 25 train conditions.

\begin{table}[h!]
\centering
\caption{
ComboSciPlex held-out test conditions.
}
\label{tab:combosciplex_split}
\setlength{\tabcolsep}{12pt}
\begin{tabular}{@{}ll@{}}
\toprule
Condition 1 & Condition 2 \\
\midrule
Panobinostat & Crizotinib \\
Panobinostat & Curcumin \\
Panobinostat & SRT1720 \\
Panobinostat & Sorafenib \\
SRT2104 & Alvespimycin \\
control & Alvespimycin \\
control & Dacinostat \\
\bottomrule
\end{tabular}
\end{table}

\clearpage
{\color{blue}\section{Post-Training scLDM}
\label{app:scldm}

To test whether our post-training procedure can generalize to other single-cell flow matching models,
we apply it to scLDM~\citep{palla2025scalable},
a single-cell generator trained with flow matching in a learned latent space.
We keep scLDM's genetic-perturbation benchmark on Replogle, splits,
released checkpoints, sampler, and default guidance weight,
change only the post-training step,
and evaluate with the metrics of Table~\ref{tab:pop_all_settings}.

\paragraph{Base model.}
scLDM encodes each cell into $128$ latent tokens of dimension $16$
with a variational autoencoder (VAE).
A diffusion transformer ($12$ layers, width $512$)
predicts the velocity of a linear interpolation path in this latent space,
conditioned on a condition token,
and a negative binomial (NB) decoder maps latents back to counts.
scLDM's default guidance weight on Replogle is $1$,
which reduces classifier-free guidance to the conditional velocity;
we use it for both rollouts and evaluation.

\paragraph{Replogle.}
The data combine Perturb-seq screens in four cell lines
(K562, RPE1, Jurkat, and HepG2)~\citep{replogle2022mapping},
filtered as in State~\citep{adduri2025predicting}.
Each condition is a pair of cell line and targeted gene.
Following scLDM's released metadata,
we use its $2{,}000$ highly variable genes
and hold out $372$ gene edits in HepG2 for testing.
The edits of the other three cell lines and the remaining HepG2 edits are used for training,
so every test edit is observed in at least one other cell line but never in HepG2.
Since scLDM does not release its processed files,
we rebuild them from the public raw counts
($616{,}865$ training and $26{,}548$ test cells).
With scLDM's own evaluation protocol,
the released checkpoints reproduce all six Replogle values reported by
\citet{palla2025scalable} within $1.1\%$.

\paragraph{Rewards.}
All rewards use $\log(1+x)$ of counts normalized to $10^4$ per cell,
restricted to the $2{,}000$ highly variable genes.
Rollout latents are decoded with the frozen VAE,
and each reward of a latent is averaged over $4$ independent NB decodes to reduce decoder noise.
Each condition forms one sampling group.
Because scLDM has no control-cell input, rollouts within a group are independent given $c$,
so the generated reference set in \eqref{eq:empirical_witness}
contains all other rollouts of the same condition.
We combine all three rewards with \eqref{eq:reward_combination}.
The support and pathway rewards follow \eqref{eq:support_reward} and \eqref{eq:discrimination_reward},
built from the training cells of each condition,
split at random into reference and calibration halves,
and computed within each cell line,
with the non-targeting cells of that cell line as the control population.
To isolate each reward, we also train three single-reward ablations
with the same settings and training seed,
keeping only the witness, support, or pathway reward.

\paragraph{Optimization and validation.}
We apply the NFT objective in \eqref{eq:nft_implementation_loss}
to the transformer's latent velocity, with the VAE frozen.
scLDM is trained with label dropout for classifier-free guidance.
We disable it during post-training,
so the current, data-collection, and reference velocities in \eqref{eq:nft_implementation_loss}
are all conditioned on the condition token.
Table~\ref{tab:scldm_settings} lists the settings.
Test cells are used only for the final evaluation.
The validation set consists of the $58$ HepG2 training edits
that scLDM's released size-factor statistics mark as absent from its own training data,
and all of their cells are excluded from the rewards.
On these edits, DS declines after about $130$ iterations while the distribution metrics keep improving,
so we report the model at a fixed budget of $120$ iterations.

\begin{table}[htbp]
\color{blue}
\centering
\caption{NFT post-training settings for scLDM.}
\label{tab:scldm_settings}
\small
\begin{tabular}{@{}ll@{}}
\toprule
Parameter & Value \\
\midrule
Training rewards & $r_{\mathrm{wit}}$, $r_{\mathrm{sup}}$, $r_{\mathrm{path}}$ \\
$\lambda_{\mathrm{sup}}$, $\lambda_{\mathrm{path}}$; $\alpha_{\mathrm{sup}}$, $\alpha_{\mathrm{path}}$
& $6$, $12$; $0.05$, $0.2$ \\
Groups $\times$ rollouts per iteration & $16\times256$\textsuperscript{a} \\
Rollout sampler & Euler, $20$ steps \\
NB decodes per rollout for the reward & $4$ \\
Training draws of $(t,\vx_0)$ per endpoint & $2$ \\
Microbatch (endpoints), inner epochs & $256$, $1$ \\
Optimizer, learning rate & AdamW, $10^{-5}$ \\
$\gamma$, $C$, $\lambda_{\mathrm{ref}}$ & $1$, $5$, $10^{-2}$ \\
Maximum gradient norm & $1$ \\
Data-collection policy EMA decay (per iteration) & $0.9$ \\
Evaluation EMA decay (per update) & $0.99$ \\
Reported iteration & $120$ (fixed budget) \\
Training seed & $0$ \\
\bottomrule
\end{tabular}

\vspace{2pt}
{\footnotesize \textsuperscript{a}$8$ HepG2 training edits and $8$ conditions drawn from all training conditions
with at least $30$ reference cells ($5{,}333$ of $5{,}391$).}
\end{table}

\paragraph{Evaluation.}
For each test cell, we generate one cell conditioned on its condition,
using scLDM's adaptive-step ODE sampler (dopri5) and the NB decoder,
with $3$ generation seeds.
We compute the metrics of Table~\ref{tab:pop_all_settings}
with the definitions and code of Appendix~\ref{app:evaluation_metrics},
on the same $\log(1+x)$ features of the $2{,}000$ highly variable genes.
Metrics are computed per held-out edit.
The control population is the HepG2 non-targeting training cells,
the center of $\rho(\hat{\Delta})$ is the equally weighted mean of the HepG2 training-edit centroids,
and DS ranks each edit among the $372$ test edits.
NES-Sp.\ uses the $937$ Reactome pathways with at least five genes in the panel.
Each condition has as many generated cells as test cells.
As in Table~\ref{tab:holdout_reward_ablation}, we also report the mean calibrated support and pathway rewards,
with scorers rebuilt from the real test cells of each condition.

\paragraph{Results.}
As shown in Table~\ref{tab:scldm_results},
post-training with all three rewards improves all nine metrics and both single-cell rewards over scLDM,
for example Energy by $11.9\%$ and MMD by $36.6\%$.
Over the $372$ held-out edits, the paired $t$-statistics against scLDM exceed $5$ in absolute value
for every metric except DS ($1.4$).

The single-reward ablations follow the same trend as the scDFM ablations in Section~\ref{sec:ablations}:
each reward mainly improves one category of metrics.
The witness alone gives the best distribution metrics,
the pathway reward alone gives the best NES-Sp., DE-Sp., and $\rho(\Delta)$,
and the support reward alone gives the highest $\mathrm{sup}$.
Combining the three rewards balances these improvements
and is the only variant that improves all nine metrics over scLDM.

\begin{table}[htbp]
\color{blue}
\centering
\caption{
\textbf{Post-training scLDM with each reward.}
Rows give the training rewards; the first row is the released scLDM.
Metrics follow Table~\ref{tab:pop_all_settings}, computed per test condition
($372$ held-out HepG2 edits)
and averaged over conditions and $3$ generation seeds.
$\mathrm{sup}$ and $\mathrm{path}$ are mean calibrated rewards of the generated cells
under scorers rebuilt from the real test cells ($\alpha=0.05$).
Bold indicates best; underline indicates second best.
}
\label{tab:scldm_results}
\renewcommand{\arraystretch}{1.1}
\resizebox{\linewidth}{!}{%
\begin{tabular}{@{}cccrrr@{\hspace{6pt}}r@{\hspace{6pt}}rrrrr@{\hspace{6pt}}rr@{}}
\toprule
\multicolumn{3}{c}{Training rewards}
& \multicolumn{3}{c}{\cellcolor{light-light-gray}\textbf{Distribution}}
& \cellcolor{light-light-gray}\textbf{Pathway}
& \multicolumn{5}{c}{\cellcolor{light-light-gray}\textbf{Pseudobulk-mean}}
& \multicolumn{2}{c}{\cellcolor{light-light-gray}\textbf{Reward}} \\
\cmidrule(lr){1-3}\cmidrule(lr){4-6}\cmidrule(lr){7-7}\cmidrule(lr){8-12}\cmidrule(l){13-14}
$\mathrm{wit}$ & $\mathrm{sup}$ & $\mathrm{path}$
& \textbf{\shortstack{Energy\\$\downarrow$}}
& \textbf{\shortstack{MMD\\$\downarrow$}}
& \textbf{\shortstack{SW1\\$\downarrow$}}
& \textbf{\shortstack{NES-Sp.\\$\uparrow$}}
& \textbf{\shortstack{MAE\\$\downarrow$}}
& \textbf{\shortstack{DE-Sp.\\$\uparrow$}}
& \textbf{\shortstack{$\rho(\hat{\Delta})$\\$\uparrow$}}
& \textbf{\shortstack{$\rho(\Delta)$\\$\uparrow$}}
& \textbf{\shortstack{DS\\$\uparrow$}}
& \textbf{\shortstack{$\mathrm{sup}$\\$\uparrow$}}
& \textbf{\shortstack{$\mathrm{path}$\\$\uparrow$}} \\
\midrule
$\times$ & $\times$ & $\times$
& 2.3945 & 0.0154 & 0.1979 & 0.3706 & 0.1329 & 0.3978 & 0.3167 & 0.3270 & 0.7518 & 0.9331 & 0.9798 \\
\midrule
$\checkmark$ & $\times$ & $\times$
& \textbf{2.0478} & \textbf{0.0086} & \textbf{0.1798} & 0.3833 & \textbf{0.1184} & 0.4084 & \underline{0.3314} & 0.3276 & 0.6948 & 0.9553 & 0.9804 \\
$\times$ & $\checkmark$ & $\times$
& 2.4011 & 0.0149 & 0.1953 & 0.3989 & 0.1366 & 0.4019 & 0.3110 & 0.3398 & \textbf{0.7649} & \textbf{0.9780} & 0.9810 \\
$\times$ & $\times$ & $\checkmark$
& 2.6500 & 0.0195 & 0.2060 & \textbf{0.4498} & 0.1417 & \textbf{0.4626} & 0.2349 & \textbf{0.3979} & \underline{0.7644} & 0.9297 & \textbf{0.9902} \\
\midrule
$\checkmark$ & $\checkmark$ & $\checkmark$
& \underline{2.1088} & \underline{0.0098} & \underline{0.1829} & \underline{0.4030} & \underline{0.1210} & \underline{0.4304} & \textbf{0.3369} & \underline{0.3514} & 0.7617 & \underline{0.9602} & \underline{0.9854} \\
\bottomrule
\end{tabular}%
}
\end{table}
}

\clearpage
{\color{blue}\section{From the Witness Identity to the Implemented Update}
\label{app:transfer}
\providecommand{\vV}{{\bm{V}}}

Proposition~\ref{thm:witness} concerns an ideal score-function update with the exact witness.
The implemented update (Appendix~\ref{app:algorithm}) differs in four ways:
(i) the witness is estimated from finite samples (Eq.~\eqref{eq:empirical_witness}),
(ii) rewards are standardized within each sampling group,
(iii) advantages are clipped, and
(iv) the update follows the NFT surrogate loss (Eq.~\eqref{eq:nft_loss}) rather than a likelihood gradient.
This section gives three results on (iv) and then shows how (i)--(iii) enter them.
\begin{itemize}
\item Proposition~\ref{thm:witness} has an exact counterpart for the NFT loss (Proposition~\ref{prop:nft_pairing}):
pairing the NFT gradient with the gradient of the witness along the sampler gives
the within-group covariance $\kappa$ between the witness and the advantage, for every generator.
\item At a stationary point of the implemented loss, $\kappa$ is bounded by the approximation error of the network
and the pull of the reference term (Proposition~\ref{prop:nft_stationary}).
With the exact witness, $\kappa$ is positive away from the target (Corollary~\ref{cor:transfer}),
so apart from these two terms the update has no stationary point away from the target.
\item A functional NFT step changes $D_\phi$ at first order by $-2\kappa$ plus a residual with an explicit expression (Proposition~\ref{prop:transfer}).
The residual vanishes when the sampler and the training states have the same marginals,
and it is bounded linearly by the generator's self-consistency error.
\end{itemize}
The choices (i)--(iii) enter the leading term $-2\kappa$ only through the scalar $\kappa$ (Corollary~\ref{cor:transfer}):
standardization preserves the sign of each group's contribution,
the finite-sample error preserves it in expectation when the estimation noise is symmetric,
and clipping adds an explicitly bounded term.
We treat the witness reward.
The support and pathway penalties are discussed in a remark at the end of this section.

\paragraph{Setting.}
Fix $c$ and drop it from the notation.
At an iteration, let $\vv^{\mathrm{old}}$ be the synchronized velocity,
$\Phi_{s\to t}(\cdot\mid\vu)$ its ODE flow,
$\pi_{\vu}=\pi_{\mathrm{old}}(\cdot\mid\vu)$ the law of $\Phi_{0\to1}(\vx_0\mid\vu)$ for $\vx_0\sim\mathcal N(\vzero,\mI)$,
$Q=\int\pi_{\vu}\,P_0(\mathrm d\vu)$ the generated population, and $w:=w^\phi_{Q,P}$ the current witness.
A group $g$ draws $\vu_g\sim P_0$ and $m$ rollouts $\hat\vx_{g,1},\dots,\hat\vx_{g,m}$ i.i.d.\ from $\pi_{\vu_g}$.
Each rollout receives an advantage $\widetilde A_{g,j}\in[-1,1]$ that may depend on its own group,
on other groups, and on real cells.
For training, $\vx_0\sim\mathcal N(\vzero,\mI)$ and $t\sim\mathcal U[0,1]$ are drawn independently of everything else
(the controls, all rollouts, and the real cells),
$\vx_t=(1-t)\vx_0+t\hat\vx_1$, and $\vV=\hat\vx_1-\vx_0$.
We use two families of marginals:
$p_t(\cdot\mid\vu)$, the law of the training state $\vx_t$ given $\vu$, and
$\rho_t(\cdot\mid\vu)$, the law of the sampler state $\Phi_{0\to t}(\vx_0\mid\vu)$.
Both equal $\mathcal N(\vzero,\mI)$ at $t=0$ and $\pi_{\vu}$ at $t=1$, but they may differ in between.
Write $\varphi_{\vu}(\hat\vx)=\mathbb E[\widetilde A\mid\hat\vx_1=\hat\vx,\vu]$,
$f_{\vu}=\varphi_{\vu}-\mathbb E[\varphi_{\vu}\mid\vu]$, $\mu_{\vu}=\mathbb E[w(\hat\vx_1)\mid\vu]$, and
\begin{equation}
\label{eq:transfer_kappa_def}
\kappa:=\mathbb E\big[(w(\hat\vx_{g,j})-\mu_{\vu_g})\,\widetilde A_{g,j}\big]
=\mathbb E_{\vu\sim P_0}\!\int w\,f_{\vu}\,\mathrm d\pi_{\vu},
\end{equation}
where the second form follows from the tower property.
Let
\begin{equation}
\label{eq:transfer_value}
\psi_t(\vx\mid\vu):=w\big(\Phi_{t\to1}(\vx\mid\vu)\big)
\end{equation}
be the witness of the cell that the sampler generates from the state $\vx$ at time $t$.
Finally, let $\bar\vv(\vx,t\mid\vu):=\mathbb E[\vV\mid\vx_t=\vx,\vu]$
be the velocity that flow matching would fit on the generator's own rollouts,
and $\ve:=\bar\vv-\vv^{\mathrm{old}}$ the \emph{self-consistency error}.
It vanishes only if the generator is a fixed point of flow matching on its own samples.

\paragraph{Assumptions.}
We use the following smoothness conditions, uniformly in $t\in[0,1]$ and $\vu$.
\begin{itemize}
\item[(S1)] $\vv^{\mathrm{old}}$ and $\nabla_{\vx}\vv^{\mathrm{old}}$ are continuous in $(\vx,t)$,
with $\|\nabla_{\vx}\vv^{\mathrm{old}}\|\le\ell$ and $|\vv^{\mathrm{old}}(\vzero,t\mid\vu)|\le B$.
\item[(S1$'$)] $\nabla_{\vx}\vv^{\mathrm{old}}$ is $M$-Lipschitz in $\vx$.
\item[(S2)] The field $\Delta$ defined below is $H_1$-Lipschitz in $\vx$ and satisfies $|\Delta(\vx,t\mid\vu)|\le H_0(1+|\vx|)$, uniformly up to $t=1$.
\item[(S3)] $w$ is $C^1$ with $|\nabla w|\le L_w$ and $W$-Lipschitz gradient,
and the mean distance $b_{Q'}(\vx)=\mathbb E_{Q'}d_\phi(\vx,\cdot)$ satisfies
$\sup_{\vx}|\nabla b_{Q'}(\vx)-\nabla b_{Q}(\vx)|\le K_b\,W_1(Q',Q)$ for $Q'$ near $Q$.
\end{itemize}
(S2) allows $\Delta$ to grow linearly, as it does when the self-consistency error $\ve$ grows linearly.
Under (S1), the flow is well defined and $C^1$, with $\|\nabla\Phi_{s\to t}\|\le e^{\ell|t-s|}$,
and $\pi_{\vu}$ has finite moments of all orders.
If $\phi$ is $L_\phi$-Lipschitz ($L_\phi=1$ for the implemented $\phi_{\mathrm{wit}}(\vx)=\vx$),
then $d_\phi$ is $L_\phi$-Lipschitz in each argument and $w$ is Lipschitz with constant $L_w\le2L_\phi$.
Propositions~\ref{prop:nft_pairing} and~\ref{prop:nft_stationary} use only (S1) and this Lipschitz property of $w$.
(S1$'$), (S2), and (S3) are used only for the first-order expansion of $D_\phi$
(Propositions~\ref{prop:transfer_first_order} and~\ref{prop:transfer}).

\begin{proposition}[NFT gradient at the synchronized policy]
\label{prop:nft_gradient}
Define $\Delta(\vx,t\mid\vu):=\mathbb E\big[\widetilde A\,(\vV-\vv^{\mathrm{old}})\mid\vx_t=\vx,\vu\big]$.
At $\vv_\theta=\vv^{\mathrm{old}}$, the functional gradient of the policy term of Eq.~\eqref{eq:nft_loss}
in $L^2(p_t)$ is $-2\gamma\Delta$.
A functional gradient step of size $\eta$ is therefore $\vv^{\mathrm{old}}\mapsto\vv^{\mathrm{old}}+\varepsilon\Delta$ with $\varepsilon=2\gamma\eta$.
\end{proposition}
\begin{proof}
With $r=(1+\widetilde A)/2$, differentiate
$r\|\gamma\vv_\theta+(1-\gamma)\vv^{\mathrm{old}}-\vV\|^2+(1-r)\|-\gamma\vv_\theta+(1+\gamma)\vv^{\mathrm{old}}-\vV\|^2$
in $\vv_\theta$ and evaluate at $\vv_\theta=\vv^{\mathrm{old}}$. Both branches reduce to $\vv^{\mathrm{old}}-\vV$, giving
$2\gamma(2r-1)(\vv^{\mathrm{old}}-\vV)=-2\gamma\widetilde A(\vV-\vv^{\mathrm{old}})$.
Condition on $(\vx_t,\vu)$.
\end{proof}
Since $\theta^{\mathrm{old}}\leftarrow\theta$ is synchronized at every iteration
and each iteration performs one optimizer update (Appendix~\ref{app:algorithm}),
the implemented gradient is evaluated exactly at this point.

\paragraph{An exact identity for the NFT loss.}
The next lemma is the fundamental theorem of calculus along the straight training path.
It holds for any velocity field and is the only place where the structure of the NFT training states is used.

\begin{lemma}[Pathwise identity]
\label{lem:pathwise}
Assume (S1). Let $f:\mathcal X\to\mathbb R$ be locally Lipschitz, $\tau\in(0,1]$,
and $\psi^{f,\tau}_s:=f\circ\Phi_{s\to\tau}(\cdot\mid\vu)$ for $s\in[0,\tau]$.
For every $\vu$ and $\hat\vx_1$, and almost every $\vx_0$,
\begin{equation}
\label{eq:pathwise}
f(\vx_\tau)-f\big(\Phi_{0\to\tau}(\vx_0\mid\vu)\big)
=\int_0^\tau\nabla\psi^{f,\tau}_s(\vx_s)\cdot\big(\vV-\vv^{\mathrm{old}}(\vx_s,s\mid\vu)\big)\,\mathrm ds .
\end{equation}
\end{lemma}
\begin{proof}
Let $\vy(s)=\Phi_{s\to\tau}(\vx_s\mid\vu)$, so that $\vy(0)=\Phi_{0\to\tau}(\vx_0\mid\vu)$ and $\vy(\tau)=\vx_\tau$.
Under (S1) the flow is $C^1$ in $(s,\vx)$ with
$\partial_s\Phi_{s\to\tau}(\vx)=-\nabla\Phi_{s\to\tau}(\vx)\,\vv^{\mathrm{old}}(\vx,s)$,
and $\tfrac{\mathrm d}{\mathrm ds}\vx_s=\vV$, so
$\vy'(s)=\nabla\Phi_{s\to\tau}(\vx_s)\big(\vV-\vv^{\mathrm{old}}(\vx_s,s)\big)$.
Since $\vy$ is $C^1$ and $f$ is locally Lipschitz, $f\circ\vy$ is Lipschitz on $[0,\tau]$, so $f(\vy(\tau))-f(\vy(0))=\int_0^\tau(f\circ\vy)'(s)\,\mathrm ds$.
Wherever $f$ is differentiable at $\vy(s)$, the chain rule gives
$(f\circ\vy)'(s)=\nabla f(\vy(s))\cdot\vy'(s)=\nabla\psi^{f,\tau}_s(\vx_s)\cdot(\vV-\vv^{\mathrm{old}}(\vx_s,s))$.
By Rademacher's theorem, $f$ is differentiable outside a Lebesgue-null set $N$.
For $s<1$, $\vx_s$ has a density given $\hat\vx_1$ and $\Phi_{s\to\tau}$ is a diffeomorphism,
so $\mathbb P(\vy(s)\in N)=0$; by Fubini's theorem, almost surely $\vy(s)\notin N$ for almost every $s$.
\end{proof}

\begin{proposition}[NFT counterpart of the witness identity]
\label{prop:nft_pairing}
Assume (S1). For every Lipschitz $f:\mathcal X\to\mathbb R$, with $\psi^f_t:=f\circ\Phi_{t\to1}(\cdot\mid\vu)$,
\begin{equation}
\label{eq:nft_pairing}
\mathbb E\big[\nabla\psi^f_t(\vx_t\mid\vu)\cdot\Delta(\vx_t,t\mid\vu)\big]
=\mathbb E\Big[\big(f(\hat\vx_{g,j})-\mathbb E[f(\hat\vx_1)\mid\vu_g]\big)\,\widetilde A_{g,j}\Big],
\end{equation}
where the expectation on the left is over the training states, $\vu\sim P_0$, $t\sim\mathcal U[0,1]$, and $\vx_t\sim p_t(\cdot\mid\vu)$.
In particular, for $f=w$ and $\psi_t$ as in \eqref{eq:transfer_value},
\begin{equation}
\label{eq:nft_pairing_witness}
\mathbb E\big[\nabla\psi_t(\vx_t\mid\vu)\cdot\Delta(\vx_t,t\mid\vu)\big]=\kappa .
\end{equation}
\end{proposition}
\begin{proof}
Apply Lemma~\ref{lem:pathwise} with $\tau=1$, multiply by $\widetilde A$, and take expectations.
On the right, $|\nabla\psi^f_t|$ is bounded and $\vV-\vv^{\mathrm{old}}(\vx_t,t)$ is integrable,
so Fubini's theorem and conditioning on $(\vx_t,\vu)$ give the left side of \eqref{eq:nft_pairing}.
On the left, the training noise $\vx_0$ is independent of $(\widetilde A,\hat\vx_1)$ given $\vu$,
and $\Phi_{0\to1}(\vx_0\mid\vu)\sim\pi_{\vu}$, so
$\mathbb E[\widetilde A\,f(\Phi_{0\to1}(\vx_0\mid\vu))]=\mathbb E\big[\mathbb E[\widetilde A\mid\vu]\,\mathbb E[f(\hat\vx_1)\mid\vu]\big]$.
For $f=w$, the right side of \eqref{eq:nft_pairing} is $\kappa$ by \eqref{eq:transfer_kappa_def}.
\end{proof}

Proposition~\ref{thm:witness} pairs the witness with the score $\nabla_\theta\log\pi_\theta$,
which is not available for a flow model.
Eq.~\eqref{eq:nft_pairing_witness} is its counterpart for the NFT loss:
the flow-matching residual $\vV-\vv^{\mathrm{old}}$, weighted by the advantage, takes the place of the score,
and the same within-group covariance between the witness and the weights appears.
By Proposition~\ref{prop:nft_gradient}, the derivative of the policy term of Eq.~\eqref{eq:nft_loss}
at $\vv_\theta=\vv^{\mathrm{old}}$ along a velocity perturbation $\vh$ is $-2\gamma\,\mathbb E[\Delta\cdot\vh]$.
Eq.~\eqref{eq:nft_pairing_witness} therefore says that the NFT loss decreases at rate $2\gamma\kappa$
along $\vh=\nabla\psi$.
Under (S1$'$) and (S3), this is a direction that raises the witness of every generated cell at first order:
by variation of constants, perturbing the velocity by $\varepsilon\nabla\psi$ changes $w(\vX_1)$ by
$\varepsilon\int_0^1|\nabla\psi_t(\vX_t)|^2\,\mathrm dt\ge0$ along every sampler trajectory $\vX_t=\Phi_{0\to t}(\vx_0\mid\vu)$.

\begin{proposition}[Stationary points of the implemented loss]
\label{prop:nft_stationary}
Let $\vv_\theta$ be differentiable in $\theta\in\mathbb R^p$ with Jacobian
$\mJ_\theta(\vx,t\mid\vu)=\partial_\theta\vv_\theta(\vx,t\mid\vu)$,
and let $\|\cdot\|$ be the $L^2$ norm over the training states ($\vu\sim P_0$, $t\sim\mathcal U[0,1]$, $\vx_t\sim p_t(\cdot\mid\vu)$),
with $\|\mJ_\theta\|<\infty$.
Call $\theta$ a \emph{synchronized stationary point} if the gradient of Eq.~\eqref{eq:nft_loss} with respect to $\theta$
vanishes when $\vv^{\mathrm{old}}=\vv_\theta$.
Under (S1), at such a point
\begin{equation}
\label{eq:nft_stationary}
\kappa\le\epsilon_{\tan}\,\|\Delta\|+\frac{\lambda_{\mathrm{reg}}}{\gamma}\,L_we^{\ell}\,\|\vv_\theta-\vv_{\theta_0}\|,
\qquad
\epsilon_{\tan}:=\min_{\boldsymbol\xi\in\mathbb R^p}\big\|\nabla\psi-\mJ_\theta\boldsymbol\xi\big\|,
\end{equation}
and $\|\Delta\|^2\le\mathbb E\|\vV-\vv_\theta(\vx_t,t\mid\vu)\|^2$.
In function space, where $\vv_\theta$ ranges over all square-integrable fields, and with $\lambda_{\mathrm{reg}}=0$,
stationarity means $\Delta=\vzero$ and hence $\kappa=0$.
\end{proposition}
\begin{proof}
By Proposition~\ref{prop:nft_gradient} and the chain rule, the gradient of Eq.~\eqref{eq:nft_loss} at $\vv^{\mathrm{old}}=\vv_\theta$ is
$\mathbb E\big[\mJ_\theta^\top\big(-2\gamma\Delta+2\lambda_{\mathrm{reg}}(\vv_\theta-\vv_{\theta_0})\big)\big]$.
If it vanishes, then
$\mathbb E[(\mJ_\theta\boldsymbol\xi)\cdot\Delta]=\tfrac{\lambda_{\mathrm{reg}}}{\gamma}\mathbb E[(\mJ_\theta\boldsymbol\xi)\cdot(\vv_\theta-\vv_{\theta_0})]$
for every $\boldsymbol\xi$, and \eqref{eq:nft_pairing_witness} gives
\[
\kappa=\mathbb E\big[(\nabla\psi-\mJ_\theta\boldsymbol\xi)\cdot\Delta\big]
+\tfrac{\lambda_{\mathrm{reg}}}{\gamma}\,\mathbb E\big[(\mJ_\theta\boldsymbol\xi)\cdot(\vv_\theta-\vv_{\theta_0})\big].
\]
Take $\mJ_\theta\boldsymbol\xi$ to be the $L^2$ projection of $\nabla\psi$ onto the range of $\mJ_\theta$.
Then $\|\mJ_\theta\boldsymbol\xi\|\le\|\nabla\psi\|\le L_we^{\ell}$,
since $|\nabla\psi_t|\le|\nabla w|\,\|\nabla\Phi_{t\to1}\|$,
and the Cauchy--Schwarz inequality gives \eqref{eq:nft_stationary}.
The bound on $\|\Delta\|$ follows from $|\widetilde A|\le1$ and Jensen's inequality.
In function space the stationarity condition is $-2\gamma\Delta+2\lambda_{\mathrm{reg}}(\vv_\theta-\vv_{\theta_0})=\vzero$.
\end{proof}
The quantity $\epsilon_{\tan}$ measures how well the tangent space of the network represents $\nabla\psi$,
and the second term measures the pull of the reference term toward the pretrained generator.
Up to these two terms, a synchronized stationary point must have $\kappa=0$.
For the witness reward with the exact witness, Corollary~\ref{cor:transfer} shows that $\kappa>0$ whenever $Q_\phi\ne P_\phi$
and bounds $D_\phi$ by $\kappa$ under a coverage condition.

\paragraph{First-order change along the sampler.}
For a Lipschitz perturbation $\vh$ with $|\vh(\vx,t)|\le H(1+|\vx|)$,
let $\vX_t=\Phi_{0\to t}(\vx_0\mid\vu)$ and let $\delta\vX_t$ solve
$\tfrac{\mathrm d}{\mathrm dt}\delta\vX_t=\nabla\vv^{\mathrm{old}}(\vX_t,t)\,\delta\vX_t+\vh(\vX_t,t)$ with $\delta\vX_0=\vzero$.
Define
\begin{equation}
\label{eq:transfer_linear_functional}
\mathcal L_{\vv^{\mathrm{old}}}[\vh]:=\mathbb E\big[\nabla w(\vX_1)\cdot\delta\vX_1\big]
=\int_0^1\mathbb E_{\vu}\mathbb E_{\rho_t(\cdot\mid\vu)}\big[\nabla\psi_t\cdot\vh\big]\,\mathrm dt,
\end{equation}
with the first expectation over $\vu\sim P_0$ and $\vx_0\sim\mathcal N(\vzero,\mI)$.
The second form follows from variation of constants,
$\delta\vX_1=\int_0^1\nabla\Phi_{t\to1}(\vX_t)\,\vh(\vX_t,t)\,\mathrm dt$,
and $\nabla\psi_t=\nabla\Phi_{t\to1}^\top\,\nabla w\circ\Phi_{t\to1}$.

\begin{proposition}[First-order change of the energy distance]
\label{prop:transfer_first_order}
Under (S1), (S1$'$), and (S3), let $Q_\varepsilon$ be the population generated by the velocity $\vv^{\mathrm{old}}+\varepsilon\vh$.
Then $D_\phi(Q_\varepsilon,P)=D_\phi(Q_0,P)-2\varepsilon\,\mathcal L_{\vv^{\mathrm{old}}}[\vh]+O(\varepsilon^2)$.
\end{proposition}
\begin{proof}
Expanding the bilinear form in Eq.~\eqref{eq:witness_restatement_definitions} gives the exact identity
$D_\phi(Q',P)=D_\phi(Q,P)-2\int w\,\mathrm d(Q'-Q)+D_\phi(Q',Q)$.
Couple both flows through the same $(\vu,\vx_0)$.
By (S1) and (S1$'$), $\vX^\varepsilon_1=\vX_1+\varepsilon\,\delta\vX_1+\vr_\varepsilon$ with $|\vr_\varepsilon|\le c\,\varepsilon^2(1+|\vx_0|)^2$
and $|\vX^\varepsilon_1-\vX_1|\le c\,\varepsilon(1+|\vx_0|)$ by Gr\"onwall's inequality,
so (S3) gives $\int w\,\mathrm d(Q_\varepsilon-Q_0)=\varepsilon\,\mathcal L_{\vv^{\mathrm{old}}}[\vh]+O(\varepsilon^2)$.
For the last term, $D_\phi(Q',Q)=-\int(b_{Q'}-b_Q)\,\mathrm d(Q'-Q)$,
and under the coupling this is at most $K_bW_1(Q',Q)\,\mathbb E|\vX^\varepsilon_1-\vX_1|=O(\varepsilon^2)$.
All remainders are $O(\varepsilon^2)$ in expectation because $\vx_0$ is Gaussian.
\end{proof}

\begin{proposition}[Descent of the functional NFT step without self-consistency]
\label{prop:transfer}
Under (S1), (S1$'$), (S2), and (S3), the functional NFT step $\vv^{\mathrm{old}}\mapsto\vv^{\mathrm{old}}+\varepsilon\Delta$ gives
\begin{equation}
\label{eq:transfer_rate}
D_\phi(Q_{\mathrm{new}},P)=D_\phi(Q_{\mathrm{old}},P)-2\varepsilon\kappa+\varepsilon R_{\mathrm{sc}}+O(\varepsilon^2),
\end{equation}
where the residual is the mismatch between the sampler and the training marginals,
\begin{equation}
\label{eq:transfer_residual}
R_{\mathrm{sc}}=-2\int_0^1\mathbb E_{\vu}\Big[\int\nabla\psi_t\cdot\Delta\;\mathrm d\big(\rho_t-p_t\big)(\cdot\mid\vu)\Big]\mathrm dt,
\qquad
|R_{\mathrm{sc}}|\le c_0\,\epsilon_0,
\end{equation}
with $\epsilon_0=\int_0^1\mathbb E_{p_t}\big[(1+|\vx_t|)\,|\ve|\big]\,\mathrm dt$ and
$c_0=2e^{2\ell}(1+B)\big[L_wH_1+e^{\ell}H_0(L_wM+W)\big]$.
The error measure is controlled by the excess flow-matching loss of the generator on its own rollouts,
\begin{equation}
\label{eq:transfer_excess_fm}
\epsilon_0^2\le m_2\Big(\mathbb E\|\vV-\vv^{\mathrm{old}}(\vx_t,t\mid\vu)\|^2-\mathbb E\|\vV-\bar\vv(\vx_t,t\mid\vu)\|^2\Big),
\qquad m_2=\mathbb E\,(1+|\vx_t|)^2 .
\end{equation}
In particular, the update decreases $D_\phi$ at first order whenever $2\kappa>R_{\mathrm{sc}}$, for which $2\kappa>c_0\epsilon_0$ suffices,
and $R_{\mathrm{sc}}=0$ whenever $\rho_t=p_t$ for almost every $t$,
which holds when the generator is self-consistent ($\ve=\vzero$).
\end{proposition}
\begin{proof}
By Proposition~\ref{prop:transfer_first_order} and \eqref{eq:transfer_linear_functional},
the first-order change is $-2\mathcal L_{\vv^{\mathrm{old}}}[\Delta]=-2\int_0^1\mathbb E_{\vu}\mathbb E_{\rho_t}[\nabla\psi_t\cdot\Delta]\,\mathrm dt$.
By \eqref{eq:nft_pairing_witness}, $\kappa=\int_0^1\mathbb E_{\vu}\mathbb E_{p_t}[\nabla\psi_t\cdot\Delta]\,\mathrm dt$.
Subtracting gives \eqref{eq:transfer_rate} with the stated $R_{\mathrm{sc}}$.

For the bound, fix $\vu$ and $t\in(0,1)$ and let $F_t=\nabla\psi_t\cdot\Delta(\cdot,t)$.
By (S1) and (S3), $|\nabla\psi_t|\le L_we^{\ell(1-t)}$.
By (S1$'$) and Gr\"onwall's inequality, $\nabla\Phi_{t\to1}$ is $Me^{2\ell(1-t)}$-Lipschitz,
so $\nabla\psi_t$ is $e^{2\ell(1-t)}(L_wM+W)$-Lipschitz.
With (S2), $F_t$ is locally Lipschitz with $|\nabla F_t(\vx)|\le\Lambda_t(1+|\vx|)$,
where $\Lambda_t=e^{\ell(1-t)}L_wH_1+e^{2\ell(1-t)}H_0(L_wM+W)$.
By (S1), $|\Phi_{s\to t}(\vx)|\le e^{\ell}(|\vx|+B)$, so $1+|\Phi_{s\to t}(\vx)|\le e^{\ell}(1+B)(1+|\vx|)$.
Apply Lemma~\ref{lem:pathwise} with $f=F_t$ and $\tau=t$ and take expectations.
Since $\vx_t\sim p_t$, $\Phi_{0\to t}(\vx_0)\sim\rho_t$, and $\mathbb E[\vV-\vv^{\mathrm{old}}\mid\vx_s,\vu]=\ve$,
\begin{equation}
\label{eq:transfer_marginal_gap}
\int F_t\,\mathrm d(p_t-\rho_t)=\int_0^t\mathbb E_{p_s}\big[\nabla\psi^{F_t,t}_s\cdot\ve\big]\,\mathrm ds,
\qquad
\Big|\int F_t\,\mathrm d(p_t-\rho_t)\Big|\le\Lambda_te^{\ell}(1+B)\int_0^te^{\ell(t-s)}\,\mathbb E_{p_s}\big[(1+|\vx_s|)|\ve|\big]\,\mathrm ds,
\end{equation}
because $|\nabla\psi^{F_t,t}_s(\vx)|\le e^{\ell(t-s)}\Lambda_t\big(1+|\Phi_{s\to t}(\vx)|\big)$.
Integrating over $t$ and exchanging the order of integration,
$\int_s^1\Lambda_te^{\ell(t-s)}\,\mathrm dt\le e^{\ell}L_wH_1+e^{2\ell}H_0(L_wM+W)$, which gives $|R_{\mathrm{sc}}|\le c_0\epsilon_0$.
The identity in \eqref{eq:transfer_marginal_gap} holds for every Lipschitz function in place of $F_t$ and every $t\in(0,1]$,
so $\ve=\vzero$ implies $p_t=\rho_t$ for every $t$.
For \eqref{eq:transfer_excess_fm}, $\bar\vv=\mathbb E[\vV\mid\vx_t,\vu]$ gives the Pythagorean identity
$\mathbb E\|\vV-\vv^{\mathrm{old}}\|^2=\mathbb E\|\vV-\bar\vv\|^2+\mathbb E\|\ve\|^2$,
and the Cauchy--Schwarz inequality gives $\epsilon_0^2\le m_2\int_0^1\mathbb E_{p_t}|\ve|^2\,\mathrm dt$.
\end{proof}

The role of Proposition~\ref{prop:transfer} is to show that the residual is first order in the self-consistency error,
so that the descent rate $2\kappa$ is stable when the generator only approximately fits its own samples.
The rate itself is governed by $\kappa$, and Corollary~\ref{cor:transfer} shows how each implementation choice enters it.

\begin{remark}[The two parts of the NFT direction]
Let $a:=\mathbb E[\widetilde A\mid\vx_t,\vu]\in[-1,1]$ and
$\Delta_{\mathrm{tilt}}:=\operatorname{Cov}\big(f_{\vu}(\hat\vx_1),\vV\mid\vx_t,\vu\big)$.
Given $(\hat\vx_1,\vu)$, $\widetilde A$ depends only on other rollouts, other groups, and real cells,
which are independent of $(\vx_0,t)$.
Hence $\mathbb E[\widetilde A\,\vV\mid\vx_t,\vu]=\operatorname{Cov}(\varphi_{\vu},\vV\mid\vx_t,\vu)+a\,\bar\vv$, and
\begin{equation}
\label{eq:nft_decomposition}
\Delta=\Delta_{\mathrm{tilt}}+a\,\ve .
\end{equation}
The first part is a reweighting of the generator's own samples by their advantages:
because $\vx_0$ is drawn independently of $\hat\vx_1$, the reweighted law
$\pi_{\vu,\varepsilon}=(1+\varepsilon f_{\vu})\,\pi_{\vu}$ has flow-matching velocity
$\mathbb E[(1+\varepsilon f_{\vu})\vV\mid\vx_t,\vu]/\mathbb E[1+\varepsilon f_{\vu}\mid\vx_t,\vu]
=\bar\vv+\varepsilon\Delta_{\mathrm{tilt}}/(1+\varepsilon\,\mathbb E[f_{\vu}\mid\vx_t,\vu])$.
The second part is proportional to the self-consistency error.
When $\ve=\vzero$, we have $\Delta=\Delta_{\mathrm{tilt}}$, $\rho_t=p_t$, and $R_{\mathrm{sc}}=0$.
\end{remark}

\begin{corollary}[What each implementation choice does]
\label{cor:transfer}
For the witness reward, $\widetilde A_{g,j}=\operatorname{clip}(z_j(\widehat w_g),-C,C)/C$, and by exchangeability
\begin{equation}
\label{eq:transfer_kappa}
\kappa=\frac1C\,\mathbb E\big[s_g\,\rho_g(\widehat w,w)\big]+\beta_{\mathrm{clip}},
\qquad
\beta_{\mathrm{clip}}=\mathbb E\Big[\tfrac1m\textstyle\sum_j(w_j-\mu_{\vu})\big(\widetilde A_j-z_j(\widehat w)/C\big)\Big],
\end{equation}
where $s_g$ is the within-group standard deviation (divisor $m$) of $w(\hat\vx_{g,\cdot})$
and $\rho_g$ the within-group correlation.
\begin{enumerate}
\item \emph{Standardization.} With centering only, a group contributes $\operatorname{Cov}_g(\widehat w,w)=s_g\widehat s_g\rho_g$.
Standardization divides this by $\widehat s_g>0$, the within-group standard deviation of $\widehat w$,
so it reweights groups but preserves the sign of each group's contribution.
\item \emph{Exact witness without clipping} ($C\ge\sqrt{m-1}$). Then $\rho_g=1$, $\beta_{\mathrm{clip}}=0$, and $\kappa=\mathbb E[s_g]/C\ge0$.
Equality forces $w$ to be constant $\pi_{\vu}$-a.s.\ for $P_0$-a.e.\ $\vu$.
Under (S1) the flow map from $\mathcal N(\vzero,\mI)$ is a diffeomorphism, so $\pi_{\vu}$ has full support.
Since $w$ is continuous, it is then constant everywhere, and Corollary~\ref{cor:witness_gap} gives
$D_\phi=\mathbb E_Pw-\mathbb E_Qw=0$, i.e., $Q_\phi=P_\phi$.
Thus $\kappa$ has no spurious zeros: $\kappa>0$ whenever $Q_\phi\ne P_\phi$.
Moreover, suppose the target is covered by the generator in feature space:
there are laws $P_{\vu}$ with $\int P_{\vu}\,P_0(\mathrm d\vu)=P_\phi$ and density ratio
$\mathrm dP_{\vu}/\mathrm d\pi_{\vu,\phi}\le\Lambda$ for $P_0$-a.e.\ $\vu$, where $\pi_{\vu,\phi}$ is the feature law of $\pi_{\vu}$
(for instance $P_{\vu}=P_\phi$ for every $\vu$).
Then, for $m\ge2$,
\begin{equation}
\label{eq:transfer_domination}
D_\phi(Q,P)\le\frac{m\,\Lambda}{2(m-1)}\,\mathbb E[s_g]=\frac{m\,\Lambda\,C}{2(m-1)}\,\kappa .
\end{equation}
Indeed, $w$ depends on $\vx$ only through $\phi(\vx)$, and Corollary~\ref{cor:witness_gap} gives
$D_\phi=\mathbb E_{\vu}\operatorname{Cov}_{\pi_{\vu,\phi}}(\mathrm dP_{\vu}/\mathrm d\pi_{\vu,\phi},w)$.
The within-group covariance with divisor $m-1$ is unbiased for this covariance,
and it is at most $\tfrac{m}{m-1}\,s_g\,\Lambda/2$,
since values in $[0,\Lambda]$ have standard deviation at most $\Lambda/2$.
\item \emph{Finite-sample witness.} The reference set of Eq.~\eqref{eq:empirical_witness} excludes the whole group,
so $\widehat w_{g,j}=w(\hat\vx_{g,j})+\xi_g(\hat\vx_{g,j})$,
where $\xi_g$ is a random function, independent of the rollouts of group $g$, with $\mathbb E[\xi_g(\vx)]=0$ for every $\vx$.
It collects the errors of the generated reference average and of the real-cell average,
both of which are independent of group $g$ and unbiased.
Hence the estimate is conditionally unbiased and the unstandardized covariance is unbiased,
$\mathbb E[\operatorname{Cov}_g(\widehat w,w)]=\mathbb E[s_g^2]$.
After standardization, the expected contribution of each group to $\kappa$, taken over the reference draw, is nonnegative
under the assumption that $\xi_g$ and $-\xi_g$ have the same law.
This symmetry holds approximately by the central limit theorem, since $\xi_g$ averages over the $K-1$ other groups and the real cells:
writing the centered vectors as $\boldsymbol a$ (from $w$) and $\boldsymbol n$ (from $\xi_g$),
with $\boldsymbol n=x\,\boldsymbol a/|\boldsymbol a|+\boldsymbol y$ and $\boldsymbol y\perp\boldsymbol a$,
the contribution is $|\boldsymbol a|\,G(|\boldsymbol a|+x)/(\sqrt mC)$ with $G(s)=s/\sqrt{s^2+|\boldsymbol y|^2}$ odd and nondecreasing,
and pairing $\boldsymbol n$ with $-\boldsymbol n$ gives $|\boldsymbol a|[G(|\boldsymbol a|+x)-G(x-|\boldsymbol a|)]\ge0$.
\item \emph{Clipping.} Samuelson's inequality gives $|z_j|\le\sqrt{m-1}$,
so $\beta_{\mathrm{clip}}=0$ whenever $C\ge\sqrt{m-1}$.
With our $m=8$ and $C=2<\sqrt7$, $\sum_jz_j^2=m=2C^2$, so at most one rollout per group has $|z_j|>C$.
Since $|\operatorname{clip}(z_j,-C,C)-z_j|\le\sqrt{m-1}-C$ for that rollout,
\[
|\beta_{\mathrm{clip}}|\le\frac{\sqrt{m-1}-C}{mC}\,\mathbb E\Big[\max_j|w(\hat\vx_{g,j})-\mu_{\vu_g}|\Big],
\]
with prefactor $(\sqrt7-2)/16\approx0.040$ in our setting; its sign is not determined.
The within-group part of the contribution keeps its sign.
With the exact witness, a group contributes
$\tfrac1m\sum_j(w_j-\bar w_g)\widetilde A_j+(\bar w_g-\mu_{\vu})\,\tfrac1m\sum_j\widetilde A_j$,
where $\bar w_g$ is the group mean.
The first term equals $\tfrac{s_g}{mC}\sum_jz_j\operatorname{clip}(z_j,-C,C)$.
When one rollout is clipped, $\sum_jz_j\operatorname{clip}(z_j,-C,C)=C|z_1|+m-z_1^2\ge1+C\sqrt{m-1}$,
so the first term is at least $(1+C\sqrt{m-1})/m$ of its unclipped value $s_g/C$,
which is $(1+2\sqrt7)/8\approx0.79$ in our setting.
Clipping can change the sign of a group's contribution only through the second term,
in which $|\tfrac1m\sum_j\widetilde A_j|\le(\sqrt{m-1}-C)/(mC)$.
In expectation the second term is dominated by the first:
$s_g\ge\tfrac1m\sum_j|w_j-\bar w_g|$ and, by Jensen's inequality,
$\mathbb E|w_1-\bar w_g|\ge\tfrac{m-1}{m}\mathbb E|w_1-\mu_{\vu}|\ge\tfrac{m-1}{m}\mathbb E|\bar w_g-\mu_{\vu}|$,
so $\mathbb E|\bar w_g-\mu_{\vu}|\le\tfrac{m}{m-1}\mathbb E[s_g]$ and
\begin{equation}
\label{eq:transfer_clip_floor}
\kappa\ge c_{\mathrm{clip}}\,\frac{\mathbb E[s_g]}{C},
\qquad
c_{\mathrm{clip}}=\frac1m\Big[1+C\sqrt{m-1}-\frac{m}{m-1}\big(\sqrt{m-1}-C\big)\Big]\approx0.69
\end{equation}
for $m=8$ and $C=2$.
Hence, with the exact witness and the implemented clipping, $\kappa>0$ whenever $Q_\phi\ne P_\phi$,
and \eqref{eq:transfer_domination} holds with $\kappa$ replaced by $\kappa/c_{\mathrm{clip}}$.
In the large-group limit, $\widetilde A=h_{\vu}(w)$ with $h_{\vu}$ nondecreasing and non-constant,
so $\kappa=\mathbb E_{\vu}\operatorname{Cov}(w,h_{\vu}(w))\ge0$ by Chebyshev's association inequality.
\end{enumerate}
\end{corollary}

\paragraph{Optimality of stationary points.}
Combine Proposition~\ref{prop:nft_stationary} with items~2 and~4 of Corollary~\ref{cor:transfer}.
For the witness reward with the exact witness and the implemented clipping ($m=8$, $C=2$),
every synchronized stationary point satisfies
\[
D_\phi(Q,P)\le\frac{m\,\Lambda\,C}{2(m-1)\,c_{\mathrm{clip}}}
\Big[\epsilon_{\tan}\,\|\Delta\|+\frac{\lambda_{\mathrm{reg}}}{\gamma}\,L_we^{\ell}\,\|\vv_\theta-\vv_{\theta_0}\|\Big]
\]
whenever the coverage condition with constant $\Lambda$ holds.
If $\nabla\psi$ lies in the tangent space of the network at $\theta$ ($\epsilon_{\tan}=0$) and $\lambda_{\mathrm{reg}}=0$,
there is no synchronized stationary point with $Q_\phi\ne P_\phi$, and this conclusion needs no coverage condition.

\begin{remark}[Support and pathway penalties]
With $R_{g,j}=z_j(\widehat w_g)-\mathsf P_{g,j}$ (Eq.~\eqref{eq:reward_combination}),
where $\mathsf P$ collects the weighted support and pathway penalties, the same calculation gives
$\kappa=\tfrac1C\mathbb E\big[s_g\big(\rho_g(\widehat w,w)-s^{\mathsf P}_g\rho_g(\mathsf P,w)\big)/s^R_g\big]+\beta_{\mathrm{clip}}$,
with $s^{\mathsf P}_g,s^R_g$ the within-group standard deviations of $\mathsf P$ and $R$.
The penalties change the leading term $\kappa$ only through their within-group standard deviation and correlation
with the witness (they also change the clipping term),
and only in groups that contain a cell below its calibration threshold ($s^{\mathsf P}_g=0$ otherwise).
Descent of $D_\phi$ alone is not guaranteed in this case,
since the penalties are designed to trade population fit for single-cell plausibility.
What the combined reward ranks is the witness minus the penalties:
within each group,
$\widehat s_g\,(R_{g,j}-\bar R_g)=(\widehat w_{g,j}-\bar{\widehat w}_g)-\widehat s_g\,(\mathsf P_{g,j}-\bar{\mathsf P}_g)$,
so ordering the rollouts by $R$ is ordering them by $\widehat w-\widehat s_g\mathsf P$.
The first part is the witness, whose policy gradient decreases the discrepancy term of Eq.~\eqref{eq:post_training_objective}
(Proposition~\ref{thm:witness}), and the second part is the single-cell term $r_{\mathrm{bio}}$ of that objective,
instantiated by the support and pathway rewards and weighted by a group-dependent multiplier $\widehat s_g$.
\end{remark}

\begin{remark}[Scope]
Propositions~\ref{prop:nft_pairing} and~\ref{prop:nft_stationary} apply to the parametric update.
Proposition~\ref{prop:transfer} concerns the functional step of Proposition~\ref{prop:nft_gradient}.
A parametric step with SGD or a fixed diagonal preconditioner changes the velocity by
$2\gamma\eta\,\mathcal K\Delta+O(\eta^2)$, where $\mathcal K$ is the preconditioned tangent kernel of the network.
By Proposition~\ref{prop:transfer_first_order}, it decreases $D_\phi$ at first order exactly when
$\mathcal L_{\vv^{\mathrm{old}}}[\mathcal K\Delta]>0$, which reduces to $\mathcal L_{\vv^{\mathrm{old}}}[\Delta]>0$ when $\mathcal K$ is a multiple of the identity.
\end{remark}

\begin{remark}[Reference term]
At the first iteration $\vv^{\mathrm{old}}=\vv_{\theta_0}$, so the reference term of Eq.~\eqref{eq:nft_loss} has zero gradient
and Proposition~\ref{prop:transfer} applies verbatim.
Afterwards it adds $2\lambda_{\mathrm{reg}}(\vv^{\mathrm{old}}-\vv_{\theta_0})$ to the functional gradient,
which pulls the generator back toward the pretrained model.
Its first-order effect on $D_\phi$ is not signed:
it implements the proximity constraint of Eq.~\eqref{eq:post_training_objective},
which trades population fit against closeness to the pretrained generator.
Its effect at a stationary point is the second term of \eqref{eq:nft_stationary}.
\end{remark}
}

\clearpage
{\color{blue}\section{Additional Quantitative Analysis}
\label{app:additional_quant}

This appendix reports six additional analyses:
a tuned classifier-free guidance baseline (Appendix~\ref{app:cfg}),
the dependence of the discrimination score on effect size (Appendix~\ref{app:ds_scale}),
direct differentiable energy-distance fine-tuning (Appendix~\ref{app:direct_energy}),
variability across training seeds and folds (Appendix~\ref{app:seeds}),
the energy distance during the first post-training updates (Appendix~\ref{app:energy_trend}),
and the dependence of generated cells on the control cell $\vu$ (Appendix~\ref{app:u_dependence}).
All metrics follow Table~\ref{tab:pop_all_settings}.

\subsection{Comparison with Classifier-Free Guidance}
\label{app:cfg}

scDFM is trained without condition dropout and does not support guidance,
so we run this comparison on scLDM with the Replogle setup of Appendix~\ref{app:scldm}.
We sweep the guidance weight $w\in\{0.5,0.75,1,1.25,1.5,1.75,2\}$ in
$\vv=\vv_\varnothing+w(\vv_c-\vv_\varnothing)$~\citep{ho2022classifier}
and keep the weight with the best average rank over the nine metrics ($w=1.5$).
As shown in Table~\ref{tab:cfg},
the best guidance weight lowers Energy by $1.1\%$ and MMD by $5.8\%$,
whereas post-training lowers them by $11.9\%$ and $36.6\%$
and is better on seven of the nine metrics.
Table~\ref{tab:cfg} also reports $w=2$, the largest weight in the sweep.

\begin{table}[H]
\centering
\caption{\textbf{Tuned classifier-free guidance on scLDM (Replogle).}
Mean $\pm$ std over $3$ generation seeds.
Bold indicates best.}
\label{tab:cfg}
\renewcommand{\arraystretch}{1.1}
\resizebox{\linewidth}{!}{%
\begin{tabular}{@{}lcrrrrrrrrr@{}}
\toprule
\textbf{Model} & $w$
& \textbf{\shortstack{Energy\\$\downarrow$}}
& \textbf{\shortstack{MMD\\$\downarrow$}}
& \textbf{\shortstack{SW1\\$\downarrow$}}
& \textbf{\shortstack{NES-Sp.\\$\uparrow$}}
& \textbf{\shortstack{MAE\\$\downarrow$}}
& \textbf{\shortstack{DE-Sp.\\$\uparrow$}}
& \textbf{\shortstack{$\rho(\hat{\Delta})$\\$\uparrow$}}
& \textbf{\shortstack{$\rho(\Delta)$\\$\uparrow$}}
& \textbf{\shortstack{DS\\$\uparrow$}} \\
\midrule
scLDM & $1$
& $2.3945_{\pm0.0159}$ & $0.0154_{\pm0.0003}$ & $0.1979_{\pm0.0007}$ & $0.3706_{\pm0.0060}$ & $0.1329_{\pm0.0006}$ & $0.3978_{\pm0.0026}$ & $0.3167_{\pm0.0056}$ & $0.3270_{\pm0.0016}$ & $0.7518_{\pm0.0032}$ \\
scLDM + guidance & $1.5$
& $2.3692_{\pm0.0123}$ & $0.0145_{\pm0.0002}$ & $0.1955_{\pm0.0005}$ & $0.3936_{\pm0.0066}$ & $0.1334_{\pm0.0005}$ & $0.4207_{\pm0.0021}$ & $0.3683_{\pm0.0039}$ & $0.3498_{\pm0.0021}$ & $0.7983_{\pm0.0040}$ \\
scLDM + guidance & $2$
& $2.4466_{\pm0.0096}$ & $0.0155_{\pm0.0002}$ & $0.1972_{\pm0.0004}$ & $0.3959_{\pm0.0040}$ & $0.1372_{\pm0.0004}$ & $\mathbf{0.4304}_{\pm0.0061}$ & $\mathbf{0.3942}_{\pm0.0029}$ & $\mathbf{0.3562}_{\pm0.0024}$ & $\mathbf{0.8124}_{\pm0.0039}$ \\
scLDM + \method{} & $1$
& $\mathbf{2.1088}_{\pm0.0121}$ & $\mathbf{0.0098}_{\pm0.0002}$ & $\mathbf{0.1829}_{\pm0.0005}$ & $\mathbf{0.4030}_{\pm0.0035}$ & $\mathbf{0.1210}_{\pm0.0004}$ & $\mathbf{0.4304}_{\pm0.0020}$ & $0.3369_{\pm0.0045}$ & $0.3514_{\pm0.0007}$ & $0.7617_{\pm0.0044}$ \\
\bottomrule
\end{tabular}%
}
\end{table}

\paragraph{Why guidance raises $\rho(\hat{\Delta})$ and DS.}
As $w$ grows, $\rho(\hat{\Delta})$ and DS keep rising (Table~\ref{tab:cfg}).
We measure the effect size of a prediction as the $L_1$ norm of its pseudobulk effect
(pseudobulk minus the control mean) divided by that of the real effect, averaged over test perturbations.
As shown in Table~\ref{tab:cfg_scale}, scLDM already overestimates the effect size,
guidance enlarges it further, and \method{} is closest to the real size.
$L_1$ DS increases when predicted effects are scaled up~\citep{liu2025effects};
following their fix, we rescale each predicted effect to the $L_1$ norm of the real effect before computing DS (norm-matched DS).
The Pearson correlations are computed on pseudobulks of about $71$ generated cells per perturbation,
so sampling noise attenuates them, and a larger effect stands out more against this noise.
We estimate this noise from the three independent generation seeds:
the correlation between the pseudobulks of two seeds of the same model measures their reliability,
and dividing the observed correlation by the square root of the mean reliability
gives the noise-corrected correlation~\citep{spearman1904proof}.
The perturbation-specific correlation uses each pseudobulk minus its mean over the $372$ test perturbations.
Differences are tested with a paired bootstrap over test perturbations ($10{,}000$ resamples).
After norm matching, guidance at $w=1.5$ and \method{} are tied on DS, whereas guidance at $w=2$ keeps a lead of $1.0$ point.
After noise correction, \method{} is significantly higher than guidance on $\rho(\Delta)$ at both weights ($p\le0.002$).
On $\rho(\hat{\Delta})$, the gap at $w=1.5$ narrows from $0.031$ to $0.010$ (guidance still higher, $p=0.003$);
this remainder comes from how the average of the test perturbations differs from the training centroid that $\rho(\hat{\Delta})$ subtracts, which is the same for every test perturbation,
since the perturbation-specific correlations of guidance at $w=1.5$ and \method{} are tied ($p=0.31$).
Most of the advantage of guidance on these scores thus comes from a larger effect size,
which moves predictions away from the real effect and worsens MAE and the distribution metrics.

\begin{table}[H]
\centering
\caption{\textbf{Effect size, norm-matched DS, and noise-corrected correlations on scLDM (Replogle).}
Effect size is relative to the real effect ($1$ is exact).
Seed-to-seed agreement is the mean correlation between the perturbation-specific pseudobulks of two generation seeds.
Correlations are corrected for sampling noise using three generation seeds.
Bold indicates best.}
\label{tab:cfg_scale}
\renewcommand{\arraystretch}{1.1}
\begin{tabular}{@{}lcrrrrrr@{}}
\toprule
\textbf{Model} & $w$
& \textbf{\shortstack{Effect\\size}}
& \textbf{\shortstack{Seed-to-seed\\agreement}}
& \textbf{\shortstack{Norm-matched\\DS $\uparrow$}}
& \textbf{\shortstack{$\rho(\Delta)$\\$\uparrow$}}
& \textbf{\shortstack{$\rho(\hat{\Delta})$\\$\uparrow$}}
& \textbf{\shortstack{Perturbation-\\specific $\uparrow$}} \\
\midrule
scLDM & $1$ & $1.27$ & $0.343$ & $0.763$ & $0.490$ & $0.481$ & $0.534$ \\
scLDM + guidance & $1.5$ & $1.34$ & $0.479$ & $0.790$ & $0.495$ & $0.504$ & $0.534$ \\
scLDM + guidance & $2$ & $1.41$ & $0.569$ & $\mathbf{0.801}$ & $0.480$ & $\mathbf{0.505}$ & $0.528$ \\
scLDM + \method{} & $1$ & $\mathbf{1.14}$ & $0.428$ & $0.791$ & $\mathbf{0.506}$ & $0.494$ & $\mathbf{0.536}$ \\
\bottomrule
\end{tabular}
\end{table}

\subsection{Discrimination Score and Effect Size on Norman}
\label{app:ds_scale}

On Norman, \method{} lowers DS by $1.9$--$3.1\%$ (Table~\ref{tab:pop_all_settings}).
DS ranks perturbations by the $L_1$ distance between absolute control-relative pseudobulk effects
(Appendix~\ref{app:evaluation_metrics}), which favors larger predicted effects~\citep{liu2025effects}.
Over the $308$ held-out condition--fold pairs in the four settings,
scDFM overestimates the effect size in $69\%$, and \method{} is closer to the real effect size in $66\%$.
With norm-matched DS (Appendix~\ref{app:cfg}), the two models differ by at most $0.2$ points on Norman,
and no difference is significant (paired Wilcoxon test over conditions, $p\ge0.29$; Table~\ref{tab:ds_scale}).
Distribution-level metrics capture what DS cannot.
Within a perturbation, cells can respond strongly, weakly, or not at all~\citep{bunne2023learning,dong2023causal,papalexi2021characterizing},
and two populations with the same mean shift but different responding fractions receive the same DS but different energy distances.
Energy distance is the standard measure of perturbation effects in scPerturb~\citep{peidli2024scperturb},
benchmarks evaluate the full distribution of responses~\citep{wu2025perturbench,li2026systematic},
State trains with a distributional loss to capture cellular heterogeneity~\citep{adduri2025predicting},
and mean-based metrics can reward predictions that collapse to an average response~\citep{mejia2025diversity}.

\begin{table}[H]
\centering
\caption{\textbf{DS before and after norm matching.}
Norman settings are four-fold means.}
\label{tab:ds_scale}
\renewcommand{\arraystretch}{1.1}
\begin{tabular}{@{}lrrrr@{}}
\toprule
& \multicolumn{2}{c}{\textbf{DS $\uparrow$}} & \multicolumn{2}{c}{\textbf{Norm-matched DS $\uparrow$}} \\
\cmidrule(lr){2-3}\cmidrule(lr){4-5}
\textbf{Setting} & scDFM & \method{} & scDFM & \method{} \\
\midrule
Additive & $0.9743$ & $0.9558$ & $0.9954$ & $0.9943$ \\
Holdout single & $0.8904$ & $0.8630$ & $0.9124$ & $0.9125$ \\
Holdout double & $0.9144$ & $0.8911$ & $0.9622$ & $0.9600$ \\
ComboSciPlex & $0.8776$ & $0.8980$ & $0.8980$ & $0.8980$ \\
\bottomrule
\end{tabular}
\end{table}

\subsection{Comparison with Direct Energy-Distance Fine-Tuning}
\label{app:direct_energy}

We replace NFT with fine-tuning that backpropagates the scaled energy distance
through the $20$-point RK4 sampler,
keeping everything else identical to the witness-only run on ComboSciPlex
(Table~\ref{tab:holdout_reward_ablation}), including the learning rate of $5\times10^{-7}$.
To keep the model close to the pretrained generator,
we apply the reference-velocity penalty of NFT at the states of the sampling chain.
Because of z-scoring, the energy distance in this objective has a different scale from the reward used by NFT,
so we rescale the penalty weight by the same factor, to $30$,
which keeps the penalty-to-objective ratio identical to that of NFT.
Backpropagating through the sampler runs out of memory on an 80\,GB H100
without gradient checkpointing,
and each step costs $5.6\times$ an NFT step.
As shown in Table~\ref{tab:direct_energy},
at a comparable compute budget, NFT is better than or tied with direct fine-tuning on all nine metrics.
The support and pathway rewards are calibrated tail ranks and are not differentiable,
so direct fine-tuning can optimize only the witness term,
whereas NFT needs only reward values and optimizes the combined reward of \eqref{eq:reward_combination}.

\begin{table}[H]
\centering
\caption{\textbf{Direct energy-distance fine-tuning versus witness-only NFT on ComboSciPlex.}
GPU time is measured on one H100.
Bold indicates best.}
\label{tab:direct_energy}
\renewcommand{\arraystretch}{1.1}
\resizebox{\linewidth}{!}{%
\begin{tabular}{@{}lcrrrrrrrrr@{}}
\toprule
\textbf{Method} & \textbf{\shortstack{Steps /\\GPU time}}
& \textbf{\shortstack{Energy\\$\downarrow$}}
& \textbf{\shortstack{MMD\\$\downarrow$}}
& \textbf{\shortstack{SW1\\$\downarrow$}}
& \textbf{\shortstack{NES-Sp.\\$\uparrow$}}
& \textbf{\shortstack{MAE\\$\downarrow$}}
& \textbf{\shortstack{DE-Sp.\\$\uparrow$}}
& \textbf{\shortstack{$\rho(\hat{\Delta})$\\$\uparrow$}}
& \textbf{\shortstack{$\rho(\Delta)$\\$\uparrow$}}
& \textbf{\shortstack{DS\\$\uparrow$}} \\
\midrule
scDFM & -- & 0.5177 & 0.0462 & 0.0654 & 0.8919 & 0.0242 & 0.8363 & 0.8418 & 0.8679 & 0.8776 \\
Direct fine-tuning & $40$ / $48$\,min & 0.4529 & 0.0361 & 0.0586 & 0.8905 & 0.0234 & 0.8403 & 0.8484 & 0.8725 & \textbf{0.8980} \\
NFT, witness only & $200$ / $43$\,min & \textbf{0.4413} & \textbf{0.0358} & \textbf{0.0581} & \textbf{0.9028} & \textbf{0.0231} & \textbf{0.8410} & \textbf{0.8567} & \textbf{0.8797} & \textbf{0.8980} \\
\bottomrule
\end{tabular}%
}
\end{table}

\subsection{Variability across Training Seeds}
\label{app:seeds}

We retrain \method{} on ComboSciPlex with three training seeds (Table~\ref{tab:seeds_csp}).
Every mean gain over scDFM is at least $3.8$ times the seed-to-seed standard deviation,
and at least $10$ times for every metric except NES-Sp.
For NES-Sp., every seed ($0.8968$, $0.9003$, $0.8992$) is above scDFM ($0.8919$).
DS is identical across seeds because ComboSciPlex has only seven test conditions.

Norman has four folds, each a separate post-training run evaluated on its own held-out perturbations.
Since the test sets differ across folds,
Table~\ref{tab:norman_paired} reports the paired gain within each fold
(\method{} minus scDFM; means in Table~\ref{tab:pop_all_settings}).
Energy, MMD, SW1, and MAE improve in every fold of all three settings.
NES-Sp.\ improves in all four folds on additive and holdout single;
on holdout double, two folds improve ($+0.017$, $+0.012$) and two are slightly lower ($-0.003$, $-0.004$).

\begin{table}[H]
\centering
\caption{\textbf{Three training seeds on ComboSciPlex.}
Mean $\pm$ std over training seeds.
Bold indicates best.}
\label{tab:seeds_csp}
\renewcommand{\arraystretch}{1.1}
\resizebox{\linewidth}{!}{%
\begin{tabular}{@{}lrrrrrrrrr@{}}
\toprule
\textbf{Model}
& \textbf{\shortstack{Energy\\$\downarrow$}}
& \textbf{\shortstack{MMD\\$\downarrow$}}
& \textbf{\shortstack{SW1\\$\downarrow$}}
& \textbf{\shortstack{NES-Sp.\\$\uparrow$}}
& \textbf{\shortstack{MAE\\$\downarrow$}}
& \textbf{\shortstack{DE-Sp.\\$\uparrow$}}
& \textbf{\shortstack{$\rho(\hat{\Delta})$\\$\uparrow$}}
& \textbf{\shortstack{$\rho(\Delta)$\\$\uparrow$}}
& \textbf{\shortstack{DS\\$\uparrow$}} \\
\midrule
scDFM & 0.5177 & 0.0462 & 0.0654 & 0.8919 & 0.0242 & 0.8363 & 0.8418 & 0.8679 & 0.8776 \\
\method{}
& $\mathbf{0.4470}_{\pm0.0026}$ & $\mathbf{0.0362}_{\pm0.0003}$ & $\mathbf{0.0585}_{\pm0.0002}$ & $\mathbf{0.8988}_{\pm0.0018}$ & $\mathbf{0.0231}_{\pm0.0001}$ & $\mathbf{0.8423}_{\pm0.0006}$ & $\mathbf{0.8542}_{\pm0.0008}$ & $\mathbf{0.8774}_{\pm0.0006}$ & $\mathbf{0.8980}_{\pm0.0000}$ \\
\bottomrule
\end{tabular}%
}
\end{table}

\begin{table}[H]
\centering
\caption{\textbf{Paired gains on Norman.}
\method{} minus scDFM within each fold, mean $\pm$ std over the four folds.}
\label{tab:norman_paired}
\renewcommand{\arraystretch}{1.1}
\resizebox{\linewidth}{!}{%
\begin{tabular}{@{}lrrrrrrrrr@{}}
\toprule
\textbf{Setting}
& \textbf{\shortstack{Energy\\$\downarrow$}}
& \textbf{\shortstack{MMD\\$\downarrow$}}
& \textbf{\shortstack{SW1\\$\downarrow$}}
& \textbf{\shortstack{NES-Sp.\\$\uparrow$}}
& \textbf{\shortstack{MAE\\$\downarrow$}}
& \textbf{\shortstack{DE-Sp.\\$\uparrow$}}
& \textbf{\shortstack{$\rho(\hat{\Delta})$\\$\uparrow$}}
& \textbf{\shortstack{$\rho(\Delta)$\\$\uparrow$}}
& \textbf{\shortstack{DS\\$\uparrow$}} \\
\midrule
Additive
& $-0.0697_{\pm0.0170}$ & $-0.0092_{\pm0.0013}$ & $-0.0066_{\pm0.0011}$ & $+0.0097_{\pm0.0078}$ & $-0.0013_{\pm0.0005}$ & $-0.0011_{\pm0.0016}$ & $+0.0146_{\pm0.0065}$ & $+0.0074_{\pm0.0045}$ & $-0.0184_{\pm0.0133}$ \\
Holdout single
& $-0.0622_{\pm0.0142}$ & $-0.0103_{\pm0.0020}$ & $-0.0066_{\pm0.0012}$ & $+0.0206_{\pm0.0092}$ & $-0.0009_{\pm0.0005}$ & $-0.0030_{\pm0.0033}$ & $+0.0251_{\pm0.0079}$ & $+0.0099_{\pm0.0058}$ & $-0.0274_{\pm0.0256}$ \\
Holdout double
& $-0.0436_{\pm0.0256}$ & $-0.0067_{\pm0.0030}$ & $-0.0049_{\pm0.0017}$ & $+0.0055_{\pm0.0107}$ & $-0.0005_{\pm0.0003}$ & $-0.0067_{\pm0.0036}$ & $+0.0092_{\pm0.0122}$ & $+0.0033_{\pm0.0076}$ & $-0.0233_{\pm0.0222}$ \\
\bottomrule
\end{tabular}%
}
\end{table}

\subsection{Energy Distance during Post-Training}
\label{app:energy_trend}

To check that the implemented NFT updates lower the energy distance in practice (Appendix~\ref{app:transfer}),
we score the non-EMA weights after each of the first $20$ updates on ComboSciPlex
(the recipe of Table~\ref{tab:pop_all_settings})
on a fixed probe of $32$ training batches with fixed initial noise,
so that changes come only from the updates.
Energy is scaled by $s_{\mathrm{wit},c}$ from \eqref{eq:gene_witness_reward}.
As shown in Table~\ref{tab:energy_trend},
after $20$ updates the energy distance is lower on all $32$ batches
(paired $t$-test, $p<10^{-22}$), with either the full or the witness-only reward.
By the end of post-training, the witness alone lowers the test-set Energy
by $6.5$--$14.8\%$ in all four settings (Table~\ref{tab:holdout_reward_ablation}).

\begin{table}[H]
\centering
\caption{\textbf{Training-set energy distance over the first $20$ NFT updates on ComboSciPlex.}
Mean over $32$ fixed probe batches.}
\label{tab:energy_trend}
\renewcommand{\arraystretch}{1.1}
\begin{tabular}{@{}lrrrrrr@{}}
\toprule
Update & $0$ & $5$ & $10$ & $15$ & $20$ & Change \\
\midrule
Full reward & 0.2837 & 0.2803 & 0.2740 & 0.2661 & 0.2593 & $-8.6\%$ \\
Witness only & 0.2837 & 0.2804 & 0.2734 & 0.2665 & 0.2613 & $-7.9\%$ \\
\bottomrule
\end{tabular}
\end{table}

\subsection{Dependence on the Control Cell}
\label{app:u_dependence}

We test whether generated cells depend on their control cell $\vu$
(Section~\ref{sec:problem_setup}),
using all $308$ held-out condition--fold pairs in the four settings.
Each condition has $128$ generated cells, each from a distinct control cell,
and scDFM and \method{} receive identical $(\vu,\vx_0)$ inputs.
For each condition, we compute the mean Pearson correlation between each generated cell and its own $\vu$,
both centered across the condition's cells,
and test it against shuffled control cells (permutation test, $199$ permutations).
This test detects an added $\vu$-dependent component once it explains about $0.5\%$ of the generated variance.
As shown in Table~\ref{tab:u_dependence}, both models are at chance level,
since about $5\%$ of conditions are expected to reach $p<0.05$ if the output ignores $\vu$.
The variation among generated cells thus comes from the noise $\vx_0$ rather than from $\vu$,
consistent with the flow-matching objective of \eqref{eq:flow_matching_objective},
in which $\vu$ and $\vx_1$ are sampled independently given $c$.

\begin{table}[H]
\centering
\caption{\textbf{Dependence of generated cells on their control cell} over $308$ held-out condition--fold pairs.}
\label{tab:u_dependence}
\renewcommand{\arraystretch}{1.1}
\begin{tabular}{@{}lcc@{}}
\toprule
Model & Correlation with own $\vu$ & Conditions with $p<0.05$ \\
\midrule
scDFM & 0.001 & $6.2\%$ \\
\method{} & 0.002 & $6.2\%$ \\
\bottomrule
\end{tabular}
\end{table}
}

\clearpage
\section{\texorpdfstring{\textcolor{blue}{Additional Qualitative Comparison}}{Additional Qualitative Comparison}}
\label{app:additional_visualizations}

\begin{figure}[htbp]
\centering
\includegraphics[width=\linewidth]{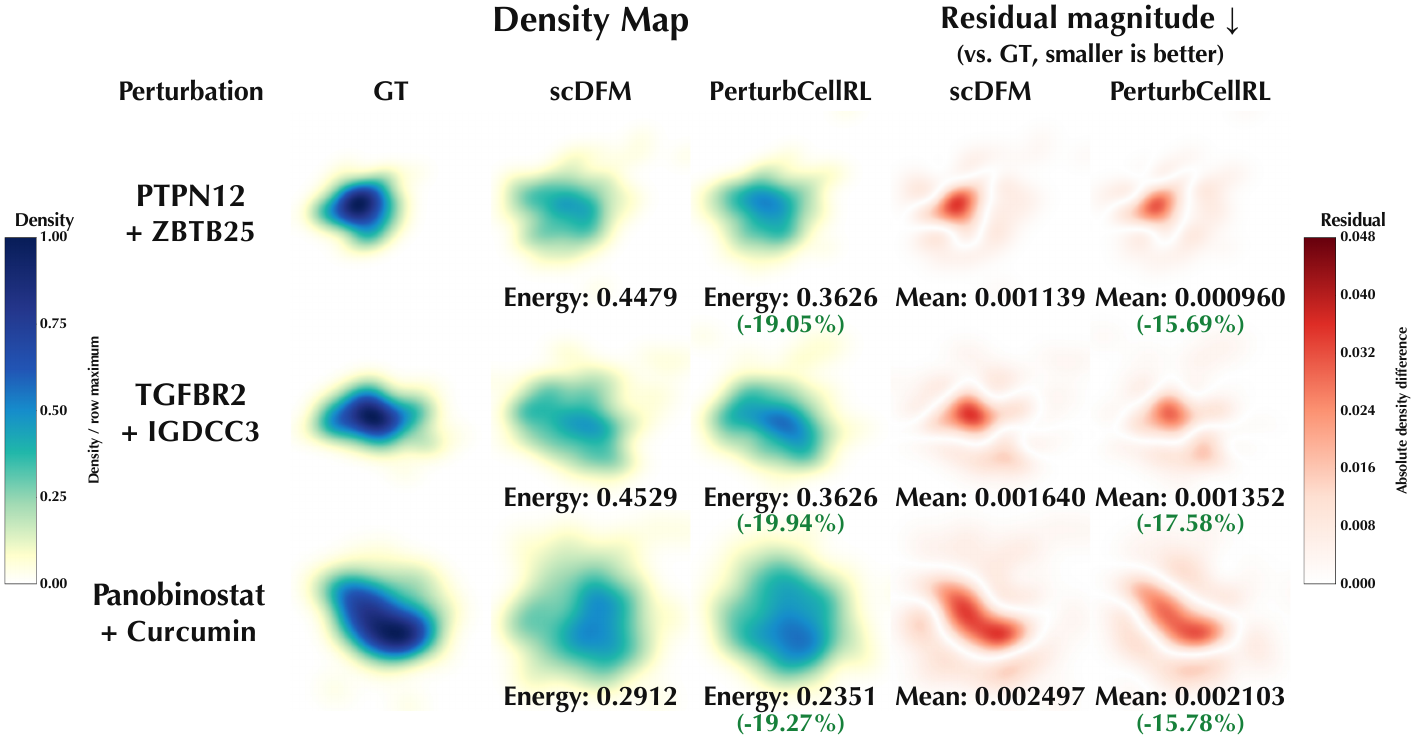}
\caption{
\textbf{Cell-density comparisons for additional perturbations.}
Rows show PTPN12 + ZBTB25, TGFBR2 + IGDCC3,
and Panobinostat + Curcumin.
The first three columns compare ground truth (GT), scDFM, and \method{}.
Densities are normalized by the maximum within each row.
The last two columns show each model's absolute density difference
from GT, on a shared residual scale; smaller values are better.
Annotations report Energy distance and its relative reduction.
}
\label{fig:visualization_supplementary}
\end{figure}

Figure~\ref{fig:visualization_supplementary} extends the qualitative comparison
in Figure~\ref{fig:visualization} to three additional perturbations,
including double-gene and drug-combination conditions.

\FloatBarrier

\clearpage
\section{Responsible Use}
\label{app:responsible_use}

\emph{\method{}} is intended as a decision-support method for
prioritizing biological hypotheses, not as a replacement for
experimental validation.
Predictions can be wrong when perturbations are outside the
training distribution, when cell states are poorly represented,
or when reward models encode incomplete biological knowledge.
Any candidate therapeutic or biological conclusion suggested by
the model should be validated with independent experiments.

\end{document}